\documentclass[letterpaper]{article} % DO NOT CHANGE THIS
\usepackage[preprint]{aaai2027} % DO NOT CHANGE THIS
\usepackage[hyphens]{url} % DO NOT CHANGE THIS
\usepackage{graphicx} % DO NOT CHANGE THIS
\usepackage{natbib} % DO NOT CHANGE THIS AND DO NOT ADD OPTIONS
\usepackage{caption} 
\usepackage{algorithm}
\usepackage{algorithmic}
\usepackage{amsmath,amssymb,amsfonts}
\usepackage{xspace}
\usepackage[table]{xcolor}
\usepackage{multirow}
\usepackage{booktabs}
\usepackage{tabularx}
\usepackage{array}
\usepackage{subcaption} 

\usepackage{pifont}

\newcommand{\cmark}{\ding{51}}
\newcommand{\xmark}{\ding{55}}

\usepackage[table]{xcolor}

\providecolor{fullkvgray}{gray}{0.92}
\providecolor{qevictyellow}{RGB}{255,249,196}

\newcolumntype{V}{
  !{\color{black!40}\vrule width 0.35pt}
}

\providecommand{\lbbest}[1]{\textbf{#1}}
\providecommand{\lbsecond}[1]{\underline{#1}}
\newcommand{\pair}[2]{#1/#2}

\newcommand{\fullkv}{FullKV\xspace}

\newcommand{\TopK}{\operatorname{TopK}}

\definecolor{qevictrow}{RGB}{255,250,205}
\definecolor{fullkvrow}{RGB}{238,238,238}
\definecolor{sectionrow}{RGB}{242,245,248}
\definecolor{qevictrow}{RGB}{255,248,210}
\definecolor{quantrow}{RGB}{239,247,255}
\definecolor{qevictyellow}{RGB}{255,248,210}
\definecolor{qevictrow}{RGB}{255,250,205}
\definecolor{quantrow}{RGB}{240,248,255}
\definecolor{qevictyellow}{RGB}{255,248,210}
\definecolor{fullkvgray}{RGB}{238,238,238}

\newcolumntype{V}{
  !{\color{black!40}\vrule width 0.35pt}
}

\title{QEvict: Recoverable Quantized KV Eviction for Attention-Drift-Robust Long-Context Decoding}

\author{
Ayushman Garg\textsuperscript{\rm 1,*},
Akshita Gupta\textsuperscript{\rm 1,*},
Shaswata Bhattacharya\textsuperscript{\rm 1,*},
Abhishek Gupta\textsuperscript{\rm 2,*},
Sandeep Kumar\textsuperscript{\rm 2,3},
Manoj Kumar\textsuperscript{\rm 1}
}

\affiliations{
\textsuperscript{\rm 1}Mehta Family School of Data Science and Artificial Intelligence,
Indian Institute of Technology Roorkee\\
\textsuperscript{\rm 2}Yardi School of Artificial Intelligence,
Indian Institute of Technology Delhi\\
\textsuperscript{\rm 3}Department of Electrical Engineering,
Indian Institute of Technology Delhi
}

\def\myarch{\texttt{QEvict}\xspace}

\def\fmm{\texttt{Future Missed Mass}\xspace}

\def\lir{\texttt{Global LIR}\xspace}

\begin{document}
\maketitle

\begingroup
\renewcommand{\thefootnote}{\fnsymbol{footnote}}
\footnotetext[1]{Equal Contribution}
\endgroup

\begin{abstract}

Autoregressive large language model inference is increasingly constrained by the memory footprint of the Key-Value (KV) cache. A dominant line of work reduces this footprint by evicting tokens that appear unimportant under attention-derived scores. However, such policies make an implicit irreversible decision: once a token is evicted, it cannot become useful again. We show that this assumption is brittle during decoding. Token and window importance drift as generated queries evolve, causing standard eviction policies to permanently discard states that later receive substantial attention under the full-cache model. To characterize this behaviour, we introduce \emph{Future Missed Mass} and \emph{Global LIR}, two diagnostics that measure future attention assigned to discarded states and the reactivation of historically inactive regions. We propose \myarch, a three-tier KV-cache management scheme that replaces binary retain-or-delete eviction with \emph{recoverable eviction}. \myarch maintains high-confidence windows in full precision, stores intermediate windows in a quantized recoverable tier, and deletes only the lowest-confidence windows. During decoding, cumulative attention scores update window importance and when a quantized window becomes important again, it is dequantized and promoted to the full-precision. Under a fixed memory budget, this design preserves broader historical context while retaining exact full precision for the most important regions. Across long-context understanding, retrieval, and reasoning benchmarks, \myarch consistently improves over representative eviction and quantization baselines, reducing missed attention and improving information retention.

\end{abstract}

\section{Introduction}
\label{sec:introduction}

Transformer-based large language models (LLMs) have achieved remarkable success across language understanding, generation, and reasoning \citep{vaswani2017attention,brown2020language,touvron2023llama}. Their deployment in long-context settings, however, is increasingly constrained by the key--value (KV) cache \citep{pope2023efficiently}. Because the cache grows linearly with sequence length and batch size, its memory footprint can rival or exceed that of the model parameters at long context lengths. This limits batch capacity, increases inference latency, and raises serving costs. Hardware-aware attention kernels improve computational efficiency \citep{dao2023flashattention2}, while memory-management systems reduce allocation overhead \citep{kwon2023pagedattention}; neither changes the fundamental storage complexity of the KV cache.

\begin{table}[t]
\centering
\caption{
Comparison of \myarch{} with representative KV-cache compression methods.
A \cmark\ indicates the presence of a desirable property, while
\xmark\ indicates its absence.
}
\label{tab:qevict_comparison}

\begingroup
\scriptsize
\setlength{\tabcolsep}{2pt}
\renewcommand{\arraystretch}{0.88}

\resizebox{1\columnwidth}{!}{%
\begin{tabular}{@{}lcccc@{}}
\toprule
\textbf{Method}
& \shortstack{\textbf{Selective}\\\textbf{Eviction}}
& \shortstack{\textbf{Low-Bit}\\\textbf{Retention}}
& \shortstack{\textbf{Window}\\\textbf{Routing}}
& \shortstack{\textbf{Dynamic}\\\textbf{Recovery}} \\
\midrule
StreamingLLM~\citep{xiao2024streamingllm}
& \cmark & \xmark & \xmark & \xmark \\
SnapKV~\citep{li2024snapkv}
& \cmark & \xmark & \xmark & \xmark \\
AdaKV~\citep{feng2024adakv}
& \cmark & \xmark & \xmark & \xmark \\
CriticalKV~\citep{feng2025criticalkv}
& \cmark & \xmark & \xmark & \xmark \\
DefensiveKV~\citep{feng2026defensivekv}
& \cmark & \xmark & \xmark & \xmark \\
\midrule
KIVI~\citep{liu2024kivi}
& \xmark & \cmark & \xmark & \xmark \\
KVQuant~\citep{hooper2024kvquant}
& \xmark & \cmark & \xmark & \xmark \\
ZipCache~\citep{he2024zipcache}
& \xmark & \cmark & \xmark & \xmark \\
\midrule
\rowcolor{gray!15}
\textbf{\myarch{} (Ours)}
& \cmark & \cmark & \cmark & \cmark \\
\bottomrule
\end{tabular}%
}

\endgroup
\end{table}

A substantial body of work therefore compresses the KV cache through eviction. StreamingLLM~\citep{xiao2024streamingllm} retains a small set of attention sinks together with recent tokens, while H$_2$O \citep{zhang2023h2o} exploits attention sparsity to preserve accumulated heavy hitters. Subsequent methods refine the selection process through prompt-time observation \citep{li2024snapkv}, adaptive budget allocation across attention heads \citep{feng2024adakv}, value-aware importance estimation \citep{feng2025criticalkv}, and robustness to uncertain future attention \citep{feng2026defensivekv}. In parallel, quantization methods such as KIVI~\citep{liu2024kivi}, KVQuant~\citep{hooper2024kvquant}, and ZipCache~\citep{he2024zipcache} retain broader historical coverage by representing KV states at reduced precision.

Despite this progress, prevailing eviction methods retain a fundamental limitation: cache management is typically formulated as a hard and irreversible token-level decision. Natural language is structured through local lexical cohesion, entity continuity, and discourse relations that extend across neighbouring tokens and sentences~\citep{grosz-etal-1995-centering,hearst-1997-text,passonneau-litman-1997-discourse, barzilay-lapata-2008-modeling,koshorek-etal-2018-text}. Contextualized representations are also strongly shaped by their surrounding context~\citep{ethayarajh-2019-contextual}, and Transformer attention captures linguistic dependencies such as syntax and coreference across related tokens~\citep{clark-etal-2019-bert}. Consistent with this structure, explicit modelling of contiguous spans improves tasks that require question answering, coreference resolution, and relation extraction~\citep{joshi-etal-2020-spanbert}. Independently selecting individual tokens can therefore fragment locally coherent evidence by retaining isolated high-scoring states while discarding neighbouring states that contribute to their interpretation.

A second limitation is that token importance changes throughout decoding. The relevance of cached states is query-dependent, and the subset required for attention can vary substantially across decoding steps~\citep{pmlr-v235-tang24l,pmlr-v235-ribar24a}. A state that appears unimportant under the current query may become critical later when the model resolves a reference, retrieves supporting evidence, or advances along a new reasoning trajectory. Because eviction is irreversible, information removed during a period of low apparent relevance cannot be recovered when its importance re-emerges.

Quantization preserves broader historical coverage by representing cached states at reduced precision. Existing methods, however, generally do not jointly manage cache residency and numerical precision. They lack an explicit mechanism for dynamically moving historical states among full-precision execution, compact recoverable storage, and permanent removal. This leaves a central design gap:

\begin{quote}
    \emph{How can the KV cache be compressed aggressively while preserving coherent historical regions whose relevance may emerge only later?}
\end{quote}

\noindent To address this gap, we propose \textbf{\myarch}, a recoverable three-tier KV-cache hierarchy over contiguous windows. \myarch periodically ranks historical windows using cumulative attention, retaining high-importance windows in full precision, storing intermediate windows in a compact low-bit tier, and evicting the remainder. Quantized windows participate in attention and can be promoted when their importance re-emerges, converting eviction from a one-shot decision into a dynamic process of demotion, recovery, and promotion.

Window-level routing preserves local context and reduces sensitivity to transient token-level fluctuations, while the recoverable tier expands historical coverage under a fixed memory budget. To prevent error accumulation across repeated tier transitions, \myarch maintains a persistent quantized backing store: each window is quantized only on its first demotion, and the resulting low-bit representation is reused for all subsequent recoveries and promotions. The implementation supports GQA~\cite{ainslie2023gqa} and FlashAttention-2~\cite{dao2023flashattention2}, while selectively materializing SDPA attention at routing steps to compute cumulative attention scores.

Our design is motivated by a systematic study of decoding-time cache dynamics. We find that historical importance is highly concentrated but temporally unstable: a small subset of windows dominates attention at any given step, yet the identity of this subset changes throughout generation, and previously inactive windows can later become relevant. To characterize this behaviour, we introduce \fmm and \lir, which measure future attention assigned to discarded states and the reactivation of historically inactive windows, respectively. We additionally use \emph{Selection Churn} to quantify instability between consecutive cache assignments. Together, these observations motivate preserving uncertain historical states without allocating full-precision memory to all of them.

We evaluate \myarch on LongBench~\cite{bai2024longbench}, RULER~\cite{hsieh2024ruler}, and GSM8K~\cite{cobbe2021training} across three instruction-tuned LLMs and multiple KV-memory budgets. The evaluation covers realistic long-context understanding, controlled retrieval and reasoning, and autoregressive multi-step generation. Across these settings, \myarch consistently improves the quality--memory trade-off over representative eviction and quantization baselines. Our primary contributions are:

\begin{itemize}
\item A systematic analysis of decoding-time cache dynamics, including two new diagnostics, \fmm and \lir, together with Selection Churn for measuring routing instability.

\item \textbf{\myarch}, a recoverable window-level KV-cache hierarchy that dynamically routes historical context among full-precision, low-bit, and evicted states, supported by a write-once quantization ledger.

\item Extensive evaluation across three benchmarks and three model families, demonstrating improved quality--memory trade-offs under stringent KV-cache budgets.

\end{itemize}

\section{Related Work}
\label{sec:related_work}

The KV cache grows linearly with sequence length and can dominate the memory cost of long-context autoregressive inference. Existing methods primarily reduce this overhead through eviction, which retains only selected historical states, or quantization, which stores a larger portion of the cache at lower precision. Eviction offers selective retention under strict budgets, whereas quantization provides broader historical coverage. \myarch combines these advantages through a dynamic hierarchy in which low-bit storage serves as a recoverable intermediate state.

\begin{figure*}[!t]
    \centering

    \begin{subfigure}[t]{0.49\textwidth}
        \centering
        \includegraphics[
            width=\linewidth,
            height=0.225\textheight,
            keepaspectratio
        ]{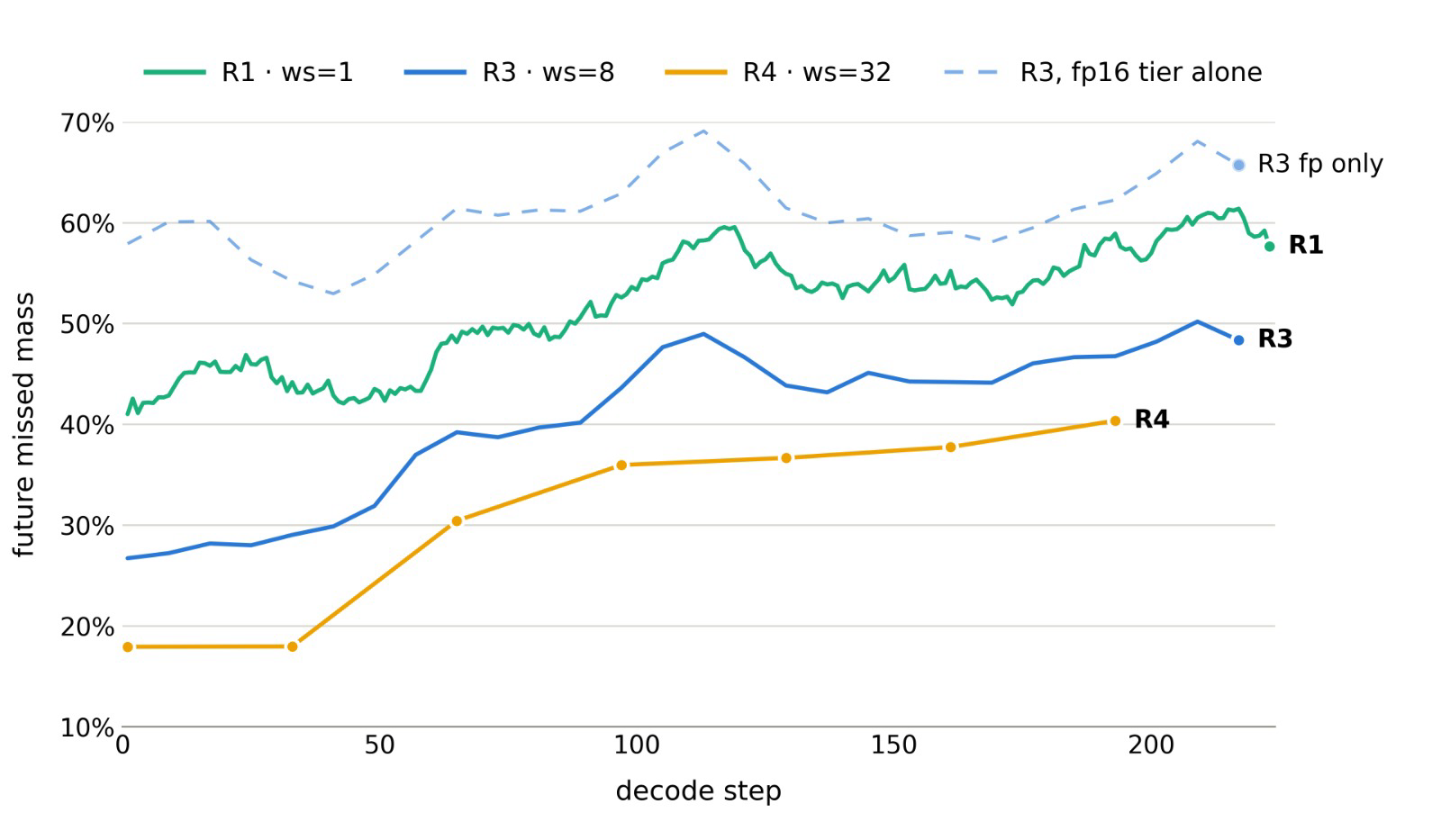}
        \caption{\textbf{FMM over decoding.} Window policies reduce attention assigned to discarded states. The dashed curve treats the R3 tier as inaccessible, isolating the contribution of recoverable low-bit windows.}
        \label{fig:obs1}
    \end{subfigure}
    \hfill
    \begin{subfigure}[t]{0.49\textwidth}
        \centering
        \includegraphics[
            width=\linewidth,
            height=0.225\textheight,
            keepaspectratio
        ]{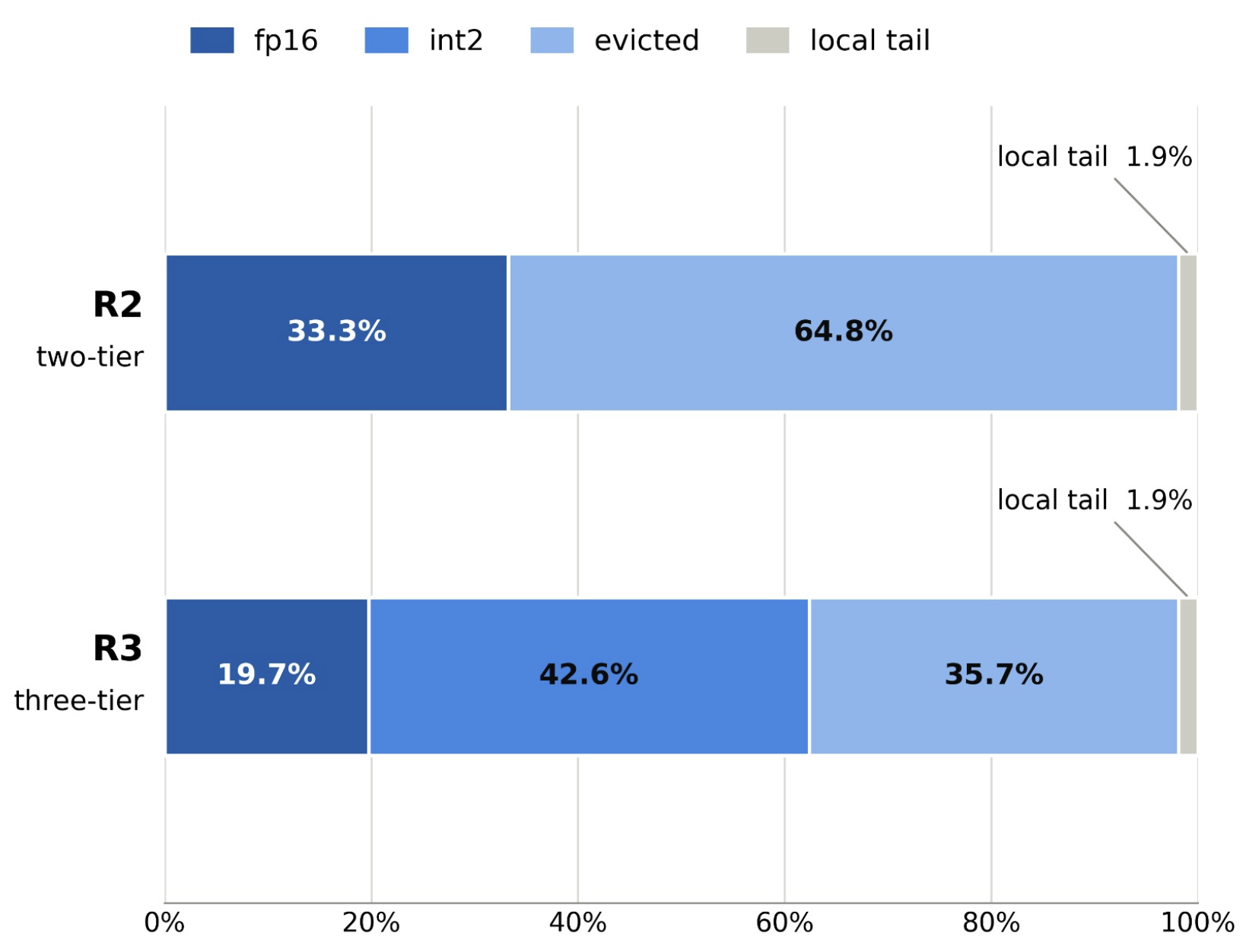}
        \caption{ \textbf{Attention-mass allocation at a matched cache budget.} The three-tier policy preserves a broad intermediate region in INT2 and reduces the mass assigned to permanent eviction. }
        \label{fig:obs2}
    \end{subfigure}

    \caption{
    \textbf{Diagnostics motivating window-level routing and recoverable low-bit retention.} 
    (a) Contiguous-window routing reduces Future Missed Mass, while the gap between the solid and dashed R3 curves shows the future utility of the quantized tier. 
    (b) At the same KV-cache budget, three-tier allocation preserves substantially more ground-truth attention mass than binary full-precision retention and eviction.
    }
    \label{fig:observation_summary}
\end{figure*}

\paragraph{Local Context and Dynamic Relevance.}
Natural language exhibits local lexical, entity, and discourse coherence across neighbouring positions \citep{grosz-etal-1995-centering,hearst-1997-text, passonneau-litman-1997-discourse,barzilay-lapata-2008-modeling, koshorek-etal-2018-text}. Contextual representations and Transformer attention similarly encode dependencies across related spans \citep{ethayarajh-2019-contextual,clark-etal-2019-bert, joshi-etal-2020-spanbert}, motivating windows as a coherent cache-management unit. Relevance is also query-dependent: Quest~\citep{pmlr-v235-tang24l} and SparQ Attention~\citep{pmlr-v235-ribar24a} selectively access cache regions required by the current query. \myarch complements these approaches by preserving historical states whose future relevance has not yet emerged.

\paragraph{KV-Cache Eviction.} StreamingLLM~\citep{xiao2024streamingllm} retains initial attention sinks and a recent-token window, while H$_2$O~\citep{zhang2023h2o} preserves heavy-hitter tokens using accumulated attention. SnapKV~\citep{li2024snapkv} estimates prompt importance from an observation window, AdaKV \citep{feng2024adakv} adapts cache allocation across attention heads, and CriticalKV~\citep{feng2025criticalkv} incorporates value-related information into cache selection. DefensiveKV and Layer-DefensiveKV \citep{feng2026defensivekv} further account for uncertainty in future attention. Although these methods improve scoring and budget allocation, they retain a binary cache state. A token is either preserved or permanently removed. Consequently, a state discarded during a period of low apparent importance cannot be recovered when its relevance re-emerges. Token-level selection can also fragment related context by assigning different retention decisions to neighbouring states. 

\paragraph{KV-Cache Quantization.} KIVI~\citep{liu2024kivi} applies asymmetric 2-bit quantization with different granularities for keys and values. KVQuant~\citep{hooper2024kvquant} combines pre-RoPE key quantization, non-uniform representations, and outlier handling for sub-4-bit compression. ZipCache~\citep{he2024zipcache} assigns saliency-dependent precision at token granularity. 

 Prior eviction and quantization methods leave two coupled challenges unresolved: token-level decisions can fragment context and fluctuate across routing steps, while static cache states cannot recover information whose importance changes during decoding. \myarch addresses these limitations by routing contiguous windows among full-precision, recoverable low-bit, and evicted states. This design preserves broader historical coverage under a fixed memory budget, stabilizes cache selection through window-level aggregation, and supports promotion and demotion as window importance evolves. Section~\ref{sec:observations} empirically examines the attention concentration, selection instability, and importance revival that motivate this three-tier hierarchy.

\begin{figure*}[t!]
\centering 
\includegraphics[width=0.7\linewidth]{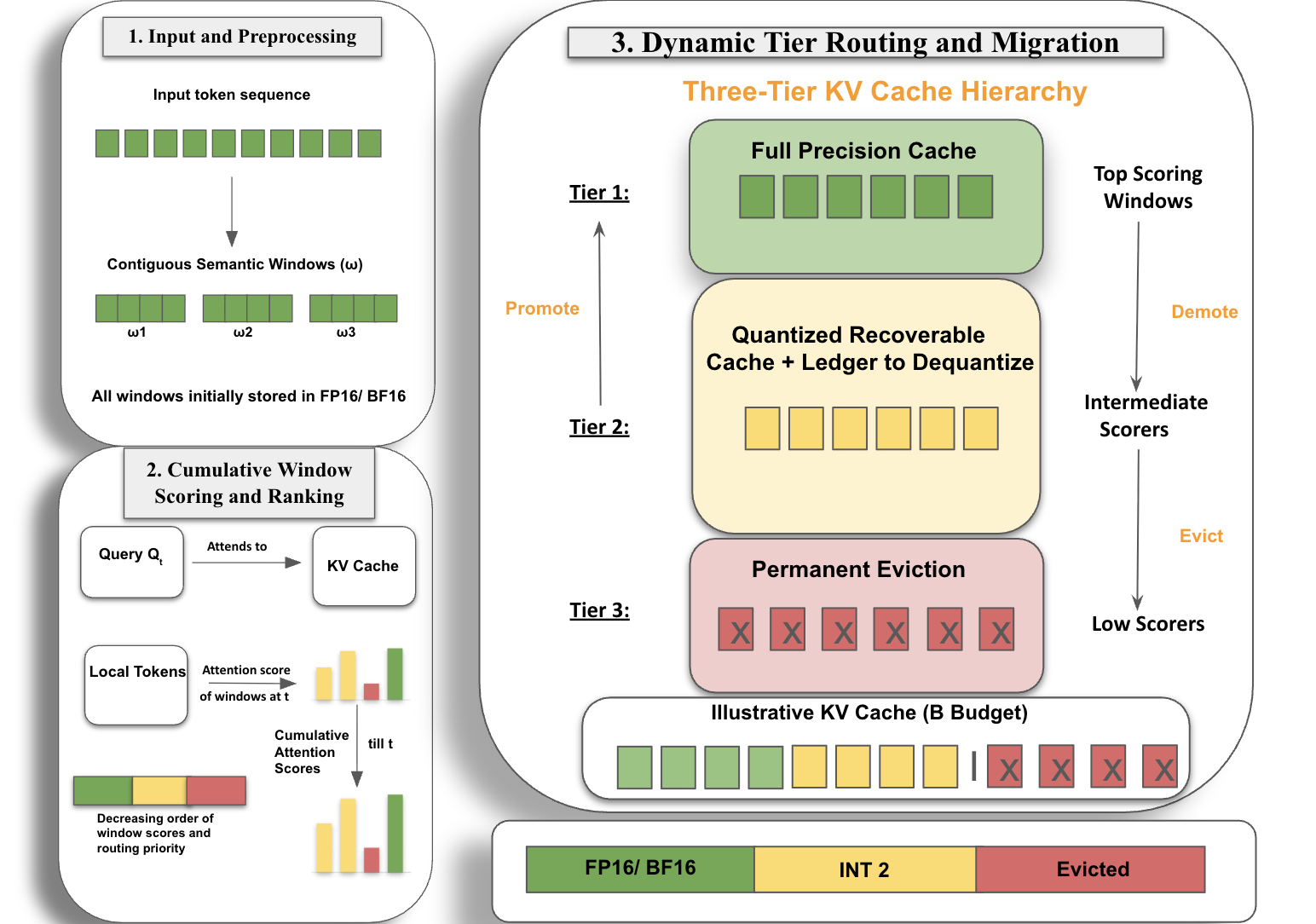} 
\caption{\textbf{Overall workflow of the \myarch framework.} Stage 1 partitions the input sequence into contiguous KV windows while preserving sink and recent tokens in full precision. Stage 2 accumulates attention scores over each window to produce a stable importance ranking. Stage 3 routes windows under a fixed byte budget into a full-precision tier, a recoverable INT2 tier that remains available for attention and later promotion, or permanent eviction, with dynamic promotion and demotion as importance evolves.} 
\label{fig:overall_pipeline} 
\end{figure*}

\section{Observations and Motivation for \myarch}
\label{sec:observations}

Conventional KV-cache eviction assumes that token-level importance is sufficiently stable for permanent deletion and that a binary retain-or-evict decision adequately captures the utility of historical states. We examine these assumptions using attention traces from an uncompressed \fullkv reference, which exposes the future attention assigned to states already removed by a compressed policy.

\paragraph{Diagnostic setup.}
We study Llama-3.1-8B-Instruct with a \(512\)-token prefill, \(256\) generated tokens, a \(20\%\) KV-cache budget, and five protected sink tokens. R1 denotes token-level eviction with window size \(\Omega=1\), R2 denotes two-tier window eviction with \(\Omega=8\), and R3 and R4 denote three-tier policies with \(\Omega=8\) and \(\Omega=32\), respectively. R2 and R3 use the same window size and measured cache budget, providing a controlled comparison between binary eviction and recoverable low-bit retention. Complete experimental and metric definitions are provided in Appendix~\ref{app:observation_details}.

\paragraph{Observation I: Windows stabilize cache decisions.}
We introduce \fmm (FMM), which measures the fraction of future \fullkv attention assigned to states already discarded by a policy, and use \emph{Selection Churn}, the Jaccard distance between historical sets retained at consecutive routing events. As shown in Figure~\ref{fig:obs1}, the window-based policies incur substantially lower FMM than token-level eviction throughout decoding. Historical-set churn per routing event decreases from \(0.017\) for R1 to \(0.0012\) for R3 and \(0.018\) for a larger window; R4. Window aggregation therefore suppresses short-lived selection changes, although larger windows introduce a coarser allocation trade-off.

\paragraph{Observation II: Binary eviction discards a useful middle region.}
Figure~\ref{fig:obs2} compares R2 and R3 at the same window size and byte budget. R2 retains \(33.3\%\) of the \fullkv attention mass in full precision and evicts \(64.8\%\). R3 retains \(19.7\%\) in full precision and preserves another \(42.6\%\) in INT2, reducing the evicted share to \(35.7\%\). The INT2 scores maintain a cosine agreement of \(0.9824\) with the corresponding \fullkv scores. Moreover, disabling access to the R3 quantized tier produces the FMM curve in Figure~\ref{fig:obs1}, showing that these windows remain relevant to subsequent decoding. The ranked cache therefore admits three operational actions: retain in full precision, preserve compactly, or evict.

\paragraph{Observation III: Importance is persistent but not static.}
We further introduce \lir, which measures how often a window re-enters the full-precision set after sustained inactivity. For R3, oracle top-ranked windows exhibit a \(0.98\%\) revival rate, confirming that the dominant region is largely persistent. The deployed full-precision policy exhibits a higher \(6.18\%\) revival rate, while lagged transitions show that demotion is substantially more frequent than promotion. Thus, most important windows remain stable, but a non-negligible subset leaves and later re-enters the active set.

Together, these motivate three design choices: contiguous windows as the routing unit, a recoverable low-bit tier between full-precision retention and eviction, and bidirectional migration as importance evolves. Section~\ref{sec:methodology} formalizes the resulting byte-constrained cache hierarchy.

% ============================================================
% RULER 32K: Combined eviction and quantization comparison
% ============================================================
\begin{table*}[!h]
\centering

\caption{\textbf{RULER performance at 32K context length on Llama-3.1-8B-Instruct.} We compare \myarch with representative KV-cache eviction methods under a matched 20\% cache budget and with quantization baselines at comparable memory footprints. Scores are string-match percentages, and KV memory is reported relative to the full-precision cache. Excluding Full-KV, the best and second-best distinct results within each comparison group are shown in bold and underlined, respectively.}
\label{tab:ruler_32k_combined_llama}

\vspace{-1mm}

\begingroup
\fontsize{5.4}{5.9}\selectfont
\setlength{\tabcolsep}{1.6pt}
\renewcommand{\arraystretch}{0.90}

\resizebox{\textwidth}{!}{%
\begin{tabular}{@{}lVcVccVccccccccVccVc@{}}
\toprule

\multirow{2}{*}{\textbf{Method}} &
\multirow{2}{*}{
  \shortstack{\textbf{KV}\\\textbf{memory}}
} &
\multicolumn{2}{cV}{\textbf{Aggregation}} &
\multicolumn{8}{cV}{\textbf{Needle-in-a-Haystack}} &
\multicolumn{2}{cV}{\textbf{Question Answering}} &
\multicolumn{1}{c}{\textbf{Tracking}} \\

\cmidrule(lr){3-4}
\cmidrule(lr){5-12}
\cmidrule(lr){13-14}
\cmidrule(lr){15-15}

& &
\textbf{CWE} &
\textbf{FWE} &
\textbf{MK-1} &
\textbf{MK-2} &
\textbf{MK-3} &
\textbf{MQ} &
\textbf{MV} &
\textbf{S-1} &
\textbf{S-2} &
\textbf{S-3} &
\textbf{QA-1} &
\textbf{QA-2} &
\textbf{VT} \\

\midrule
\multicolumn{15}{c}{%
  \emph{Llama-3.1-8B-Instruct, RULER 32K}%
} \\
\cmidrule(lr){1-15}

\rowcolor{fullkvgray}
Full-KV
& 100\%
& 45.22
& 94.13
& 99.60
& 99.60
& 99.40
& 98.75
& 99.10
& 100.00
& 100.00
& 100.00
& 79.80
& 54.80
& 99.24 \\

% ============================================================
% Eviction comparison
% ============================================================
\midrule
\multicolumn{15}{c}{%
  \textsc{Eviction: 20\% KV-Cache Budget}%
} \\
\cmidrule(lr){1-15}

StreamingLLM
& 20\%
& 0.04
& \underline{93.40}
& 23.20
& 19.20
& 23.00
& 20.90
& 20.10
& \underline{22.20}
& 17.60
& 20.80
& 23.40
& 39.00
& 30.76 \\

% H$_2$O
% & 20\%
% & \underline{31.36}
% & 79.73
% & 99.20
% & 43.20
% & 55.20
% & \underline{99.00}
% & 91.60
% & \textbf{100.00}
% & 96.80
% & 20.00
% & 81.60
% & \underline{51.20}
% & 94.24 \\

SnapKV
& 20\%
& 14.56
& 70.40
& 98.40
& 94.40
& 72.80
& 98.80
& 98.00
& \textbf{100.00}
& 97.60
& 48.00
& \underline{82.40}
& \textbf{52.00}
& 97.92 \\

AdaKV
& 20\%
& 19.20
& 75.20
& 99.40
& 93.60
& 88.00
& \textbf{99.40}
& \textbf{99.40}
& \textbf{100.00}
& 99.20
& 56.00
& 77.60
& \underline{51.20}
& \textbf{99.52} \\

CriticalKV
& 20\%
& 26.80
& 88.80
& 91.60
& 29.40
& 19.40
& 95.00
& 93.60
& \textbf{100.00}
& \underline{99.60}
& 42.40
& 40.80
& 40.20
& 97.76 \\

DefensiveKV
& 20\%
& 22.94
& 90.00
& \textbf{99.80}
& 86.80
& \underline{97.00}
& 98.65
& 97.90
& \textbf{100.00}
& \textbf{100.00}
& 97.40
& 68.80
& 45.80
& 98.76 \\

Layer-Def.\ KV
& 20\%
& 17.86
& 90.80
& \underline{99.60}
& \textbf{99.40}
& \textbf{99.00}
& 98.85
& \underline{98.45}
& \textbf{100.00}
& \textbf{100.00}
& \textbf{100.00}
& 73.00
& 47.60
& 98.56 \\

\rowcolor{qevictyellow}
\textbf{\myarch}
& 20\%
& \textbf{43.98}
& \textbf{93.60}
& \textbf{99.80}
& \underline{97.00}
& 77.80
& 97.45
& 95.90
& \textbf{100.00}
& \textbf{100.00}
& \underline{98.80}
& \textbf{84.60}
& \underline{51.20}
& \underline{98.84} \\

% ============================================================
% Quantization comparison
% ============================================================
\midrule
\multicolumn{15}{c}{\textsc{Quantization}} \\
\cmidrule(lr){1-15}

\rowcolor{qevictyellow}
\textbf{\myarch}
& 20\%
& 43.98
& \textbf{93.60}
& \textbf{99.80}
& \textbf{97.00}
& 77.80
& \underline{97.45}
& \underline{95.90}
& \textbf{100.00}
& \textbf{100.00}
& \textbf{98.80}
& 84.60
& 51.20
& \textbf{98.84} \\

KIVI-2b
& $\sim$20\%
& 20.76
& 91.20
& \underline{98.00}
& \underline{90.20}
& 21.20
& 92.40
& \textbf{96.35}
& \underline{99.80}
& \underline{98.40}
& \underline{90.40}
& 75.60
& 50.20
& 81.68 \\

ZipCache-4b (70\%)
& $\sim$22\%
& 44.80
& \underline{93.53}
& \textbf{99.80}
& 84.20
& 6.60
& \textbf{97.60}
& 95.25
& \textbf{100.00}
& \textbf{100.00}
& 80.00
& 76.80
& 50.40
& \underline{97.60} \\

KVQuant-2b (s1\%)
& $\sim$15\%
& \underline{64.27}
& 68.21
& 61.05
& 52.55
& \underline{80.88}
& 76.06
& 75.30
& 61.82
& 62.98
& 89.99
& \textbf{98.00}
& \underline{95.00}
& 65.86 \\

KVQuant-3b (s1\%)
& $\sim$22\%
& \textbf{64.38}
& 70.49
& 63.51
& 63.15
& \textbf{89.83}
& 75.56
& 75.46
& 63.64
& 63.64
& 90.00
& \underline{97.80}
& \textbf{96.20}
& 70.07 \\

\bottomrule
\end{tabular}%
}

\endgroup
\end{table*}

\begin{figure*}[t]
    \centering

    \begin{subfigure}[t]{0.32\textwidth}
        \centering
        \includegraphics[
            width=\linewidth,
            height=0.22\textheight,
            keepaspectratio
        ]{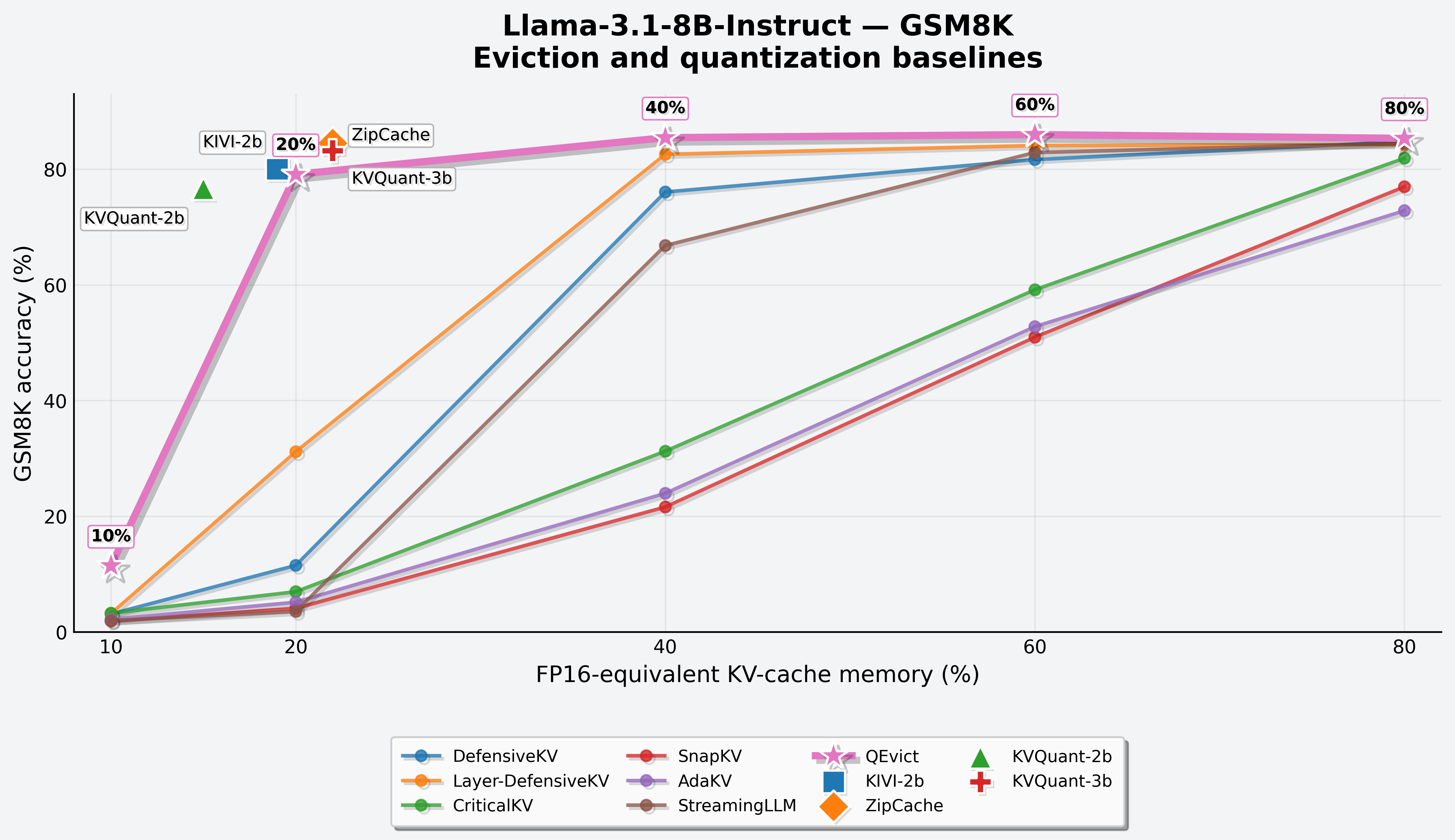}
        \caption{Llama-3.1-8B-Instruct.}
        \label{fig:gsm8k_llama}
    \end{subfigure}
    \hfill
    \begin{subfigure}[t]{0.32\textwidth}
        \centering
        \includegraphics[
            width=\linewidth,
            height=0.22\textheight,
            keepaspectratio
        ]{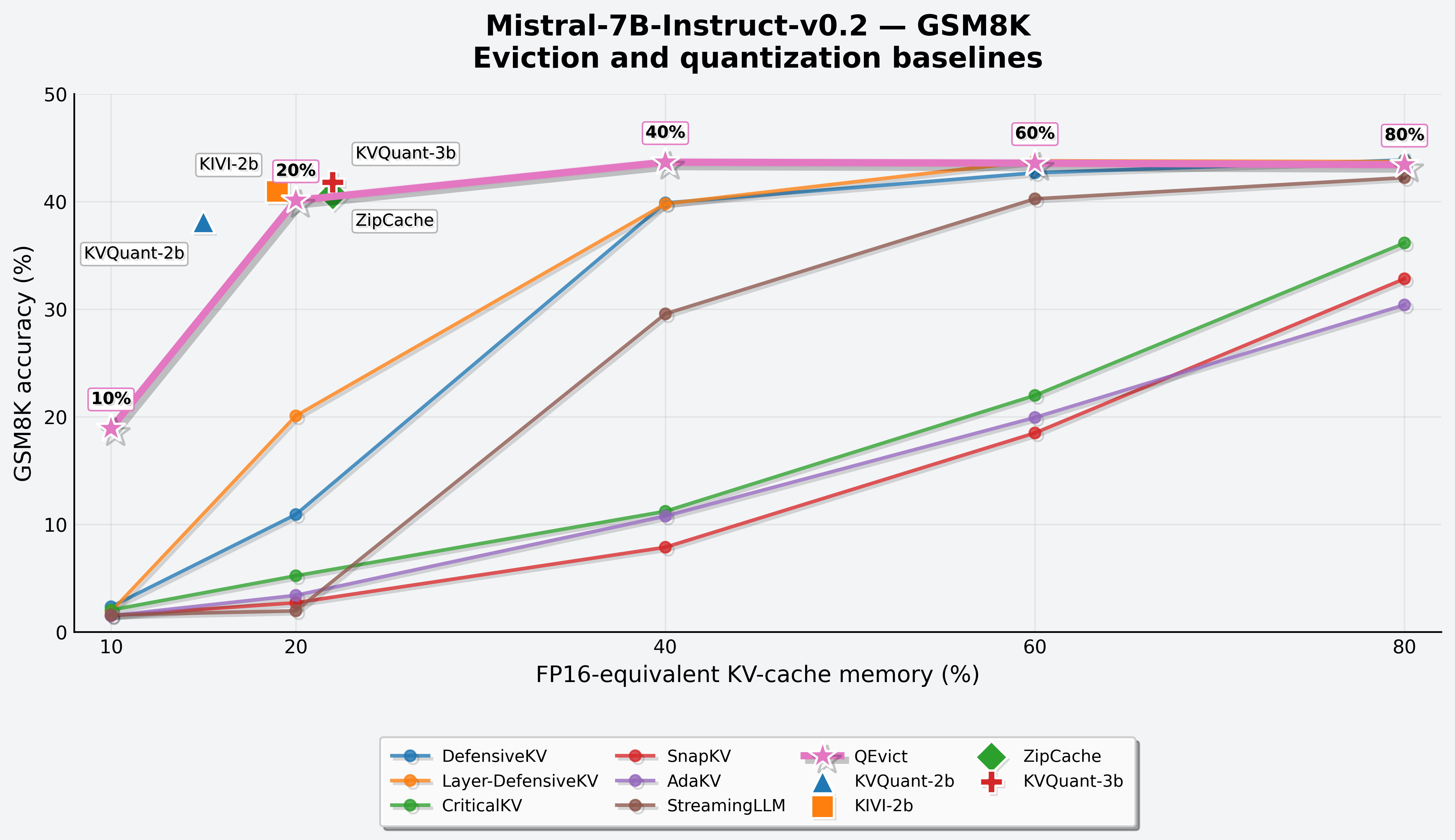}
        \caption{Mistral-7B-Instruct-v0.2.}
        \label{fig:gsm8k_mistral}
    \end{subfigure}
    \hfill
    \begin{subfigure}[t]{0.32\textwidth}
        \centering
        \includegraphics[
            width=\linewidth,
            height=0.22\textheight,
            keepaspectratio
        ]{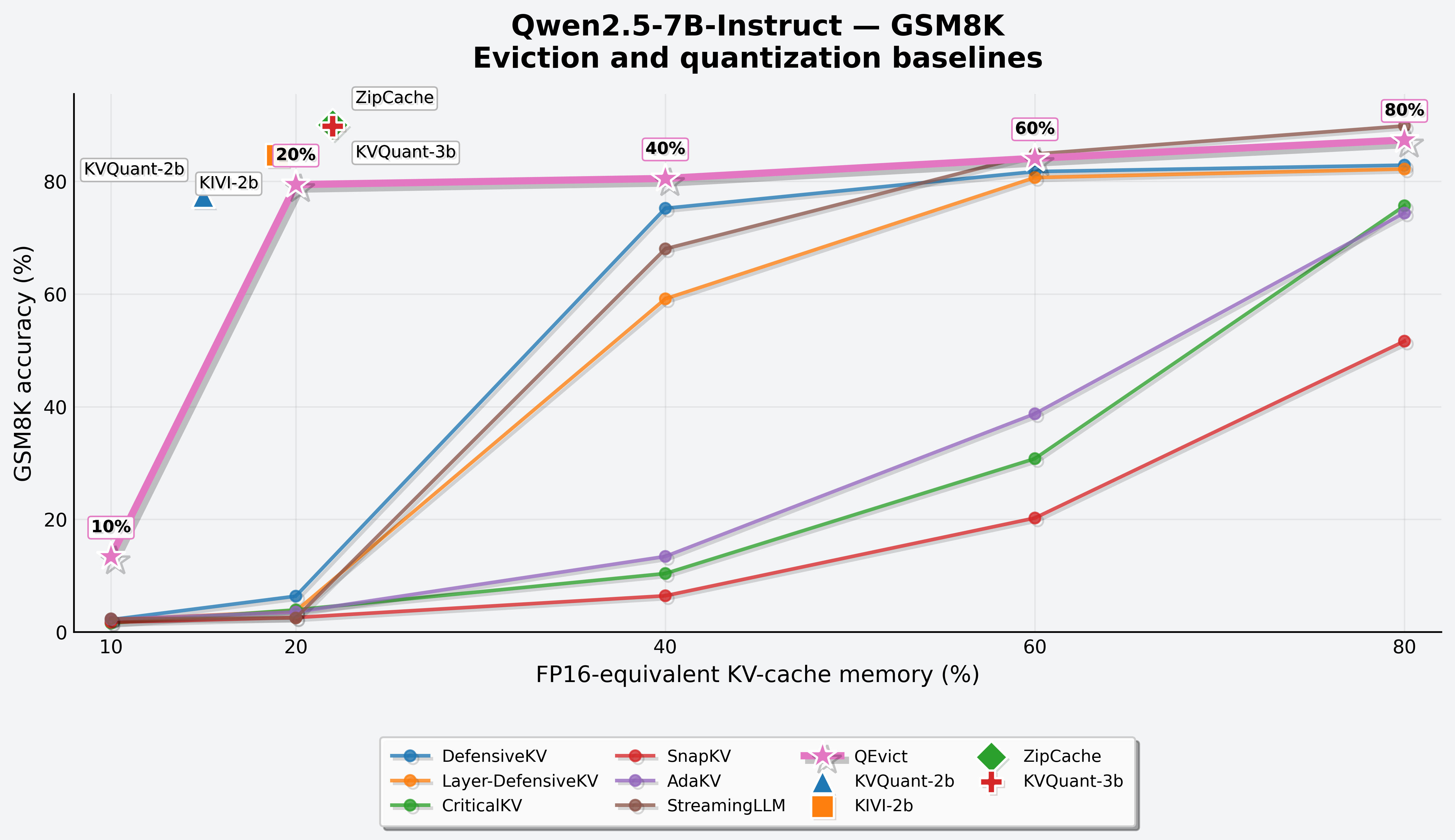}
        \caption{Qwen2.5-7B-Instruct.}
        \label{fig:gsm8k_qwen}
    \end{subfigure}

    \caption{\textbf{GSM8K accuracy under KV-cache compression.}
    Accuracy--memory trade-offs of \myarch and representative eviction and quantization baselines across three instruction-tuned language models.}
    \label{fig:gsm8k_budget_sweep}
\end{figure*}

% ============================================================
% LongBench: combined eviction and quantization comparison
%
% The existing \pair macro is used for each
% Llama-3.1-8B-Instruct / Mistral-7B-Instruct-v0.2 score pair.
% These helpers style either model independently.
% ============================================================
\providecommand{\lbbest}[1]{\textbf{#1}}
\providecommand{\lbsecond}[1]{\underline{#1}}

\begin{table*}[!h]
\centering

\caption{\textbf{LongBench performance on Llama-3.1-8B-Instruct and Mistral-7B-Instruct-v0.2.} Each cell reports the Llama score followed by the Mistral score. We compare \myarch with representative KV-cache eviction methods under matched \(20\%\), \(10\%\), and \(5\%\) budgets, and with quantization baselines at comparable memory footprints. KV memory is reported relative to the full-precision cache. Excluding Full-KV, the best and second-best distinct results for each model within every comparison group are shown in bold and underlined, respectively.}

\label{tab:longbench_combined_two_models}

\vspace{-1mm}

\begingroup
\fontsize{5.4}{5.9}\selectfont
\setlength{\tabcolsep}{1.6pt}
\renewcommand{\arraystretch}{0.90}

\resizebox{\textwidth}{!}{%
\begin{tabular}{@{}lVcVcccVcccVcccVccc@{}}
\toprule

\multirow{2}{*}{\textbf{Method}} &
\multirow{2}{*}{
  \shortstack{\textbf{KV}\\\textbf{memory}}
} &
\multicolumn{3}{cV}{\textbf{Single-Doc QA}} &
\multicolumn{3}{cV}{\textbf{Multi-Doc QA}} &
\multicolumn{3}{cV}{\textbf{Summarization}} &
\multicolumn{3}{c}{\textbf{Few-Shot Learning}} \\

\cmidrule(lr){3-5}
\cmidrule(lr){6-8}
\cmidrule(lr){9-11}
\cmidrule(lr){12-14}

& &
\textbf{Nar.QA} &
\textbf{Qasper} &
\textbf{Mul.QA} &
\textbf{Hot.QA} &
\textbf{2Wi.QA} &
\textbf{Musique} &
\textbf{Gov.Re.} &
\textbf{QMSum} &
\textbf{M.News} &
\textbf{TREC} &
\textbf{Tri.QA} &
\textbf{SAMSum} \\

\midrule
\multicolumn{14}{c}{
  \emph{Llama-3.1-8B-Instruct / Mistral-7B-Instruct-v0.2}
} \\
\cmidrule(lr){1-14}

\rowcolor{fullkvgray}
Full-KV
& 100\%
& \pair{30.5}{21.0}
& \pair{45.5}{29.4}
& \pair{55.0}{47.1}
& \pair{56.0}{36.5}
& \pair{45.7}{21.8}
& \pair{31.3}{19.1}
& \pair{35.1}{32.6}
& \pair{25.6}{24.0}
& \pair{27.3}{27.1}
& \pair{73.0}{71.0}
& \pair{91.7}{86.2}
& \pair{43.7}{43.0} \\

% ============================================================
% Eviction: 20% budget
% ============================================================
\midrule
\multicolumn{14}{c}{
  \textsc{Eviction: 20\% KV-Cache Budget}
} \\
\cmidrule(lr){1-14}

StreamingLLM
& 20\%
& \pair{19.9}{15.3}
& \pair{19.6}{13.3}
& \pair{24.4}{23.9}
& \pair{40.5}{27.8}
& \pair{20.3}{14.7}
& \pair{15.1}{11.6}
& \pair{27.8}{27.4}
& \pair{20.3}{20.1}
& \pair{22.3}{21.9}
& \pair{53.5}{43.0}
& \pair{89.7}{76.9}
& \pair{40.0}{40.2} \\

% H$_2$O
% & 20\%
% & \pair{28.5}{xx.x}
% & \pair{33.4}{xx.x}
% & \pair{39.1}{xx.x}
% & \pair{47.9}{xx.x}
% & \pair{35.2}{xx.x}
% & \pair{\lbsecond{30.0}}{xx.x}
% & \pair{\lbsecond{31.2}}{xx.x}
% & \pair{23.1}{xx.x}
% & \pair{\lbsecond{25.0}}{xx.x}
% & \pair{63.0}{xx.x}
% & \pair{\lbbest{92.3}}{xx.x}
% & \pair{42.5}{xx.x} \\

SnapKV
& 20\%
& \pair{20.9}{16.7}
& \pair{28.3}{17.1}
& \pair{29.4}{31.1}
& \pair{45.7}{30.1}
& \pair{30.0}{16.8}
& \pair{22.8}{11.5}
& \pair{27.3}{26.2}
& \pair{21.6}{21.4}
& \pair{22.6}{22.8}
& \pair{50.0}{51.0}
& \pair{\lbsecond{92.2}}{\lbbest{87.2}}
& \pair{\lbbest{44.6}}{42.3} \\

AdaKV
& 20\%
& \pair{21.7}{17.1}
& \pair{28.4}{17.6}
& \pair{33.5}{32.3}
& \pair{50.8}{30.8}
& \pair{30.4}{17.2}
& \pair{21.7}{13.4}
& \pair{26.4}{25.8}
& \pair{21.6}{21.1}
& \pair{22.9}{22.9}
& \pair{55.0}{53.0}
& \pair{91.4}{\lbsecond{87.0}}
& \pair{43.5}{43.2} \\

CriticalKV
& 20\%
& \pair{\lbbest{29.9}}{19.6}
& \pair{31.1}{19.5}
& \pair{33.2}{33.8}
& \pair{51.3}{30.7}
& \pair{34.3}{18.2}
& \pair{24.9}{13.4}
& \pair{28.5}{27.6}
& \pair{22.9}{21.7}
& \pair{23.1}{23.3}
& \pair{55.0}{57.5}
& \pair{91.8}{\lbbest{87.2}}
& \pair{\lbsecond{44.1}}{41.8} \\

DefensiveKV
& 20\%
& \pair{27.1}{19.2}
& \pair{40.8}{23.1}
& \pair{47.8}{43.7}
& \pair{\lbsecond{55.3}}{34.3}
& \pair{38.7}{\lbsecond{21.7}}
& \pair{27.9}{17.3}
& \pair{29.2}{28.6}
& \pair{23.7}{21.7}
& \pair{23.9}{23.7}
& \pair{66.0}{\lbsecond{67.0}}
& \pair{91.8}{86.7}
& \pair{43.7}{\lbbest{43.8}} \\

Layer-Def.\ KV
& 20\%
& \pair{27.7}{\lbsecond{20.5}}
& \pair{\lbbest{44.3}}{\lbsecond{26.0}}
& \pair{\lbsecond{50.7}}{\lbbest{47.0}}
& \pair{53.2}{\lbsecond{35.3}}
& \pair{\lbsecond{40.7}}{\lbbest{23.3}}
& \pair{27.5}{\lbsecond{18.0}}
& \pair{30.6}{\lbsecond{30.3}}
& \pair{\lbsecond{24.1}}{\lbsecond{22.7}}
& \pair{24.6}{\lbsecond{24.7}}
& \pair{\lbsecond{69.0}}{\lbbest{71.0}}
& \pair{91.6}{86.6}
& \pair{43.7}{\lbsecond{43.5}} \\

\rowcolor{qevictyellow}
\textbf{\myarch}
& 20\%
& \pair{\lbsecond{29.69}}{\lbbest{22.6}}
& \pair{\lbsecond{42.3}}{\lbbest{27.7}}
& \pair{\lbbest{55.74}}{\lbsecond{46.9}}
& \pair{\lbbest{57.97}}{\lbbest{38.0}}
& \pair{\lbbest{48.05}}{20.8}
& \pair{\lbbest{32.99}}{\lbbest{18.3}}
& \pair{\lbbest{33.71}}{\lbbest{30.8}}
& \pair{\lbbest{25.11}}{\lbbest{24.1}}
& \pair{\lbbest{26.89}}{\lbbest{26.8}}
& \pair{\lbbest{71.0}}{\lbbest{71.0}}
& \pair{91.21}{85.5}
& \pair{41.7}{41.1} \\

% ============================================================
% Eviction: 10% budget
% ============================================================
\midrule
\multicolumn{14}{c}{
  \textsc{Eviction: 10\% KV-Cache Budget}
} \\
\cmidrule(lr){1-14}

StreamingLLM
& 10\%
& \pair{18.2}{15.8}
& \pair{16.6}{10.8}
& \pair{22.8}{22.6}
& \pair{34.7}{23.1}
& \pair{16.6}{14.3}
& \pair{11.6}{10.3}
& \pair{24.7}{24.8}
& \pair{18.8}{19.2}
& \pair{19.8}{18.8}
& \pair{51.0}{32.0}
& \pair{88.8}{72.8}
& \pair{38.3}{38.6} \\

% H$_2$O
% & 10\%
% & \pair{\lbsecond{26.9}}{xx.x}
% & \pair{\lbsecond{32.0}}{xx.x}
% & \pair{\lbbest{49.8}}{xx.x}
% & \pair{\lbsecond{53.4}}{xx.x}
% & \pair{\lbsecond{44.4}}{xx.x}
% & \pair{\lbsecond{29.2}}{xx.x}
% & \pair{\lbsecond{27.4}}{xx.x}
% & \pair{\lbsecond{23.9}}{xx.x}
% & \pair{\lbsecond{22.1}}{xx.x}
% & \pair{43.5}{xx.x}
% & \pair{91.5}{xx.x}
% & \pair{42.3}{xx.x} \\

SnapKV
& 10\%
& \pair{18.9}{14.9}
& \pair{20.8}{12.0}
& \pair{23.6}{25.2}
& \pair{43.0}{26.4}
& \pair{22.6}{13.5}
& \pair{18.2}{10.8}
& \pair{24.1}{24.3}
& \pair{19.6}{20.2}
& \pair{20.4}{20.9}
& \pair{44.5}{44.5}
& \pair{\lbbest{92.9}}{87.3}
& \pair{43.0}{40.7} \\

AdaKV
& 10\%
& \pair{18.3}{16.3}
& \pair{22.7}{13.7}
& \pair{26.1}{27.2}
& \pair{40.8}{29.1}
& \pair{22.2}{15.4}
& \pair{17.8}{11.9}
& \pair{24.3}{24.3}
& \pair{20.2}{19.9}
& \pair{20.9}{20.6}
& \pair{46.5}{48.5}
& \pair{92.2}{87.1}
& \pair{43.3}{42.2} \\

CriticalKV
& 10\%
& \pair{25.1}{16.5}
& \pair{23.4}{13.9}
& \pair{26.1}{26.4}
& \pair{40.3}{27.6}
& \pair{25.6}{15.6}
& \pair{19.7}{10.5}
& \pair{25.3}{25.4}
& \pair{21.0}{20.6}
& \pair{20.9}{21.5}
& \pair{45.5}{45.5}
& \pair{\lbsecond{92.5}}{\lbbest{87.6}}
& \pair{42.5}{42.0} \\

DefensiveKV
& 10\%
& \pair{20.9}{17.0}
& \pair{26.0}{14.8}
& \pair{34.0}{32.6}
& \pair{48.3}{31.7}
& \pair{30.0}{16.7}
& \pair{19.6}{11.6}
& \pair{25.1}{25.1}
& \pair{22.4}{20.8}
& \pair{21.5}{21.6}
& \pair{\lbsecond{54.0}}{53.0}
& \pair{91.4}{\lbsecond{87.4}}
& \pair{\lbsecond{43.7}}{\lbsecond{43.1}} \\

Layer-Def.\ KV
& 10\%
& \pair{22.6}{\lbsecond{19.2}}
& \pair{29.1}{\lbsecond{16.7}}
& \pair{\lbsecond{37.5}}{\lbsecond{38.6}}
& \pair{49.6}{\lbsecond{35.6}}
& \pair{30.2}{\lbbest{20.6}}
& \pair{21.7}{\lbsecond{14.7}}
& \pair{25.6}{\lbsecond{26.8}}
& \pair{22.6}{\lbsecond{21.4}}
& \pair{21.4}{\lbsecond{21.9}}
& \pair{52.0}{\lbsecond{64.0}}
& \pair{91.5}{\lbsecond{87.4}}
& \pair{\lbbest{44.0}}{\lbbest{43.4}} \\

\rowcolor{qevictyellow}
\textbf{\myarch}
& 10\%
& \pair{\lbbest{28.8}}{\lbbest{20.1}}
& \pair{\lbbest{34.94}}{\lbbest{23.3}}
& \pair{\lbbest{49.8}}{\lbbest{42.6}}
& \pair{\lbbest{58.4}}{\lbbest{36.5}}
& \pair{\lbbest{46.4}}{\lbsecond{18.7}}
& \pair{\lbbest{32.9}}{\lbbest{18.0}}
& \pair{\lbbest{32.1}}{\lbbest{29.6}}
& \pair{\lbbest{24.9}}{\lbbest{24.0}}
& \pair{\lbbest{24.6}}{\lbbest{25.1}}
& \pair{\lbbest{67.5}}{\lbbest{69.0}}
& \pair{89.8}{83.4}
& \pair{40.4}{39.5} \\

% ============================================================
% Eviction: 5% budget
% ============================================================
\midrule
\multicolumn{14}{c}{
  \textsc{Eviction: 5\% KV-Cache Budget}
} \\
\cmidrule(lr){1-14}

StreamingLLM
& 5\%
& \pair{14.3}{13.1}
& \pair{13.9}{9.4}
& \pair{19.7}{20.3}
& \pair{30.1}{18.5}
& \pair{14.7}{12.8}
& \pair{7.9}{8.2}
& \pair{22.2}{21.8}
& \pair{17.6}{18.8}
& \pair{17.3}{15.9}
& \pair{38.5}{19.5}
& \pair{87.8}{68.7}
& \pair{36.2}{37.5} \\

% H$_2$O
% & 5\%
% & \pair{\lbsecond{24.8}}{xx.x}
% & \pair{\lbbest{30.1}}{xx.x}
% & \pair{\lbbest{47.5}}{xx.x}
% & \pair{\lbsecond{52.4}}{xx.x}
% & \pair{\lbsecond{42.3}}{xx.x}
% & \pair{\lbsecond{28.4}}{xx.x}
% & \pair{\lbsecond{25.9}}{xx.x}
% & \pair{\lbsecond{23.7}}{xx.x}
% & \pair{\lbbest{20.9}}{xx.x}
% & \pair{40.5}{xx.x}
% & \pair{90.8}{xx.x}
% & \pair{41.3}{xx.x} \\

SnapKV
& 5\%
& \pair{18.3}{14.4}
& \pair{15.7}{11.4}
& \pair{19.8}{21.8}
& \pair{36.3}{24.0}
& \pair{13.4}{14.7}
& \pair{13.2}{8.6}
& \pair{21.6}{21.8}
& \pair{18.3}{19.2}
& \pair{17.8}{17.8}
& \pair{34.0}{38.5}
& \pair{92.0}{86.6}
& \pair{\lbsecond{42.1}}{39.5} \\

AdaKV
& 5\%
& \pair{17.0}{15.1}
& \pair{16.4}{10.6}
& \pair{19.6}{22.9}
& \pair{36.7}{24.8}
& \pair{17.1}{14.5}
& \pair{14.4}{9.0}
& \pair{21.7}{21.6}
& \pair{18.7}{19.4}
& \pair{18.4}{18.2}
& \pair{35.5}{41.0}
& \pair{\lbsecond{92.8}}{\lbsecond{87.4}}
& \pair{42.0}{39.8} \\

CriticalKV
& 5\%
& \pair{20.2}{14.7}
& \pair{17.4}{11.0}
& \pair{19.6}{22.9}
& \pair{36.0}{25.8}
& \pair{15.5}{14.3}
& \pair{15.8}{9.2}
& \pair{22.7}{23.0}
& \pair{18.7}{19.4}
& \pair{18.2}{18.2}
& \pair{34.0}{40.0}
& \pair{92.5}{85.7}
& \pair{42.0}{\lbsecond{40.9}} \\

DefensiveKV
& 5\%
& \pair{20.8}{\lbsecond{16.1}}
& \pair{18.1}{11.1}
& \pair{21.6}{24.3}
& \pair{39.8}{25.7}
& \pair{19.8}{\lbsecond{15.4}}
& \pair{14.2}{\lbsecond{10.5}}
& \pair{23.2}{23.1}
& \pair{19.5}{20.0}
& \pair{18.9}{19.0}
& \pair{\lbsecond{43.5}}{44.0}
& \pair{\lbbest{93.0}}{87.1}
& \pair{\lbbest{43.2}}{\lbsecond{40.9}} \\

Layer-Def.\ KV
& 5\%
& \pair{19.8}{15.6}
& \pair{20.1}{\lbsecond{11.8}}
& \pair{21.8}{\lbsecond{27.0}}
& \pair{40.9}{\lbsecond{28.0}}
& \pair{19.8}{14.9}
& \pair{17.8}{10.3}
& \pair{22.9}{\lbsecond{23.9}}
& \pair{19.7}{\lbsecond{20.4}}
& \pair{18.9}{\lbsecond{19.2}}
& \pair{41.5}{\lbsecond{47.0}}
& \pair{\lbbest{93.0}}{\lbbest{87.6}}
& \pair{\lbbest{43.2}}{\lbbest{42.6}} \\

\rowcolor{qevictyellow}
\textbf{\myarch}
& 5\%
& \pair{\lbbest{27.3}}{\lbbest{20.0}}
& \pair{\lbsecond{28.0}}{\lbbest{16.8}}
& \pair{\lbsecond{43.5}}{\lbbest{34.6}}
& \pair{\lbbest{56.4}}{\lbbest{34.1}}
& \pair{\lbbest{45.1}}{\lbbest{17.1}}
& \pair{\lbbest{32.5}}{\lbbest{25.1}}
& \pair{\lbbest{28.8}}{\lbbest{27.6}}
& \pair{\lbbest{23.9}}{\lbbest{22.7}}
& \pair{\lbsecond{19.3}}{\lbbest{19.9}}
& \pair{\lbbest{63.0}}{\lbbest{63.5}}
& \pair{89.6}{84.6}
& \pair{38.2}{38.2} \\

% ============================================================
% Quantization comparison
% ============================================================
\midrule
\multicolumn{14}{c}{\textsc{Quantization}} \\
\cmidrule(lr){1-14}

\rowcolor{qevictyellow}
\textbf{\myarch}
& 10\%
& \pair{\lbsecond{28.8}}{20.1}
& \pair{34.94}{23.3}
& \pair{49.8}{42.6}
& \pair{\lbbest{58.4}}{36.5}
& \pair{46.4}{18.7}
& \pair{\lbsecond{32.9}}{\lbsecond{18.0}}
& \pair{32.1}{29.6}
& \pair{24.9}{\lbsecond{24.0}}
& \pair{24.6}{25.1}
& \pair{67.5}{\lbsecond{69.0}}
& \pair{89.8}{83.4}
& \pair{40.4}{39.5} \\

\rowcolor{qevictyellow}
\textbf{\myarch}
& 20\%
& \pair{\lbbest{29.69}}{\lbbest{22.6}}
& \pair{42.3}{27.7}
& \pair{\lbbest{55.74}}{\lbsecond{46.9}}
& \pair{\lbsecond{57.97}}{\lbbest{38.0}}
& \pair{\lbsecond{48.05}}{\lbsecond{20.8}}
& \pair{\lbbest{32.99}}{\lbbest{18.3}}
& \pair{\lbbest{33.71}}{\lbsecond{30.8}}
& \pair{\lbbest{25.11}}{\lbbest{24.1}}
& \pair{\lbbest{26.89}}{\lbbest{26.8}}
& \pair{\lbsecond{71.0}}{\lbbest{71.0}}
& \pair{\lbsecond{91.21}}{\lbsecond{85.5}}
& \pair{41.7}{\lbsecond{41.1}} \\

KIVI-2b
& $\sim$20\%
& \pair{23.7}{\lbsecond{20.6}}
& \pair{36.8}{\lbbest{28.7}}
& \pair{41.1}{44.9}
& \pair{44.4}{35.5}
& \pair{30.0}{20.7}
& \pair{21.9}{\lbsecond{18.0}}
& \pair{30.5}{\lbbest{32.6}}
& \pair{24.4}{23.7}
& \pair{26.0}{26.5}
& \pair{68.0}{\lbbest{71.0}}
& \pair{87.8}{\lbbest{86.0}}
& \pair{\lbbest{44.7}}{\lbbest{43.3}} \\

ZipCache-4b (70\%)
& $\sim$22\%
& \pair{24.1}{20.4}
& \pair{\lbsecond{43.3}}{28.1}
& \pair{\lbsecond{53.9}}{45.4}
& \pair{47.4}{33.2}
& \pair{\lbbest{50.8}}{\lbbest{20.9}}
& \pair{25.2}{16.6}
& \pair{32.6}{29.9}
& \pair{22.9}{22.9}
& \pair{\lbsecond{26.5}}{\lbbest{26.8}}
& \pair{21.0}{39.5}
& \pair{91.0}{80.3}
& \pair{20.4}{37.3} \\

KVQuant-2b (s1\%)
& $\sim$15\%
& \pair{22.6}{16.0}
& \pair{36.5}{23.4}
& \pair{47.1}{41.8}
& \pair{43.8}{29.3}
& \pair{34.2}{17.3}
& \pair{23.2}{11.6}
& \pair{31.5}{27.5}
& \pair{22.9}{22.6}
& \pair{25.8}{26.0}
& \pair{65.0}{10.8}
& \pair{86.3}{68.0}
& \pair{41.9}{25.1} \\

KVQuant-3b (s1\%)
& $\sim$22\%
& \pair{28.6}{20.1}
& \pair{\lbbest{44.3}}{\lbsecond{28.6}}
& \pair{53.4}{\lbbest{47.1}}
& \pair{53.9}{\lbsecond{36.7}}
& \pair{42.7}{19.9}
& \pair{28.0}{17.3}
& \pair{\lbsecond{33.4}}{30.4}
& \pair{\lbsecond{25.0}}{23.7}
& \pair{\lbsecond{26.5}}{\lbsecond{26.7}}
& \pair{\lbbest{72.5}}{10.3}
& \pair{\lbbest{92.0}}{77.1}
& \pair{\lbsecond{44.6}}{37.0} \\

\bottomrule
\end{tabular}%
}

\endgroup
\end{table*}

\section{Methodology}
\label{sec:methodology}

\myarch manages historical KV states as contiguous windows and dynamically assigns each window to full-precision, recoverable low-bit, or evicted state under a fixed budget.

\subsection{Preliminaries and Problem Formulation}
\label{subsec:method_preliminaries}

Consider transformer layer \(\ell\) with \(H_q\) query heads and \(H_{kv}\) KV heads. At decoding step \(t\), query \(q^{\ell,h}_t\) attends over the currently accessible keys and values:
\[
    a^{\ell,h}_t
    =
    \operatorname{softmax}
    \left(
        \frac{
            q^{\ell,h}_t
            \bigl(K^\ell_{\mathrm{acc},t}\bigr)^\top
        }{\sqrt{d_{kv}}}
    \right),
    \,\, o^{\ell,h}_t
    =
    a^{\ell,h}_t V^\ell_{\mathrm{acc},t}.
\]
 
\noindent After reserving a protected sink prefix and recent region, we partition the remaining historical cache into contiguous windows \(\mathcal{W}_t\). For each window \(w\), \myarch assigns a state
\[
    z^\ell_t(w)
    \in
    \left\{
        \textsc{Full},
        \textsc{Quantized},
        \textsc{Evicted}
    \right\}.
\]
Let \(M_f(w)\) and \(M_q(w)\) denote the storage costs of a full-precision and quantized window. Assignment must satisfy
\begin{equation*}
\begin{aligned}
    B_{\mathrm{sink}}
    + B_{\mathrm{local}}
    + \sum_{w\in\mathcal{W}_t}
    \Bigl[
        \mathbf{1}\!\left[z^\ell_t(w)=\textsc{Full}\right]M_f(w)
        +
        \\
        \mathbf{1}\!\left[z^\ell_t(w)=\textsc{Quantized}\right]M_q(w)
    \Bigr]
    \leq B_{\mathrm{total}}.
\end{aligned}
\label{eq:method_budget_constraint}
\end{equation*}

Thus, \myarch jointly determines cache residency and representation
precision rather than selecting a single retained subset.

\subsection{Cumulative Window Scoring}
\label{subsec:method_scoring}

Let \(a^{\ell,h}_{\tau}(i)\) denote the attention probability assigned to token \(i\) by query head \(h\). At routing event \(t\), the cumulative score of window \(w\) is updated as

\vspace{-1 em}

\begin{equation}
\begin{aligned}
\overline{S}^{\ell}_t(w)
&=
\overline{S}^{\ell}_{t-\Omega}(w)
+
\frac{1}{H_q}
\sum_{h=1}^{H_q}
\sum_{\tau=t-\Omega+1}^{t}
\sum_{i\in w}
a^{\ell,h}_{\tau}(i),
\\
&\qquad t \equiv 0 \pmod{R}.
\end{aligned}
\label{eq:cumulative_window_score}
\end{equation} 

\vspace{-1 em}

The resulting scores induce a layer-wise ranking over historical windows. Although we use cumulative attention as the default importance estimator, the hierarchy requires only an ordering and is compatible with alternative ranking functions. More detailed discussion in Appendix~\ref{app:ranking_generality}

\noindent \myarch operates during both prefill and decoding. During prefill, prompt attention initializes the window scores and the first tier assignment. During decoding, newly generated states first enter the protected recent region, and historical windows are rescored and rerouted every \(\Omega\) generated tokens.

\subsection{\myarch: Recoverable Cache Hierarchy}
\label{subsec:method_hierarchy}

At routing event \(t\), the cache at layer \(\ell\) is partitioned
as:
\begin{equation}
    \mathcal{C}^{\ell}_t
    =
    \left(
        \mathcal{S},
        \mathcal{L}_t,
        \mathcal{F}_t,
        \mathcal{Q}_t,
        \mathcal{E}_t
    \right),
    \label{eq:qevict_cache_partition}
\end{equation}
where \(\mathcal{S}\) and \(\mathcal{L}_t\) are the protected sink and recent regions and remain in full precision. Historical windows are assigned to the full-precision tier \(\mathcal{F}_t\), the recoverable low-bit tier \(\mathcal{Q}_t\), or the evicted set \(\mathcal{E}_t\).

\paragraph{Byte-Constrained Tier Allocation.}
After reserving the protected regions, the remaining historical budget \(B_{\mathrm{hist}}\) is divided between the full-precision and quantized tiers. Given quantized-tier fraction \(q\), their capacities are

\begin{equation}
    K_f
    =
    \left\lfloor
        \frac{(1-q)B_{\mathrm{hist}}}{M_f}
    \right\rfloor,
    \qquad
    K_q
    =
    \left\lfloor
        \frac{qB_{\mathrm{hist}}}{M_q}
    \right\rfloor.
    \label{eq:qevict_tier_capacities}
\end{equation}
The per-window costs include packed codes, quantization parameters, positions, and persistent indexing state. Complete byte-level accounting and the default configuration are provided in Appendix~\ref{app:memory_accounting} and Appendix~\ref{app:QEvict_configuration}. Sensitivity to \(q\), \(\Omega\), and quantization precision is studied in Appendix~\ref{app:ablations}.

\paragraph{Migration-Stable Quantization.}
When a window first enters \(\mathcal{Q}_t\), \myarch applies asymmetric low-bit quantization. Following the distinct distributions of keys and values~\citep{liu2024kivi}, keys are quantized per channel across the token dimension, whereas values are quantized per token across the channel dimension. Keys are represented in the pre-RoPE domain together with their original absolute positions, and RoPE is reapplied after dequantization.

The codes and quantization parameters produced on the first demotion are retained and reused during later tier transitions. Promotion reconstructs the same low-bit approximation in the model's execution datatype, while a subsequent demotion reuses the existing representation instead of requantizing the reconstructed values. This prevents approximation error from compounding across repeated migrations.

\paragraph{Dynamic Routing and Recovery.}
At each routing event, the candidate pool contains the current full-precision and quantized windows together with windows that have exited the recent region:

\vspace{-1 em}

\begin{equation}
    \mathcal{W}^{\mathrm{cand}}_t
    =
    \mathcal{F}_{t-\Omega}
    \cup
    \mathcal{Q}_{t-\Omega}
    \cup
    \mathcal{L}^{\mathrm{aged}}_t.
    \label{eq:qevict_candidate_pool}
\end{equation}

% \vspace{-2 em}

Candidates are ranked by \(\overline{S}^{\ell}_t(w)\). Highest-ranked \(K_f\) windows are assigned to \(\mathcal{F}_t\), next \(K_q\) to \(\mathcal{Q}_t\), and rest to \(\mathcal{E}_t\):
\begin{equation*}
\begin{aligned}
\mathcal{F}_t &= \operatorname{TopK}_{K_f} \left( \mathcal{W}^{\mathrm{cand}}_t; \overline{S}^{\ell}_t \right)
\end{aligned}
\label{eq}
\end{equation*}

\begin{equation}
\begin{aligned}
\mathcal{Q}_t &= \operatorname{TopK}_{K_q} \left( \mathcal{W}^{\mathrm{cand}}_t \setminus\mathcal{F}_t; \overline{S}^{\ell}_t \right)
,\\
\mathcal{E}_t &= \mathcal{W}^{\mathrm{cand}}_t \setminus \left( \mathcal{F}_t\cup\mathcal{Q}_t \right).
\end{aligned}
\label{eq}
\end{equation}

\noindent Quantized windows remain available to attention and continue accumulating scores. A window can therefore be promoted when its importance increases or demoted when it declines:
\begin{equation}
    \textsc{Full}
    \rightleftarrows
    \textsc{Quantized}
    \longrightarrow
    \textsc{Evicted}.
    \label{eq:qevict_state_transition}
\end{equation}

Only windows outside the combined capacities \(K_f+K_q\) are permanently removed.

\paragraph{Cache Execution.}
Full-precision and quantized windows are stored separately. Before attention, active low-bit windows are dequantized, their keys are re-rotated at the original positions, and all accessible states are restored in chronological order. Standard decoding uses FlashAttention-2, while SDPA is invoked only at prefill initialization and routing events to expose the attention probabilities required for scoring. Further execution details are provided in Appendix~\ref{app:efficiency}.

\section{Experiments}
\label{sec:experiments}

\paragraph{Setup.}
We evaluate \myarch on Llama-3.1-8B-Instruct~\citep{dubey2024llama3}, Qwen2.5-7B-Instruct~\citep{yang2025qwen3}, and Mistral-7B-Instruct-v0.2~\citep{jiang2023mistral} using LongBench~\citep{bai2024longbench}, RULER~\citep{hsieh2024ruler}, and GSM8K~\citep{cobbe2021training}. We compare against two baseline families: permanent KV-cache eviction methods and global KV-cache quantization methods. Eviction baselines are evaluated under matched measured-memory budgets, while quantization baselines are compared at their corresponding performance--memory operating points.

Unless otherwise stated, \myarch uses routing interval \(\Omega=8\), five sink tokens, \(32\) recent tokens, an INT2 recoverable tier, and quantized-tier fraction \(q=0.70\) across LongBench, RULER, and GSM8K. Complete baseline configurations, decoding protocols, and byte-level memory accounting are provided in Appendix~\ref{app:experiments}.

\paragraph{Long-context understanding.}
We evaluate 12 LongBench tasks across single-document QA, multi-document QA, summarization, and few-shot learning at \(5\%\), \(10\%\), and \(20\%\) KV-memory budgets. Table~\ref{tab:longbench_combined_two_models} reports results for Llama and Mistral, with Qwen results in Appendix~\ref{app:longbench}.

\myarch achieves the highest macro-average across all six model--budget settings. At \(20\%\) memory, it scores \(46.4\) on Llama and \(37.8\) on Mistral, compared with \(44.0\) and \(37.4\) for the strongest matched-memory eviction baseline. The gains increase at \(5\%\) memory to \(9.7\) and \(4.7\) points, respectively, highlighting the value of recoverable low-bit retention under tight budgets. \myarch also achieves the best \(20\%\) macro-average among quantization methods; at only \(10\%\) memory, it remains within \(1.2\) points on Llama and \(1.8\) points on Mistral of the strongest quantization baseline despite using roughly half the KV memory. Further analyses are provided in Appendix~\ref{app:pareto}.

\paragraph{Long-range retrieval.}
We evaluate RULER at a \(32\)K context length and a \(20\%\) KV-cache budget on Llama-3.1-8B-Instruct. Table~\ref{tab:ruler_32k_combined_llama} reports all 13 aggregation, needle-in-a-haystack, question-answering, and tracking tasks. \myarch obtains a macro-average of \(87.6\), exceeding the strongest matched-memory eviction baseline, Layer-DefensiveKV, by \(1.2\) points. It also outperforms the strongest comparable-memory quantization baseline by \(8.6\) points and remains within \(2.4\) points of the uncompressed Full-KV reference. The results show that broad low-bit historical coverage is especially effective for retrieval-intensive long-context tasks.

\paragraph{Reasoning under compression.}
On GSM8K, Figure~\ref{fig:gsm8k_budget_sweep} reports accuracy--memory curves for Llama, Qwen, and Mistral over complete memory sweep. This evaluates whether the recoverable hierarchy preserves multi-step generation quality as the available cache budget decreases. Complete numerical results and model-specific comparisons with eviction and quantization baselines are provided in Appendix~\ref{app:gsm8k}.

\paragraph{Ablations.}
We vary the routing interval \(\Omega\), quantized-tier fraction \(q\), and recoverable precision. The results favour \(\Omega=8\) and \(q=0.70\) across benchmarks. INT2 provides broader historical coverage under a fixed byte budget, while INT4 trades coverage for lower quantization error. Full results are reported in Appendix~\ref{app:ablations}.

\subsection{Efficiency, Limitations, and Future Work}
\label{sec:efficiency_limitations}

\paragraph{End-to-end efficiency.}
Table~\ref{tab:efficiency} evaluates \myarch under matched model, batch, sequence-length, and backend settings. With eager SDPA, \myarch adds only \(0.5\%\) TTFT overhead, reduces TPOT by \(9.3\%\), and improves decoding throughput by \(9.8\%\) over Full-KV. Thus, reducing the full-precision attended cache offsets the cost of low-bit cache management in the eager implementation.

With FlashAttention-2, \myarch reduces peak GPU memory from \(29.54\) to \(20.78\) GB, a \(29.7\%\) reduction. Its current execution path, however, requires periodic attention-score materialization together with dequantization and cache reconstruction, which lowers decoding throughput. The resulting trade-off is backend dependent: \myarch improves eager decoding efficiency and substantially reduces FlashAttention-2 memory, while fused low-bit attention and routing kernels remain necessary to realize both benefits simultaneously.

\begin{table}[t]
\centering
\caption{\textbf{End-to-end inference efficiency on Llama-3.1-8B-Instruct.} Results use a 256-token prefill, 1024 generated tokens, batch size 32, and the default \myarch configuration. Comparisons are made within each attention backend.}
\label{tab:efficiency}

\begingroup
\small
\setlength{\tabcolsep}{3.8pt}
\renewcommand{\arraystretch}{0.94}

\resizebox{\columnwidth}{!}{%
\begin{tabular}{llrrrr}
\toprule
\textbf{Backend}
& \textbf{Method}
& \textbf{TTFT} \(\downarrow\)
& \textbf{TPOT} \(\downarrow\)
& \textbf{Throughput} \(\uparrow\)
& \textbf{Peak GPU} \(\downarrow\) \\
&
&
\textbf{(ms)}
& \textbf{(ms)}
& \textbf{(tok/s)}
& \textbf{(GB)} \\
\midrule

\multirow{2}{*}{FlashAttention-2}
& Full-KV
& \textbf{675.0}
& \textbf{59.74}
& \textbf{535.11}
& 29.54 \\

& \textbf{\myarch}
& 772.0
& 146.35
& 218.05
& \textbf{20.78} \\

\midrule

\multirow{2}{*}{Eager/SDPA}
& Full-KV
& \textbf{689.3}
& 84.05
& 379.26
& 24.38 \\

& \textbf{\myarch}
& 692.4
& \textbf{76.26}
& \textbf{416.44}
& \textbf{24.36} \\

\bottomrule
\end{tabular}%
}
\endgroup
\end{table}

\paragraph{Limitations and future work.}
The current implementation reconstructs active low-bit windows before attention and selectively falls back to SDPA at routing events to expose attention probabilities. These operations limit FlashAttention-2 throughput and introduce temporary workspace overhead. Future work will integrate dequantization, positional rotation, and mixed-precision attention directly into fused kernels, and investigate lower-cost importance estimators that avoid explicit attention-score materialization. The present evaluation also focuses on decoder-only models and fixed-size contiguous windows, leaving adaptive window boundaries and broader model architectures for future study.

\section{Conclusion}
\label{sec:conclusion}

We introduced \myarch, a byte-constrained KV-cache framework that routes historical windows across full-precision, recoverable INT2, and evicted tiers. Across LongBench, RULER, and GSM8K, \myarch improves performance--memory trade-offs over eviction and quantization baselines, while reducing peak GPU memory and improving eager decoding throughput. These results show that recoverable low-bit retention is a practical alternative to permanent eviction.

\bibliography{references}

@inproceedings{vaswani2017attention,
  title={Attention Is All You Need},
  author={Vaswani, Ashish and Shazeer, Noam and Parmar, Niki and Uszkoreit, Jakob and Jones, Llion and Gomez, Aidan N. and Kaiser, Lukasz and Polosukhin, Illia},
  booktitle={Advances in Neural Information Processing Systems},
  volume={30},
  year={2017}
}

@article{jiang2023mistral,
  title={Mistral 7B},
  author={Jiang, Albert Q and Sablayrolles, Alexandre and Mensch, Arthur and Bamford, Chris and Chaplot, Devendra Singh and de las Casas, Diego and Bressand, Florian and Lengyel, Gianna and Lample, Guillaume and Saulnier, Lucile and others},
  journal={arXiv preprint arXiv:2310.06825},
  year={2023}
}

@inproceedings{feng2026defensivekv,
  title={DefensiveKV: Taming the fragility of KV cache eviction in LLM inference},
  author={Feng, Yuan and Guo, Haoyu and Lv, Junlin and Zhou, S Kevin and Xie, Xike},
  booktitle={The Fourteenth International Conference on Learning Representations, 2026a. URL https://openreview. net/forum},
  year={2026}
}

@inproceedings{zhang2023h2o,
      title={{H2O}: Heavy-Hitter Oracle for Efficient Generative Inference of Large Language Models},
      author={Zhenyu Zhang and Ying Sheng and Tianyi Zhou and Tianlong Chen and Lianmin Zheng and Ruisi Cai and Zhao Song and Yuandong Tian and Christopher R\'{e} and Clark Barrett and others},
      booktitle={NeurIPS},
      year={2023}
}

@inproceedings{liu2023scissorhands,
      title={Scissorhands: Exploiting the Persistence of Importance Hypothesis for LLM KV Cache Compression at Test Time},
      author={Zichang Liu and Aditya Desai and Fangshuo Liao and Weitao Wang and Victor Xie and Zhaozhuo Xu and Anastasios Kyrillidis and Anshumali Shrivastava},
      booktitle={NeurIPS},
      year={2023}
}

@inproceedings{li2024snapkv,
      title={SnapKV: LLM Knows What You are Looking for Before Generation},
      author={Yuhong Li and Yingbing Huang and Bowen Yang and Bharat Venkitesh and Acyr Locatelli and Hanchen Ye and Tianle Cai and Patrick Lewis and Deming Chen},
      booktitle={NeurIPS},
      year={2024}
}

@misc{feng2024adakv,
      title={Ada-KV: Optimizing KV Cache Eviction by Adaptive Budget Allocation for Efficient LLM Inference}, 
      author={Yuan Feng and Junlin Lv and Yukun Cao and Xike Xie and S. Kevin Zhou},
      year={2024}
}

@misc{liu2024kivi,
      title={KIVI: A Tuning-Free Asymmetric 2bit Quantization for KV Cache}, 
      author={Zirui Liu and Jiayi Yuan and Hongye Jin and Shaochen Zhong and Zhaozhuo Xu and Vladimir Braverman and Beidi Chen and Xia Hu},
      year={2024}
}

@inproceedings{hooper2024kvquant,
      title={KVQuant: Towards 10 Million Context Length LLM Inference with KV Cache Quantization},
      author={Coleman Hooper and Sehoon Kim and Hiva Mohammadzadeh and Michael W Mahoney and Sophia Shao and Kurt Keutzer and Amir Gholami},
      booktitle={NeurIPS},
      year={2024}
}

@inproceedings{kwon2023pagedattention,
      title={Efficient Memory Management for Large Language Model Serving with PagedAttention},
      author={Woosuk Kwon and Zhuohan Li and Siyuan Zhuang and Ying Sheng and Lianmin Zheng and Cody Hao Yu and Joseph Gonzalez and Hao Zhang and Ion Stoica},
      booktitle={SOSP},
      year={2023}
}

@inproceedings{brown2020language,
  title={Language models are few-shot learners},
  author={Brown, Tom B and Mann, Benjamin and Ryder, Nick and Subbiah, Melanie and Kaplan, Jared and Dhariwal, Prafulla and Neelakantan, Arvind and Shyam, Pranav and Sastry, Girish and Askell, Amanda and others},
  booktitle={Advances in Neural Information Processing Systems},
  volume={33},
  pages={1877--1901},
  year={2020}
}

@article{touvron2023llama,
  title={Llama: Open and efficient foundation language models},
  author={Touvron, Hugo and Lavril, Thibaut and Izacard, Gautier and Martinet, Xavier and Lachaux, Marie-Anne and Lacroix, Timoth{\'e}e and Rozi{\`e}re, Baptiste and Goyal, Naman and Hambro, Eric and Azhar, Faisal and others},
  journal={arXiv preprint arXiv:2302.13971},
  year={2023}
}

@article{dubey2024llama3,
  title={The {Llama} 3 Herd of Models},
  author={Dubey, Abhimanyu and Jauhri, Abhinav and Pandey, Abhinav and Kadian, Abhishek and Al-Dahle, Ahmad and Letman, Aiesha and Mathur, Akhil and Schelten, Alan and Yang, Amy and Fan, Angela and others},
  journal={arXiv preprint arXiv:2407.21783},
  year={2024}
}

@article{pope2023efficiently,
  title={Efficiently scaling transformer inference},
  author={Pope, Reiner and Douglas, Sholto and Chowdhery, Aakanksha and Devlin, Jacob and Bradbury, James and Heek, Jonathan and Xiao, Kefan and Agrawal, Shivani and Dean, Jeff},
  journal={Proceedings of Machine Learning and Systems},
  volume={5},
  pages={606--624},
  year={2023}
}

@article{dao2023flashattention2,
  title={{FlashAttention-2}: Faster Attention with Better Parallelism and Work Partitioning},
  author={Dao, Tri},
  journal={arXiv preprint arXiv:2307.08691},
  year={2023}
}

@inproceedings{xiao2024streamingllm,
  title={Efficient Streaming Language Models with Attention Sinks},
  author={Xiao, Guangxuan and Tian, Yuandong and Chen, Beidi and Han, Song and Lewis, Mike},
  booktitle={International Conference on Learning Representations},
  year={2024}
}

@article{feng2025criticalkv,
  title={Identify Critical KV Cache in LLM Inference from an Output Perturbation Perspective},
  author={Feng, Yuan and Lv, Junlin and Guo, Haoyu and Cao, Yukun and Xie, Xike and Zhou, S. Kevin},
  journal={OpenReview preprint},
  year={2025}
}

@inproceedings{bai2024longbench,
  title={LongBench: A Bilingual, Multitask Benchmark for Long Context Understanding},
  author={Bai, Yushi and Lv, Xin and Zhang, Jiajie and Lyu, Hongchang and Tang, Jiankai and Huang, Zhidian and Du, Zhengxiao and Liu, Xiao and Zeng, Aohan and Hou, Lei and Dong, Yuxiao and Tang, Jie and Li, Juanzi},
  booktitle={Proceedings of the 62nd Annual Meeting of the Association for Computational Linguistics},
  year={2024}
}

@inproceedings{hsieh2024ruler,
  title={RULER: What's the Real Context Size of Your Long-Context Language Models?},
  author={Hsieh, Cheng-Ping and Sun, Simeng and Kriman, Samuel and Acharya, Shantanu and Rekesh, Dima and Jia, Fei and Zhang, Yang and Ginsburg, Boris},
  booktitle={Conference on Language Modeling},
  year={2024}
}

@inproceedings{ainslie2023gqa,
  title={Gqa: Training generalized multi-query transformer models from multi-head checkpoints},
  author={Ainslie, Joshua and Lee-Thorp, James and De Jong, Michiel and Zemlyanskiy, Yury and Lebr{\'o}n, Federico and Sanghai, Sumit},
  booktitle={Proceedings of the 2023 Conference on Empirical Methods in Natural Language Processing},
  pages={4895--4901},
  year={2023}
}

@article{he2024zipcache,
  title={Zipcache: Accurate and efficient kv cache quantization with salient token identification},
  author={He, Yefei and Zhang, Luoming and Wu, Weijia and Liu, Jing and Zhou, Hong and Zhuang, Bohan},
  journal={Advances in Neural Information Processing Systems},
  volume={37},
  pages={68287--68307},
  year={2024}
}

@article{mirzadeh2024gsm,
  title={Gsm-symbolic: Understanding the limitations of mathematical reasoning in large language models},
  author={Mirzadeh, Iman and Alizadeh, Keivan and Shahrokhi, Hooman and Tuzel, Oncel and Bengio, Samy and Farajtabar, Mehrdad},
  journal={arXiv preprint arXiv:2410.05229},
  year={2024}
}

@article{hearst-1997-text,
    title = "Text Tiling: Segmenting Text into Multi-paragraph Subtopic Passages",
    author = "Hearst, Marti A.",
    editor = "Hirschberg, Julia",
    journal = "Computational Linguistics",
    volume = "23",
    number = "1",
    year = "1997",
    address = "Cambridge, MA",
    publisher = "MIT Press",
    url = "https://aclanthology.org/J97-1003/",
    pages = "33--64"
}

@article{barzilay-lapata-2008-modeling,
    title = "Modeling Local Coherence: An Entity-Based Approach",
    author = "Barzilay",
    journal = "Computational Linguistics",
    volume = "34",
    number = "1",
    year = "2008",
    url = "https://aclanthology.org/J08-1001/",
    doi = "10.1162/coli.2008.34.1.1",
    pages = "1--34"
}

@inproceedings{ethayarajh-2019-contextual,
    title = "How Contextual are Contextualized Word Representations? {C}omparing the Geometry of {BERT}, {ELM}o, and {GPT}-2 Embeddings",
    author = "Ethayarajh, Kawin",
    editor = "Inui, Kentaro  and
      Jiang, Jing  and
      Ng, Vincent  and
      Wan, Xiaojun",
    booktitle = "Proceedings of the 2019 Conference on Empirical Methods in Natural Language Processing and the 9th International Joint Conference on Natural Language Processing (EMNLP-IJCNLP)",
    month = nov,
    year = "2019",
    address = "Hong Kong, China",
    publisher = "Association for Computational Linguistics",
    url = "https://aclanthology.org/D19-1006/",
    doi = "10.18653/v1/D19-1006",
    pages = "55--65"
}

@article{joshi-etal-2020-spanbert,
    title = "{S}pan{BERT}: Improving Pre-training by Representing and Predicting Spans",
    author = "Joshi, Mandar  and
      Chen, Danqi  and
      Liu, Yinhan  and
      Weld, Daniel S.  and
      Zettlemoyer, Luke  and
      Levy, Omer",
    editor = "Johnson, Mark  and
      Roark, Brian  and
      Nenkova, Ani",
    journal = "Transactions of the Association for Computational Linguistics",
    volume = "8",
    year = "2020",
    address = "Cambridge, MA",
    publisher = "MIT Press",
    url = "https://aclanthology.org/2020.tacl-1.5/",
    doi = "10.1162/tacl_a_00300",
    pages = "64--77"
}

@InProceedings{pmlr-v235-tang24l,
  title = 	 {{QUEST}: Query-Aware Sparsity for Efficient Long-Context {LLM} Inference},
  author =       {Tang, Jiaming and Zhao, Yilong and Zhu, Kan and Xiao, Guangxuan and Kasikci, Baris and Han, Song},
  booktitle = 	 {Proceedings of the 41st International Conference on Machine Learning},
  pages = 	 {47901--47911},
  year = 	 {2024},
  editor = 	 {Salakhutdinov, Ruslan and Kolter, Zico and Heller, Katherine and Weller, Adrian and Oliver, Nuria and Scarlett, Jonathan and Berkenkamp, Felix},
  volume = 	 {235},
  series = 	 {Proceedings of Machine Learning Research},
  month = 	 {21--27 Jul},
  publisher =    {PMLR},
  url = 	 {https://proceedings.mlr.press/v235/tang24l.html}
}

@article{grosz-etal-1995-centering,
    title = "{C}entering: A Framework for Modeling the Local Coherence of Discourse",
    author = "Grosz",
    editor = "Hirschberg, Julia",
    journal = "Computational Linguistics",
    volume = "21",
    number = "2",
    year = "1995",
    address = "Cambridge, MA",
    publisher = "MIT Press",
    url = "https://aclanthology.org/J95-2003/",
    pages = "203--225"
}

@article{passonneau-litman-1997-discourse,
    title = "Discourse Segmentation by Human and Automated Means",
    author = "Passonneau",
    editor = "Hirschberg, Julia",
    journal = "Computational Linguistics",
    volume = "23",
    number = "1",
    year = "1997",
    address = "Cambridge, MA",
    publisher = "MIT Press",
    url = "https://aclanthology.org/J97-1005/",
    pages = "103--139"
}

@inproceedings{koshorek-etal-2018-text,
    title = "Text Segmentation as a Supervised Learning Task",
    author = "Koshorek",
    editor = "Walker, Marilyn  and
      Ji, Heng  and
      Stent, Amanda",
    booktitle = "Proceedings of the 2018 Conference of the North {A}merican Chapter of the Association for Computational Linguistics: Human Language Technologies, Volume 2 (Short Papers)",
    month = jun,
    year = "2018",
    address = "New Orleans, Louisiana",
    publisher = "Association for Computational Linguistics",
    url = "https://aclanthology.org/N18-2075/",
    doi = "10.18653/v1/N18-2075",
    pages = "469--473"
}

@inproceedings{clark-etal-2019-bert,
    title = "What Does {BERT} Look at? An Analysis of {BERT}{'}s Attention",
    author = "Clark, Kevin  and
      Khandelwal, Urvashi  and
      Levy, Omer  and
      Manning, Christopher D.",
    editor = "Linzen, Tal  and
      Chrupa{\l}a, Grzegorz  and
      Belinkov, Yonatan  and
      Hupkes, Dieuwke",
    booktitle = "Proceedings of the 2019 ACL Workshop BlackboxNLP: Analyzing and Interpreting Neural Networks for NLP",
    month = aug,
    year = "2019",
    address = "Florence, Italy",
    publisher = "Association for Computational Linguistics",
    url = "https://aclanthology.org/W19-4828/",
    doi = "10.18653/v1/W19-4828",
    pages = "276--286"
}

@InProceedings{pmlr-v235-ribar24a,
  title = 	 {{S}par{Q} Attention: Bandwidth-Efficient {LLM} Inference},
  author =       {Ribar, Luka and Chelombiev, Ivan and Hudlass-Galley, Luke and Blake, Charlie and Luschi, Carlo and Orr, Douglas},
  booktitle = 	 {Proceedings of the 41st International Conference on Machine Learning},
  pages = 	 {42558--42583},
  year = 	 {2024},
  editor = 	 {Salakhutdinov, Ruslan and Kolter, Zico and Heller, Katherine and Weller, Adrian and Oliver, Nuria and Scarlett, Jonathan and Berkenkamp, Felix},
  volume = 	 {235},
  series = 	 {Proceedings of Machine Learning Research},
  month = 	 {21--27 Jul},
  publisher =    {PMLR},
  url = 	 {https://proceedings.mlr.press/v235/ribar24a.html}
}

@inproceedings{devoto2024simple,
  title={A simple and effective l\_2 norm-based strategy for kv cache compression},
  author={Devoto, Alessio and Zhao, Yu and Scardapane, Simone and Minervini, Pasquale},
  booktitle={Proceedings of the 2024 Conference on Empirical Methods in Natural Language Processing},
  pages={18476--18499},
  year={2024}
}

@article{yang2025qwen3,
  title={Qwen3 technical report},
  author={Yang, An and Li, Anfeng and Yang, Baosong and Zhang, Beichen and Hui, Binyuan and Zheng, Bo and Yu, Bowen and Gao, Chang and Huang, Chengen and Lv, Chenxu and others},
  journal={arXiv preprint arXiv:2505.09388},
  year={2025}
}

@article{cobbe2021training,
  title={Training verifiers to solve math word problems},
  author={Cobbe, Karl and Kosaraju, Vineet and Bavarian, Mohammad and Chen, Mark and Jun, Heewoo and Kaiser, Lukasz and Plappert, Matthias and Tworek, Jerry and Hilton, Jacob and Nakano, Reiichiro and others},
  journal={arXiv preprint arXiv:2110.14168},
  year={2021}
}

\newpage
\appendix
\onecolumn

\section{Algorithm}
\label{app:algorithm}

\begin{algorithm}[!h]
\caption{\myarch Dynamic Window Routing}
\label{alg:QEvict}
\begin{algorithmic}[1]
\REQUIRE Previous tiers $\mathcal{F}_{t-\Omega}$ and
$\mathcal{Q}_{t-\Omega}$
\REQUIRE Aged local windows $\mathcal{L}_{\mathrm{aged}}$,
scores $\bar S_t(w)$, capacities $K_f,K_q$, ledger $\Lambda$
\STATE $\mathcal{C}_{\mathrm{cand}}
\gets
\mathcal{F}_{t-\Omega}
\cup
\mathcal{Q}_{t-\Omega}
\cup
\mathcal{L}_{\mathrm{aged}}$
\STATE $\mathcal{F}_t
\gets
\TopK_{K_f}
\{\bar S_t(w):w\in\mathcal{C}_{\mathrm{cand}}\}$
\STATE $\mathcal{Q}_t
\gets
\TopK_{K_q}
\{\bar S_t(w):
w\in\mathcal{C}_{\mathrm{cand}}\setminus\mathcal{F}_t\}$
\STATE $\mathcal{E}_t
\gets
\mathcal{C}_{\mathrm{cand}}
\setminus
(\mathcal{F}_t\cup\mathcal{Q}_t)$

\FOR{$w\in\mathcal{F}_t$}
    \IF{$w\in\mathcal{Q}_{t-\Omega}$}
        \STATE $(\hat K_w,\hat V_w)
        \gets
        \operatorname{Dequantize}(\Lambda_w)$
        \STATE mark $\Lambda_w$ dormant
    \ENDIF
\ENDFOR

\FOR{$w\in\mathcal{Q}_t$}
    \IF{$w$ has no ledger entry}
        \STATE quantize $w$ once and create $\Lambda_w$
    \ELSIF{$\Lambda_w$ is dormant}
        \STATE reactivate $\Lambda_w$
    \ENDIF
\ENDFOR

\FOR{$w\in\mathcal{E}_t$}
    \STATE release $w$ and delete $\Lambda_w$ if present
\ENDFOR

\STATE reconstruct the effective cache in chronological order
\STATE \textbf{return}
$(\mathcal{F}_t,\mathcal{Q}_t,\mathcal{E}_t,\Lambda)$
\end{algorithmic}
\end{algorithm}

\section{Complete Experimental Details}
\label{app:experiments}

This section specifies the models, inference protocol, baselines, comparison regimes, memory accounting, and default \myarch configuration used throughout the evaluation.

\subsection{Models and Inference Protocol}
\label{app:models_protocol}

We evaluate \myarch on Llama-3.1-8B-Instruct~\citep{dubey2024llama3}, Qwen2.5-7B-Instruct~\citep{yang2025qwen3}, and Mistral-7B-Instruct-v0.2~\citep{jiang2023mistral}. \myarch is training-free and modifies only KV-cache management at inference time. Within each benchmark, all methods use the same checkpoint, tokenizer, prompt template, input and generation limits, and answer post-processing. Unless stated otherwise, decoding is greedy.

We evaluate three complementary capabilities: long-context understanding on LongBench~\citep{bai2024longbench}, controlled retrieval and aggregation on RULER~\citep{hsieh2024ruler}, and multi-step mathematical reasoning on GSM8K~\citep{cobbe2021training}.

\subsection{Baselines}
\label{app:baselines}

We compare against eviction and quantization methods, whose technical details are reviewed in Section~\ref{sec:related_work}.

\paragraph{Eviction baselines.} The eviction baselines are StreamingLLM~\citep{xiao2024streamingllm},  SnapKV~\citep{li2024snapkv}, AdaKV~\citep{feng2024adakv}, CriticalKV~\citep{feng2025criticalkv}, DefensiveKV, and Layer-DefensiveKV~\citep{feng2026defensivekv}. They cover sink- and recency-based retention, accumulated-attention scoring, prompt observation, head-adaptive allocation, value-aware importance, and drift-aware eviction.

\paragraph{Quantization baselines.} We compare against KIVI~\citep{liu2024kivi}, KVQuant~\citep{hooper2024kvquant}, and ZipCache~\citep{he2024zipcache}. These methods retain broad historical coverage at reduced precision, whereas \myarch jointly determines residency and precision through full-precision, quantized, and evicted tiers.

\paragraph{Uncompressed reference.} Full-KV stores the complete key and value cache in the model's native inference datatype and serves as the uncompressed reference.

\subsection{Comparison Regimes}
\label{app:comparison_regimes}

Eviction and quantization methods are evaluated under separate protocols because nominal compression settings do not necessarily yield the same physical memory footprint.

\paragraph{Eviction baselines.} For every method \(\mathcal{A}\), we define the measured KV-memory ratio as \(\rho_{\mathcal{A}}=M_{\mathcal{A}}/M_{\mathrm{FullKV}}\). LongBench uses \(\rho_{\mathcal{A}}\in\{0.05,0.10,0.20\}\), RULER 32K uses \(\rho_{\mathcal{A}} = 0.20\), and GSM8K uses \(\rho_{\mathcal{A}}\in\{0.10,0.20,0.40,0.60,0.80,1.0\}\) for the eviction comparisons.

\paragraph{Quantization baselines.} Quantization methods are compared at their measured performance--memory operating points rather than by nominal bit width.

\subsection{Memory Accounting}
\label{app:memory_accounting}

For a model with \(L\) layers, \(H_{kv}\) KV heads, head dimension \(d_h\), sequence length \(S\), and \(c_f\) bytes per element, the uncompressed KV-cache footprint is
\begin{equation}
    M_{\mathrm{FullKV}}
    =
    2 L H_{kv} d_h S c_f,
    \label{eq:appendix_fullkv_memory}
\end{equation}
where the factor of two accounts for keys and values.

All persistent cache-related storage is included in the measured budget, including retained KV tensors, packed codes, quantization parameters, full-precision residual regions, outliers, and indexing metadata when applicable. For \myarch,
\begin{equation}
\begin{aligned}
    M_{\mathrm{QEvict}}
    =\;&
    M_{\mathrm{sink}}
    + M_{\mathrm{recent}}
    + M_{\mathrm{FP}}
    + M_{\mathrm{Q}}
    + M_{\mathrm{quant\text{-}meta}}
    + M_{\mathrm{position}}
    + M_{\mathrm{ledger}} .
\end{aligned}
\label{eq:appendix_QEvict_memory}
\end{equation}
Here, \(M_{\mathrm{Q}}\) contains packed low-bit keys and values, \(M_{\mathrm{quant\text{-}meta}}\) contains their scales and offsets, and \(M_{\mathrm{ledger}}\) contains persistent migration records. Temporary workspaces are excluded from the KV-cache budget and captured separately by peak GPU memory.

\subsection{\myarch Configuration}
\label{app:QEvict_configuration}

Unless explicitly varied, \myarch uses routing interval \(\Omega=8\), five sink tokens, and an INT2 recoverable tier. After allocating the protected regions, a fraction \(q=0.70\) of the remaining historical budget is assigned to the quantized tier and \(1-q\) to the full-precision historical tier. This configuration is used for LongBench, RULER, and GSM8K. Sensitivity to \(\Omega\), \(q\), and quantization precision is reported in Appendix~\ref{app:ablations}.

% \input{Appendix/C_longbench_evaluation_revised}

% ============================================================
% LONGBENCH
% ============================================================

\section{LongBench Evaluation}
\label{app:longbench}

\subsection{Task Families and Metrics}
\label{app:longbench_tasks}

We evaluate 12 LongBench tasks spanning the four families summarized in Table~\ref{tab:longbench_task_families} and we store 128 recent tokens for this task.

\begin{table*}[!h]
\centering
\caption{\textbf{LongBench task families used in our evaluation.}}
\label{tab:longbench_task_families}
\small
\setlength{\tabcolsep}{5pt}
\renewcommand{\arraystretch}{1.12}

\begin{tabularx}{\textwidth}{
@{}
>{\raggedright\arraybackslash\bfseries}p{0.18\textwidth}
>{\raggedright\arraybackslash}p{0.31\textwidth}
>{\raggedright\arraybackslash}X
@{}
}
\toprule
\textbf{Task Family}
& \textbf{Datasets}
& \textbf{Evaluation Focus} \\
\midrule

Single-Document QA
& NarrativeQA, Qasper, MultiFieldQA-en
& Retrieving evidence from one long document. \\
\addlinespace[2pt]

Multi-Document QA
& HotpotQA, 2WikiMultihopQA, MuSiQue
& Retrieving and combining evidence across documents. \\
\addlinespace[2pt]

Summarization
& GovReport, QMSum, MultiNews
& Preserving information distributed throughout the input. \\
\addlinespace[2pt]

Few-Shot Learning
& TREC, TriviaQA, SAMSum
& Retaining demonstrations and task structure. \\
% \addlinespace[2pt]

% Synthetic Retrieval
% & PassageCount, PassageRetrieval-en
% & Controlled retrieval and counting. \\

\bottomrule
\end{tabularx}
\end{table*}

\providecommand{\lbbest}[1]{\textbf{#1}}
\providecommand{\lbsecond}[1]{\underline{#1}}

\begin{table*}[!h]
\centering

\caption{\textbf{LongBench performance on Qwen2.5-7B-Instruct.}
We compare \myarch{} with representative KV-cache eviction methods
under matched \(20\%\), \(10\%\), and \(5\%\) budgets, and with
quantization baselines at comparable memory footprints. KV memory is
reported relative to the full-precision cache. Excluding Full-KV, the
best and second-best distinct results within each comparison group are
shown in bold and underlined, respectively.}

\label{tab:longbench_qwen25_7b}

\vspace{-1mm}

\begingroup
\fontsize{5.4}{5.9}\selectfont
\setlength{\tabcolsep}{1.6pt}
\renewcommand{\arraystretch}{0.90}

\resizebox{\textwidth}{!}{%
\begin{tabular}{@{}lVcVcccVcccVcccVccc@{}}
\toprule

\multirow{2}{*}{\textbf{Method}} &
\multirow{2}{*}{
  \shortstack{\textbf{KV}\\\textbf{memory}}
} &
\multicolumn{3}{cV}{\textbf{Single-Doc QA}} &
\multicolumn{3}{cV}{\textbf{Multi-Doc QA}} &
\multicolumn{3}{cV}{\textbf{Summarization}} &
\multicolumn{3}{c}{\textbf{Few-Shot Learning}} \\

\cmidrule(lr){3-5}
\cmidrule(lr){6-8}
\cmidrule(lr){9-11}
\cmidrule(lr){12-14}

& &
\textbf{Nar.QA} &
\textbf{Qasper} &
\textbf{Mul.QA} &
\textbf{Hot.QA} &
\textbf{2Wi.QA} &
\textbf{Musique} &
\textbf{Gov.Re.} &
\textbf{QMSum} &
\textbf{M.News} &
\textbf{TREC} &
\textbf{Tri.QA} &
\textbf{SAMSum} \\

\midrule
\multicolumn{14}{c}{\emph{Qwen2.5-7B-Instruct}} \\
\cmidrule(lr){1-14}

\rowcolor{fullkvgray}
Full-KV
& 100\%
& 29.33
& 46.36
& 50.27
& 55.96
& 42.49
& 27.55
& 33.73
& 24.27
& 25.41
& 73.50
& 86.51
& 41.20 \\

% ===========================================================
% Eviction: 20% KV-Cache Budget
% ===========================================================
\midrule
\multicolumn{14}{c}{\textsc{Eviction: 20\% KV-Cache Budget}} \\
\cmidrule(lr){1-14}

StreamingLLM
& 20\%
& 18.56
& 20.05
& 23.83
& 35.02
& 27.09
& 14.34
& 28.50
& 18.99
& 21.05
& 57.00
& 78.49
& 42.96 \\

SnapKV
& 20\%
& 22.21
& 22.43
& 30.67
& 45.89
& 28.77
& 22.00
& 28.98
& 19.94
& 21.28
& 49.00
& 87.61
& 40.20 \\

AdaKV
& 20\%
& \lbbest{23.35}
& 22.53
& 29.57
& 44.57
& 28.65
& 21.37
& 28.61
& 20.09
& 21.45
& 53.50
& 87.89
& 40.00 \\

CriticalKV
& 20\%
& \lbsecond{22.82}
& 26.72
& 31.74
& \lbsecond{46.85}
& \lbsecond{37.09}
& \lbbest{24.95}
& 29.75
& 20.73
& 22.15
& 53.00
& \lbsecond{88.94}
& \lbsecond{45.61} \\

DefensiveKV
& 20\%
& 20.80
& 30.30
& 34.60
& 44.30
& 32.00
& 19.20
& \lbsecond{30.40}
& 20.60
& 22.40
& 37.50
& 87.20
& \lbbest{46.10} \\

Layer-Def.\ KV
& 20\%
& 20.46
& \lbsecond{30.45}
& \lbsecond{39.34}
& 46.54
& 36.16
& 21.28
& \lbbest{31.03}
& \lbsecond{21.49}
& \lbsecond{22.50}
& \lbsecond{61.00}
& 87.23
& 45.36 \\

\rowcolor{qevictyellow}
\textbf{\myarch{}}
& 20\%
& 19.97
& \lbbest{35.74}
& \lbbest{44.28}
& \lbbest{49.15}
& \lbbest{37.43}
& \lbsecond{24.37}
& 26.68
& \lbbest{22.04}
& \lbbest{25.28}
& \lbbest{67.00}
& \lbbest{89.63}
& 44.17 \\

% ===========================================================
% Eviction: 10% KV-Cache Budget
% ===========================================================
\midrule
\multicolumn{14}{c}{\textsc{Eviction: 10\% KV-Cache Budget}} \\
\cmidrule(lr){1-14}

StreamingLLM
& 10\%
& 18.20
& 16.01
& 21.25
& 27.39
& 22.83
& 11.38
& 25.88
& 18.30
& 17.87
& \lbsecond{49.50}
& 71.06
& 41.03 \\

SnapKV
& 10\%
& 19.43
& 14.06
& 24.55
& 36.48
& 23.98
& 17.13
& 26.35
& 18.34
& 18.69
& 44.50
& 87.71
& 39.17 \\

AdaKV
& 10\%
& 20.78
& 15.22
& 25.67
& 37.27
& 22.13
& 17.00
& 26.27
& 18.09
& 19.04
& 43.50
& 87.14
& 39.67 \\

CriticalKV
& 10\%
& 19.90
& 19.67
& 25.33
& \lbsecond{40.75}
& 27.36
& \lbsecond{19.21}
& 27.02
& 18.83
& 19.99
& 46.50
& \lbbest{88.64}
& \lbbest{45.18} \\

DefensiveKV
& 10\%
& 19.00
& 20.10
& 26.50
& 34.00
& \lbsecond{31.70}
& 16.40
& \lbsecond{28.00}
& \lbsecond{19.10}
& 20.20
& 27.50
& \lbsecond{87.90}
& 43.60 \\

Layer-Def.\ KV
& 10\%
& \lbsecond{21.19}
& \lbsecond{21.93}
& \lbsecond{26.81}
& 37.53
& 22.32
& 17.43
& 20.44
& 18.83
& \lbsecond{21.99}
& 43.50
& 87.51
& \lbsecond{44.62} \\

\rowcolor{qevictyellow}
\textbf{\myarch{}}
& 10\%
& \lbbest{23.88}
& \lbbest{34.46}
& \lbbest{37.71}
& \lbbest{49.22}
& \lbbest{37.13}
& \lbbest{23.56}
& \lbbest{29.35}
& \lbbest{22.10}
& \lbbest{23.72}
& \lbbest{64.50}
& 83.43
& 42.55 \\

% ===========================================================
% Eviction: 5% KV-Cache Budget
% ===========================================================
\midrule
\multicolumn{14}{c}{\textsc{Eviction: 5\% KV-Cache Budget}} \\
\cmidrule(lr){1-14}

StreamingLLM
& 5\%
& 11.98
& 12.02
& 20.28
& 28.05
& 22.13
& 9.87
& 23.42
& 17.10
& 15.82
& \lbsecond{38.50}
& 45.69
& 34.11 \\
% The supplied value for QMSum is 177.10.
% Verify whether this was intended to be 17.71.

SnapKV
& 5\%
& \lbsecond{18.17}
& 14.35
& 19.85
& 29.09
& 21.97
& 14.30
& 24.46
& 17.46
& 16.80
& 29.50
& 86.00
& 41.51 \\

AdaKV
& 5\%
& 17.46
& 14.41
& 21.06
& 29.60
& 22.84
& 12.85
& 24.42
& 17.32
& 17.32
& 33.00
& 87.80
& 41.54 \\

CriticalKV
& 5\%
& 17.87
& 14.48
& 21.77
& \lbsecond{35.36}
& 20.39
& 14.37
& 24.39
& 17.38
& 17.03
& 36.00
& \lbbest{88.45}
& \lbsecond{43.11} \\

DefensiveKV
& 5\%
& 16.80
& \lbsecond{14.90}
& 21.20
& 30.50
& \lbsecond{24.80}
& 14.30
& \lbsecond{25.60}
& \lbsecond{17.90}
& \lbsecond{17.60}
& 27.50
& \lbsecond{87.90}
& \lbbest{43.20} \\

Layer-Def.\ KV
& 5\%
& 15.86
& 14.33
& \lbsecond{22.12}
& 30.51
& 24.50
& \lbsecond{14.94}
& 25.24
& 17.19
& 17.08
& 34.00
& \lbsecond{87.90}
& 42.05 \\

\rowcolor{qevictyellow}
\textbf{\myarch{}}
& 5\%
& \lbbest{23.73}
& \lbbest{28.61}
& \lbbest{33.78}
& \lbbest{49.77}
& \lbbest{35.16}
& \lbbest{22.89}
& \lbbest{28.74}
& \lbbest{21.85}
& \lbbest{18.54}
& \lbbest{59.50}
& 84.93
& 41.48 \\

% ============================================================
% Quantization comparison
% ============================================================
\midrule
\multicolumn{14}{c}{\textsc{Quantization}} \\
\cmidrule(lr){1-14}

\rowcolor{qevictyellow}
\textbf{\myarch{}}
& 10\%
& 23.88
& 34.46
& 37.71
& \lbsecond{49.22}
& 37.13
& 23.56
& \lbsecond{29.35}
& 22.10
& 23.72
& 64.50
& 83.43
& 42.55 \\

\rowcolor{qevictyellow}
\textbf{\myarch{}}
& 20\%
& 19.97
& \lbsecond{35.74}
& 44.28
& 49.15
& 37.43
& \lbsecond{24.37}
& 26.68
& 22.04
& \lbbest{25.28}
& \lbsecond{67.00}
& \lbsecond{89.63}
& \lbbest{44.17} \\

KIVI-2b
& $\sim$20\%
& \lbbest{24.23}
& 28.25
& 42.64
& 45.90
& 38.07
& 24.15
& 25.41
& \lbsecond{22.17}
& 22.55
& 65.25
& 76.21
& \lbsecond{43.94} \\

ZipCache-4b (70\%)
& $\sim$22\%
& 22.82
& 34.53
& \lbbest{45.81}
& 48.92
& \lbbest{40.03}
& 23.50
& 29.10
& 21.65
& 23.49
& 34.50
& \lbbest{90.97}
& 40.15 \\

KVQuant-2b (s1\%)
& $\sim$15\%
& 20.98
& 32.57
& 42.58
& 41.53
& 32.25
& 16.00
& 28.99
& 21.93
& 23.02
& 57.25
& 79.54
& 35.96 \\

KVQuant-3b (s1\%)
& $\sim$22\%
& \lbsecond{24.19}
& \lbbest{37.07}
& \lbsecond{45.16}
& \lbbest{53.94}
& \lbsecond{39.43}
& \lbbest{26.24}
& \lbbest{31.38}
& \lbbest{23.59}
& \lbsecond{23.86}
& \lbbest{69.00}
& 85.44
& 39.33 \\

\bottomrule
\end{tabular}%
}

\endgroup
\end{table*}

We use the official metric for each dataset and report task-level scores.

\subsection{Complete Results}
\label{app:longbench_results}

% Retain these labels if they are referenced elsewhere.
\label{app:longbench_eviction}
\label{app:longbench_quantization}
\label{app:longbench_mistral}

All baselines, configurations, memory-accounting rules, and comparison protocols follow Appendix~\ref{app:experiments}, particularly Appendix~\ref{app:comparison_regimes}. Results for LongBench on Llama-3.1-8B-Instruct~\citep{dubey2024llama3} and Mistral-7B-Instruct-v0.2~\citep{jiang2023mistral} are reported in the main-paper Table~\ref{tab:longbench_combined_two_models} and for Qwen2.5-7B-Instruct~\citep{yang2025qwen3} is reported in Table~\ref{tab:longbench_qwen25_7b}.

% \input{Tables/Appendix/longbench_mistral_eviction}
% \input{Tables/Appendix/longbench_mistral_quantization}

% ============================================================
% LONGBENCH PERFORMANCE--MEMORY PARETO ANALYSIS
% ============================================================

\section{LongBench Performance--Memory Pareto Analysis}
\label{app:pareto}

Fixed-budget comparisons capture performance at only a small number of operating points and may obscure the overall trade-off between task quality and KV-cache memory. We therefore analyse the performance--memory Pareto frontier for each LongBench task family, model, and baseline class.

Let \((\rho_i,S_i)\) denote an operating point, where \(\rho_i\) is the measured KV-memory ratio and \(S_i\) is the corresponding task-family score. Operating point \(i\) dominates \(j\) when
\begin{equation}
    \rho_i \leq \rho_j,
    \qquad
    S_i \geq S_j,
\end{equation}
with at least one strict inequality. The Pareto frontier is the set of non-dominated operating points. Points closer to the upper-left corner are preferable because they achieve higher task performance with lower KV-memory usage.

\begin{figure*}[p]
    \centering
    \captionsetup[subfigure]{
        font=small,
        labelfont=bf,
        justification=centering,
        skip=2pt
    }

    \begin{subfigure}[t]{0.485\textwidth}
        \centering
        \includegraphics[width=\linewidth]
        {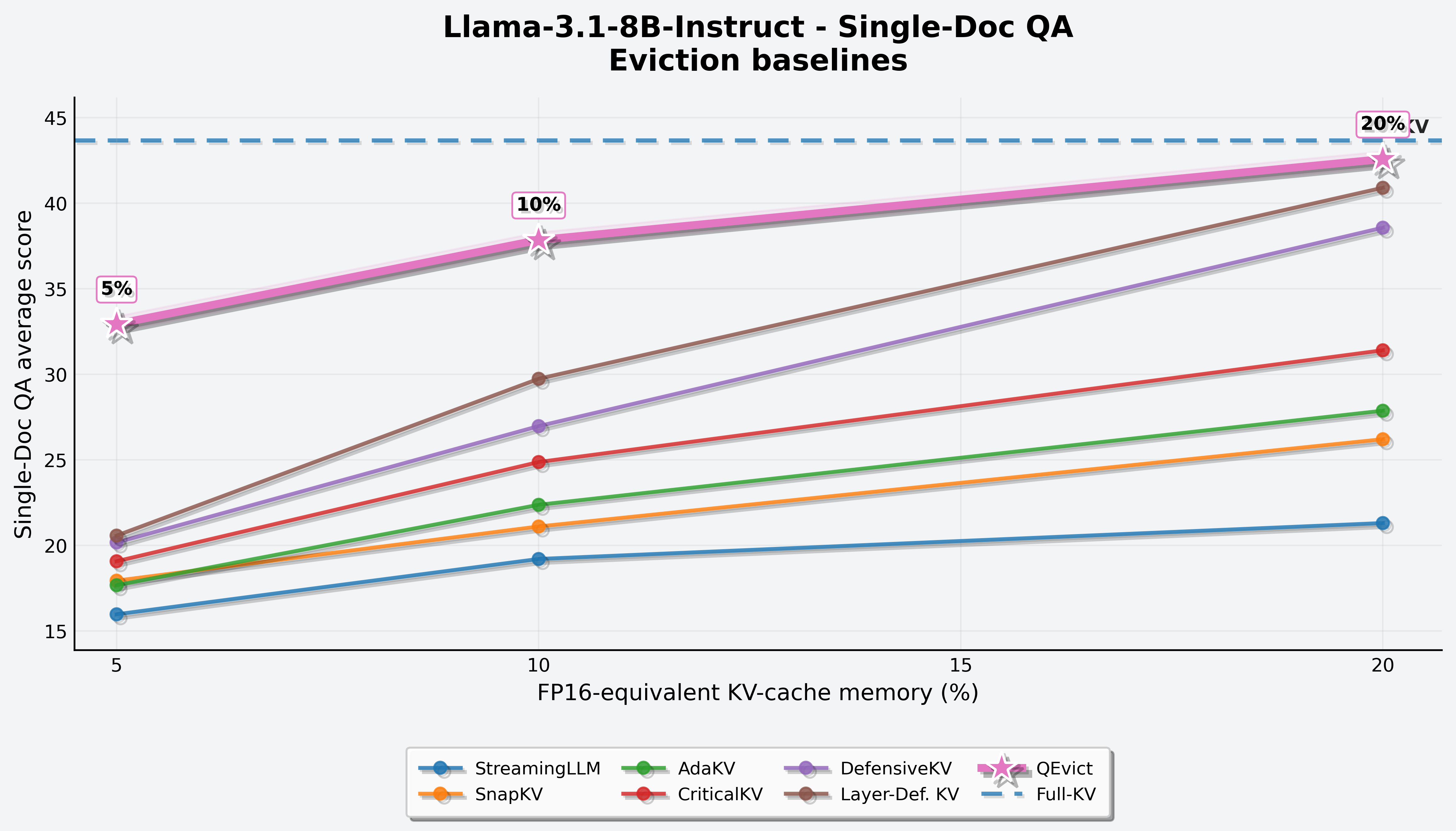}
        \caption{Single-Document QA: eviction.}
        \label{fig:llama-pareto-single-eviction}
    \end{subfigure}
    \hfill
    \begin{subfigure}[t]{0.485\textwidth}
        \centering
        \includegraphics[width=\linewidth]
        {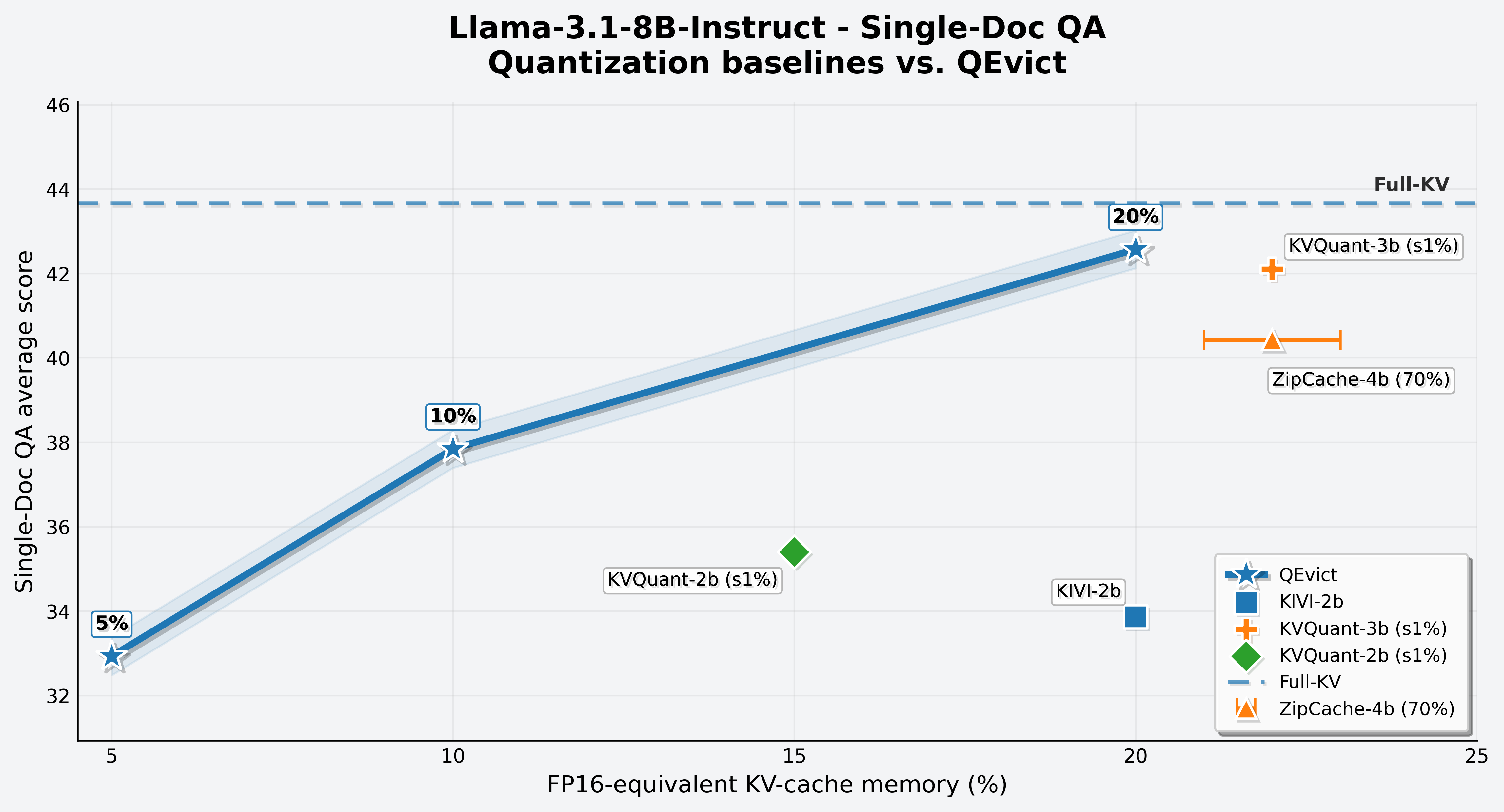}
        \caption{Single-Document QA: quantization.}
        \label{fig:llama-pareto-single-quantization}
    \end{subfigure}

    \vspace{2mm}

    \begin{subfigure}[t]{0.485\textwidth}
        \centering
        \includegraphics[width=\linewidth]
        {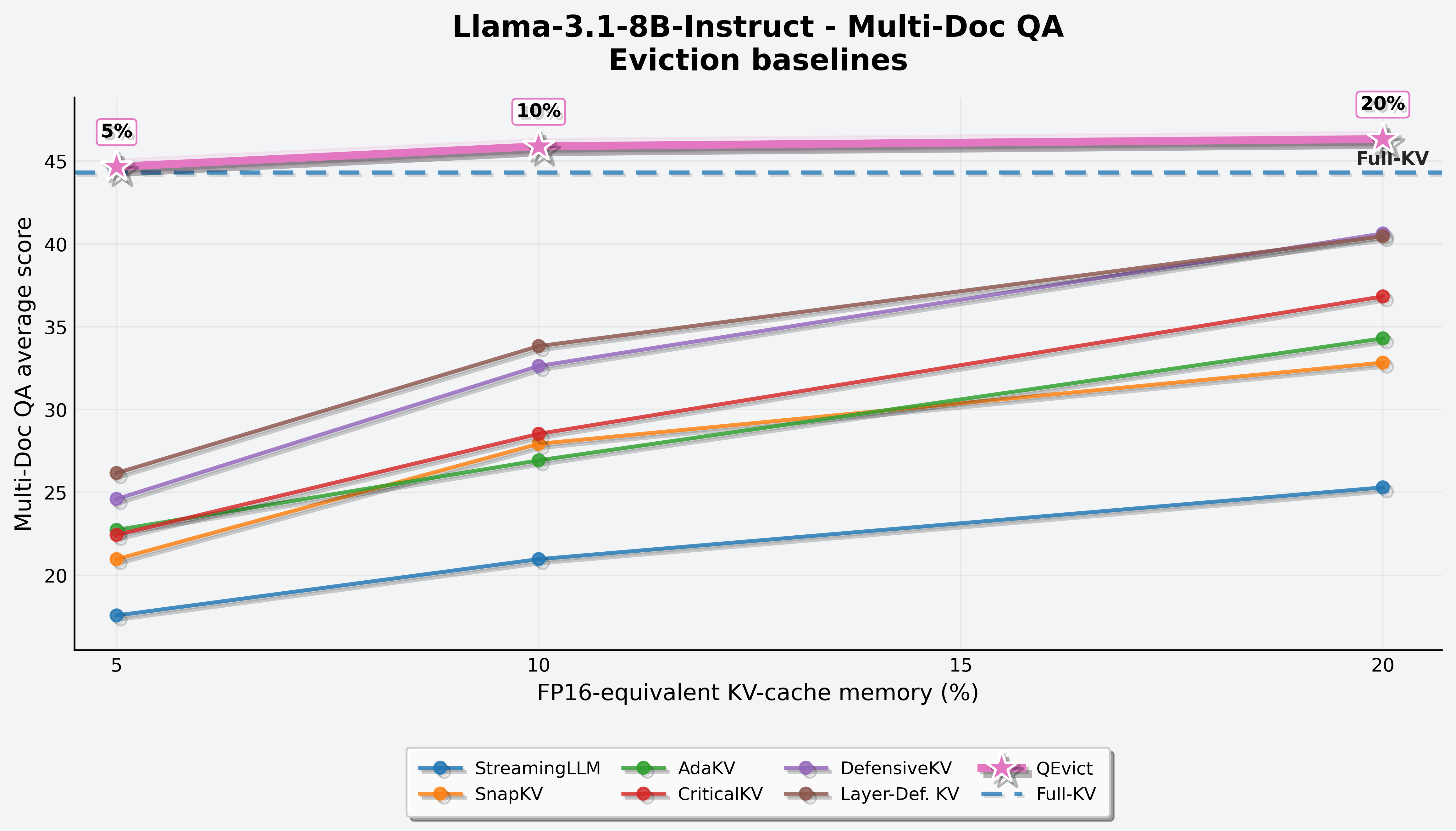}
        \caption{Multi-Document QA: eviction.}
        \label{fig:llama-pareto-multi-eviction}
    \end{subfigure}
    \hfill
    \begin{subfigure}[t]{0.485\textwidth}
        \centering
        \includegraphics[width=\linewidth]
        {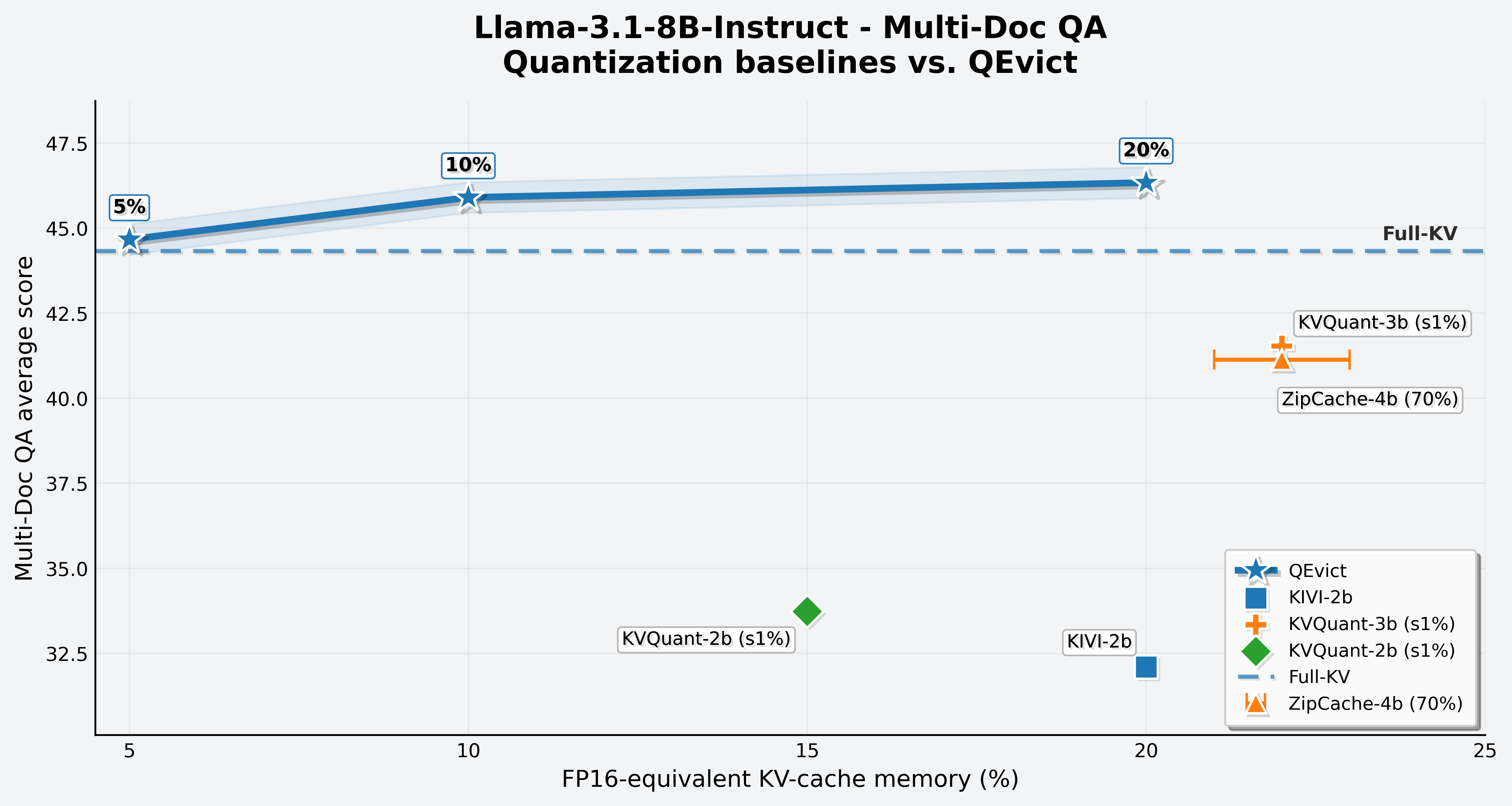}
        \caption{Multi-Document QA: quantization.}
        \label{fig:llama-pareto-multi-quantization}
    \end{subfigure}

    \vspace{2mm}

    \begin{subfigure}[t]{0.485\textwidth}
        \centering
        \includegraphics[width=\linewidth]
        {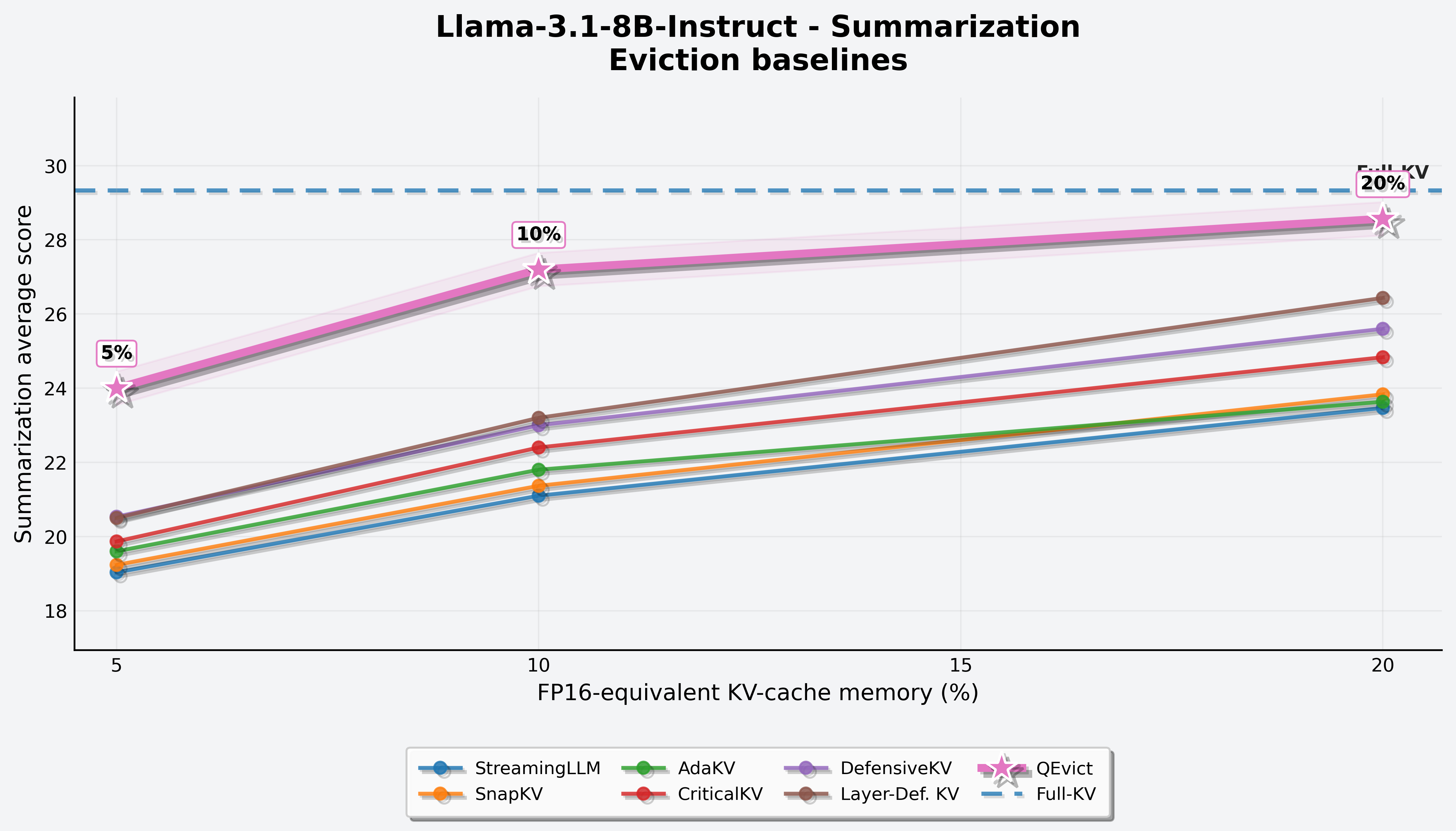}
        \caption{Summarization: eviction.}
        \label{fig:llama-pareto-summarization-eviction}
    \end{subfigure}
    \hfill
    \begin{subfigure}[t]{0.485\textwidth}
        \centering
        \includegraphics[width=\linewidth]
        {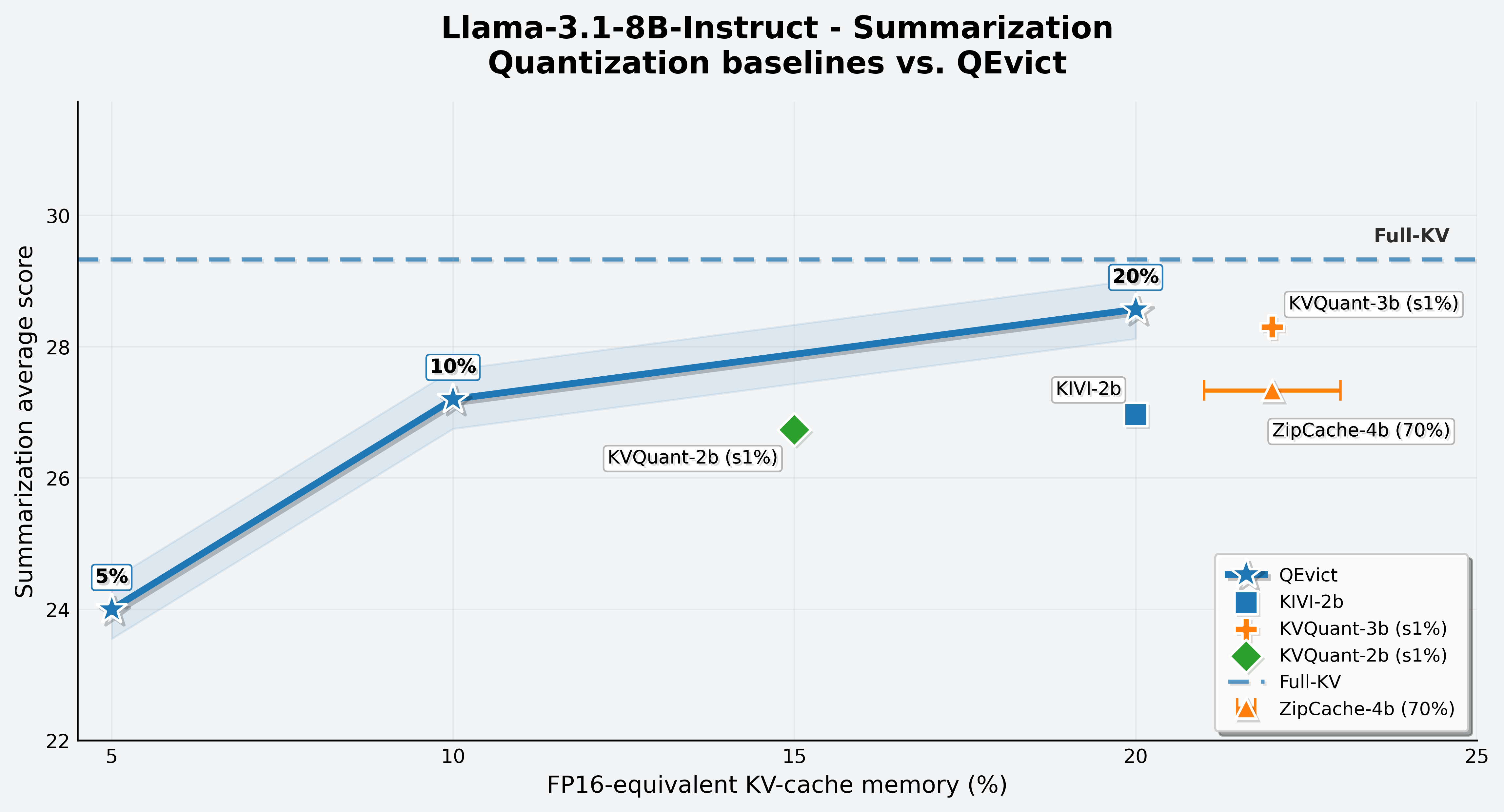}
        \caption{Summarization: quantization.}
        \label{fig:llama-pareto-summarization-quantization}
    \end{subfigure}

    \vspace{2mm}

    \begin{subfigure}[t]{0.485\textwidth}
        \centering
        \includegraphics[width=\linewidth]
        {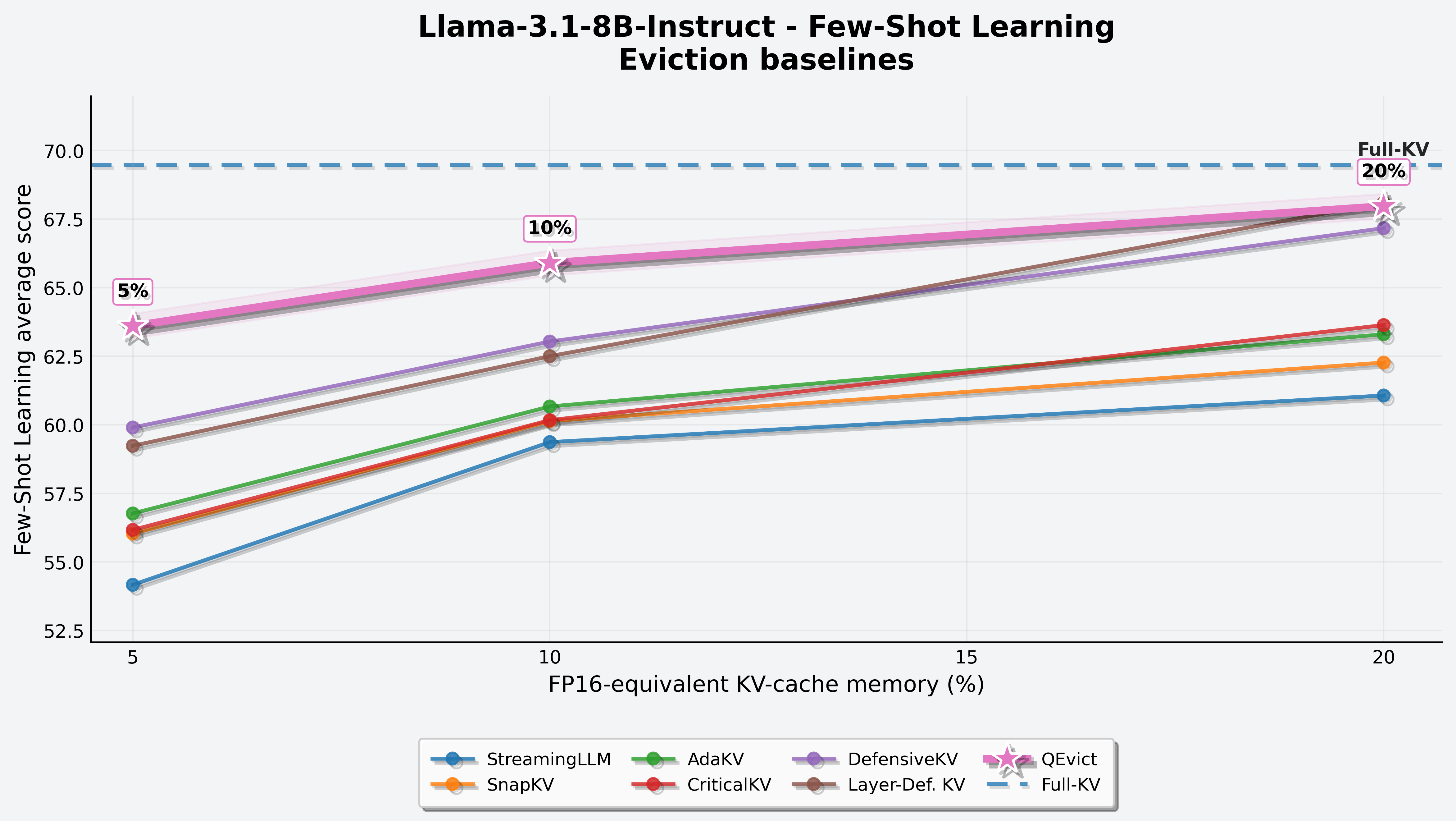}
        \caption{Few-Shot Learning: eviction.}
        \label{fig:llama-pareto-few-shot-eviction}
    \end{subfigure}
    \hfill
    \begin{subfigure}[t]{0.485\textwidth}
        \centering
        \includegraphics[width=\linewidth]
        {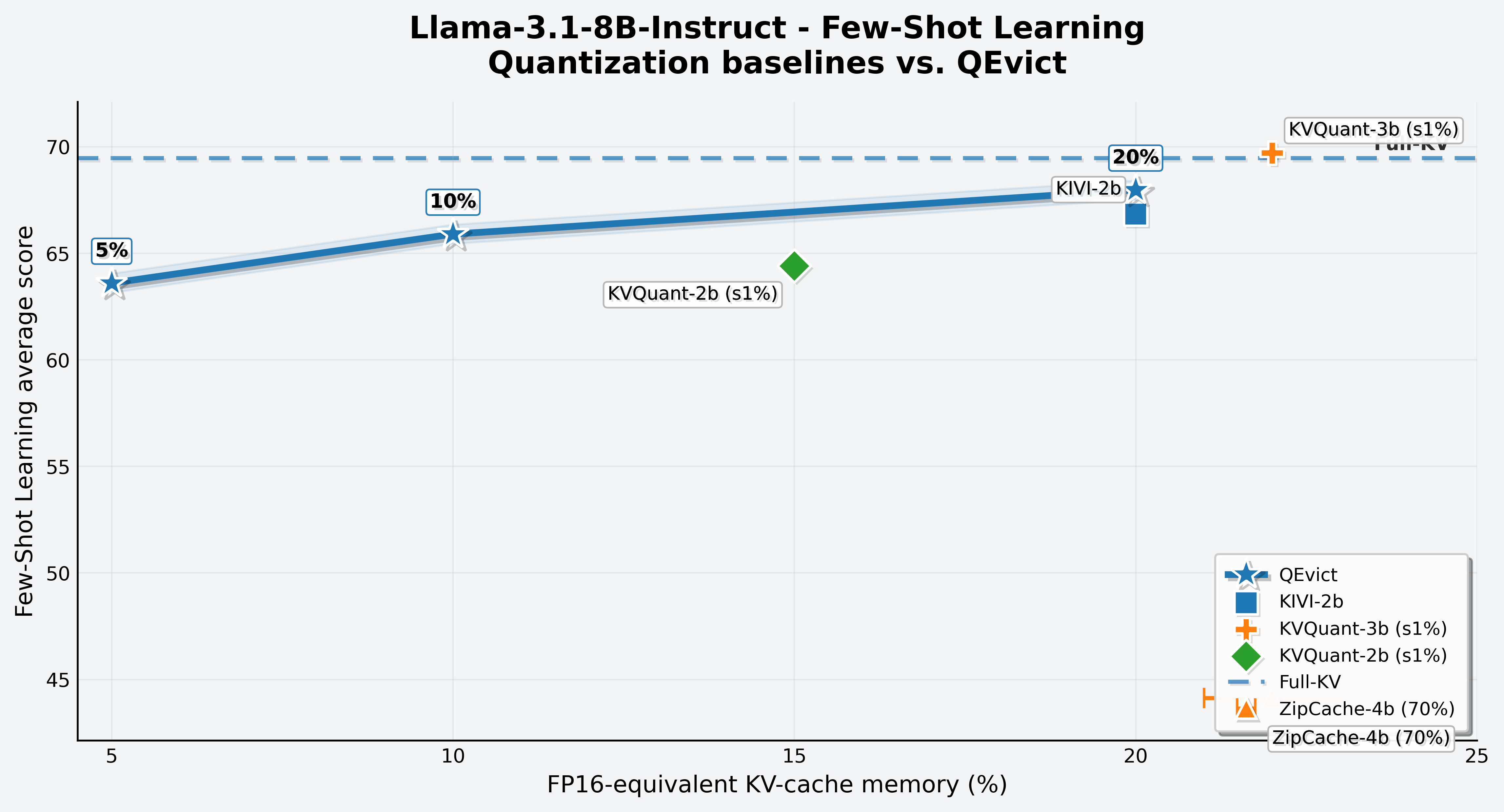}
        \caption{Few-Shot Learning: quantization.}
        \label{fig:llama-pareto-few-shot-quantization}
    \end{subfigure}

    \caption{\textbf{LongBench performance--memory Pareto analysis for Llama-3.1-8B-Instruct.} Rows correspond to Single-Document QA, Multi-Document QA, Summarization, and Few-Shot Learning. The left and right columns compare \myarch{} with eviction and quantization baselines, respectively. Dashed horizontal lines denote \fullkv{} performance.}
    \label{fig:llama-longbench-pareto}
\end{figure*}

\begin{figure*}[p]
    \centering
    \captionsetup[subfigure]{
        font=small,
        labelfont=bf,
        justification=centering,
        skip=2pt
    }

    \begin{subfigure}[t]{0.485\textwidth}
        \centering
        \includegraphics[width=\linewidth]
        {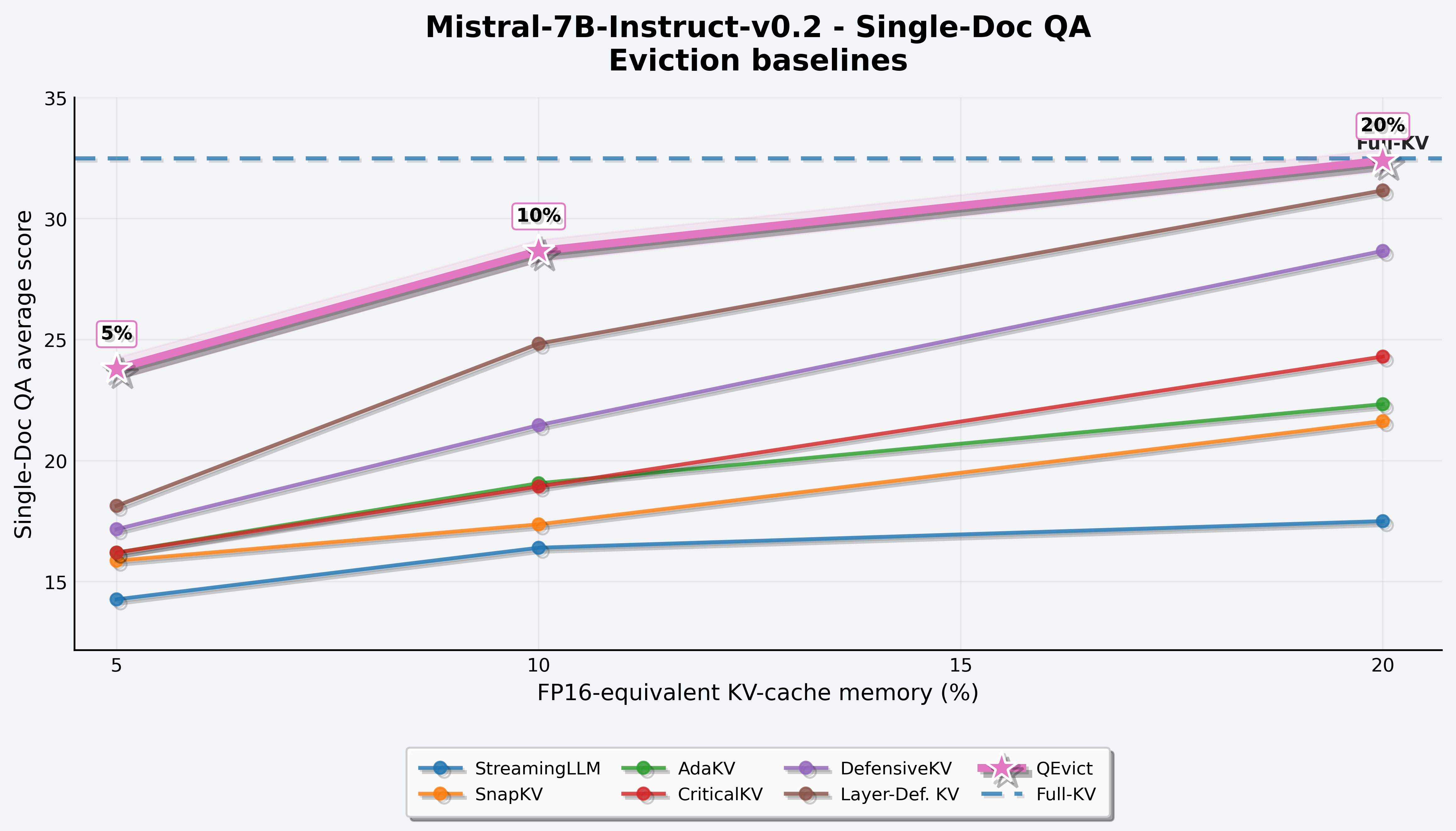}
        \caption{Single-Document QA: eviction.}
        \label{fig:mistral-pareto-single-eviction}
    \end{subfigure}
    \hfill
    \begin{subfigure}[t]{0.485\textwidth}
        \centering
        \includegraphics[width=\linewidth]
        {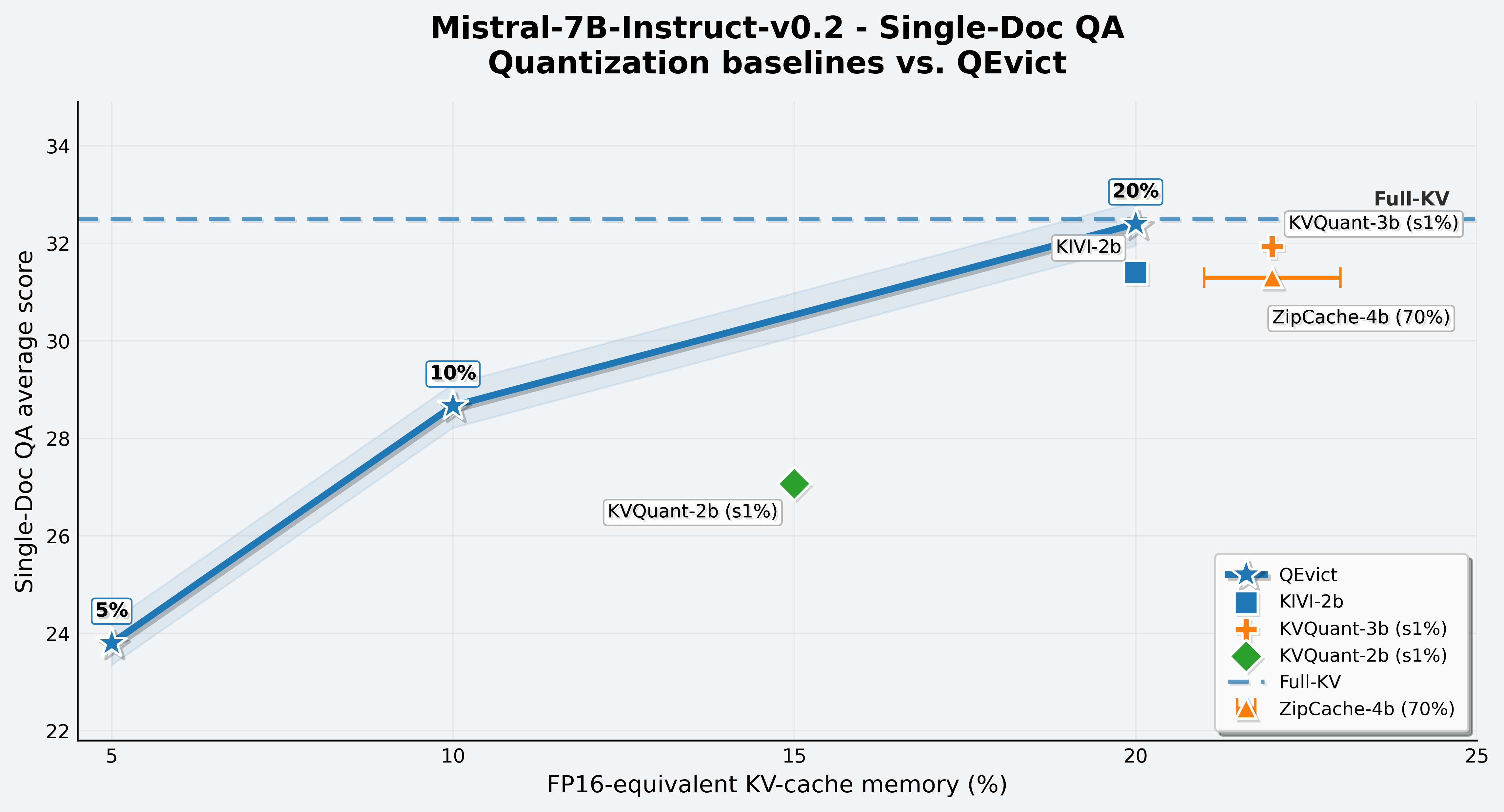}
        \caption{Single-Document QA: quantization.}
        \label{fig:mistral-pareto-single-quantization}
    \end{subfigure}

    \vspace{2mm}

    \begin{subfigure}[t]{0.485\textwidth}
        \centering
        \includegraphics[width=\linewidth]
        {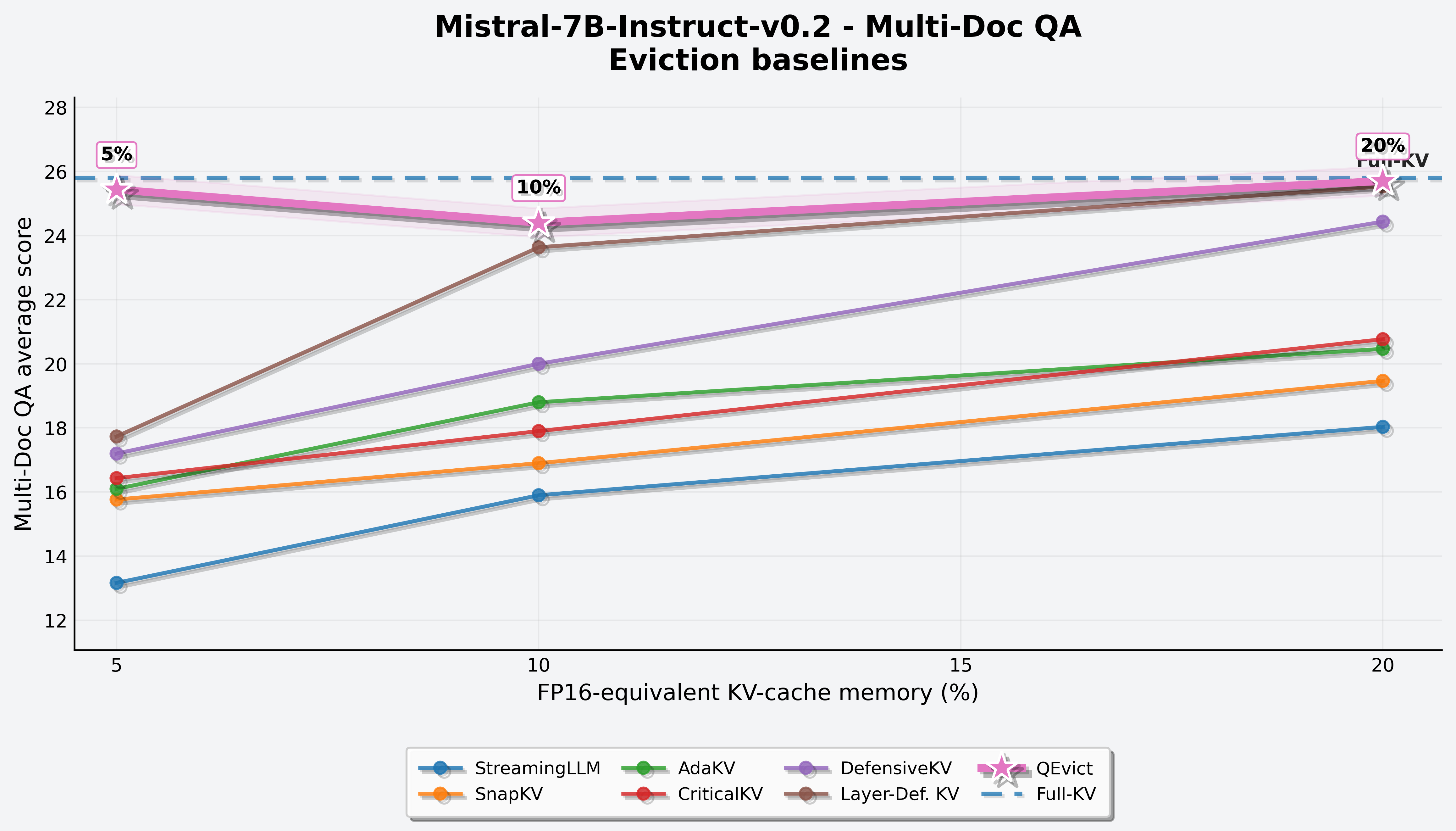}
        \caption{Multi-Document QA: eviction.}
        \label{fig:mistral-pareto-multi-eviction}
    \end{subfigure}
    \hfill
    \begin{subfigure}[t]{0.485\textwidth}
        \centering
        \includegraphics[width=\linewidth]
        {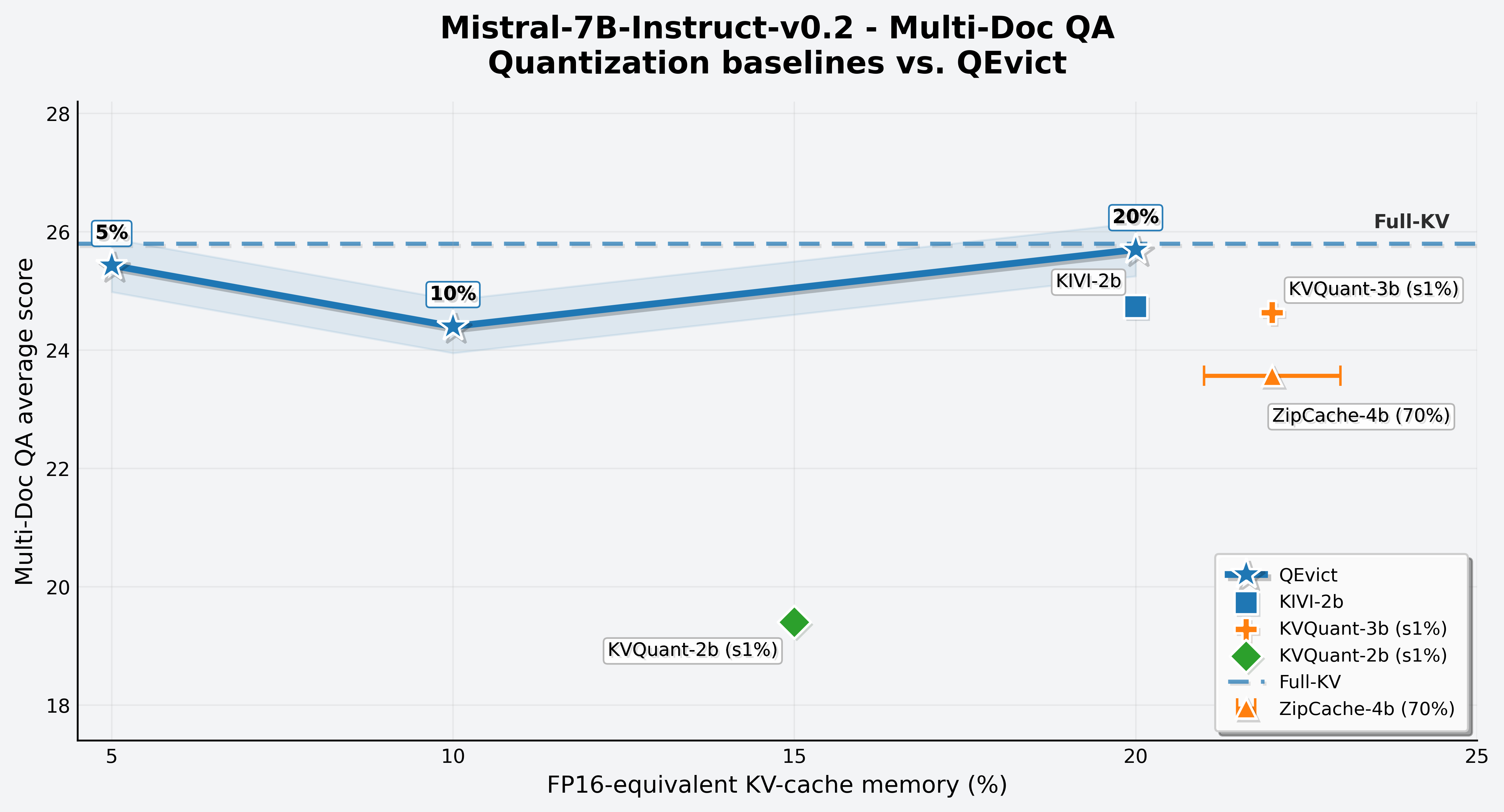}
        \caption{Multi-Document QA: quantization.}
        \label{fig:mistral-pareto-multi-quantization}
    \end{subfigure}

    \vspace{2mm}

    \begin{subfigure}[t]{0.485\textwidth}
        \centering
        \includegraphics[width=\linewidth]
        {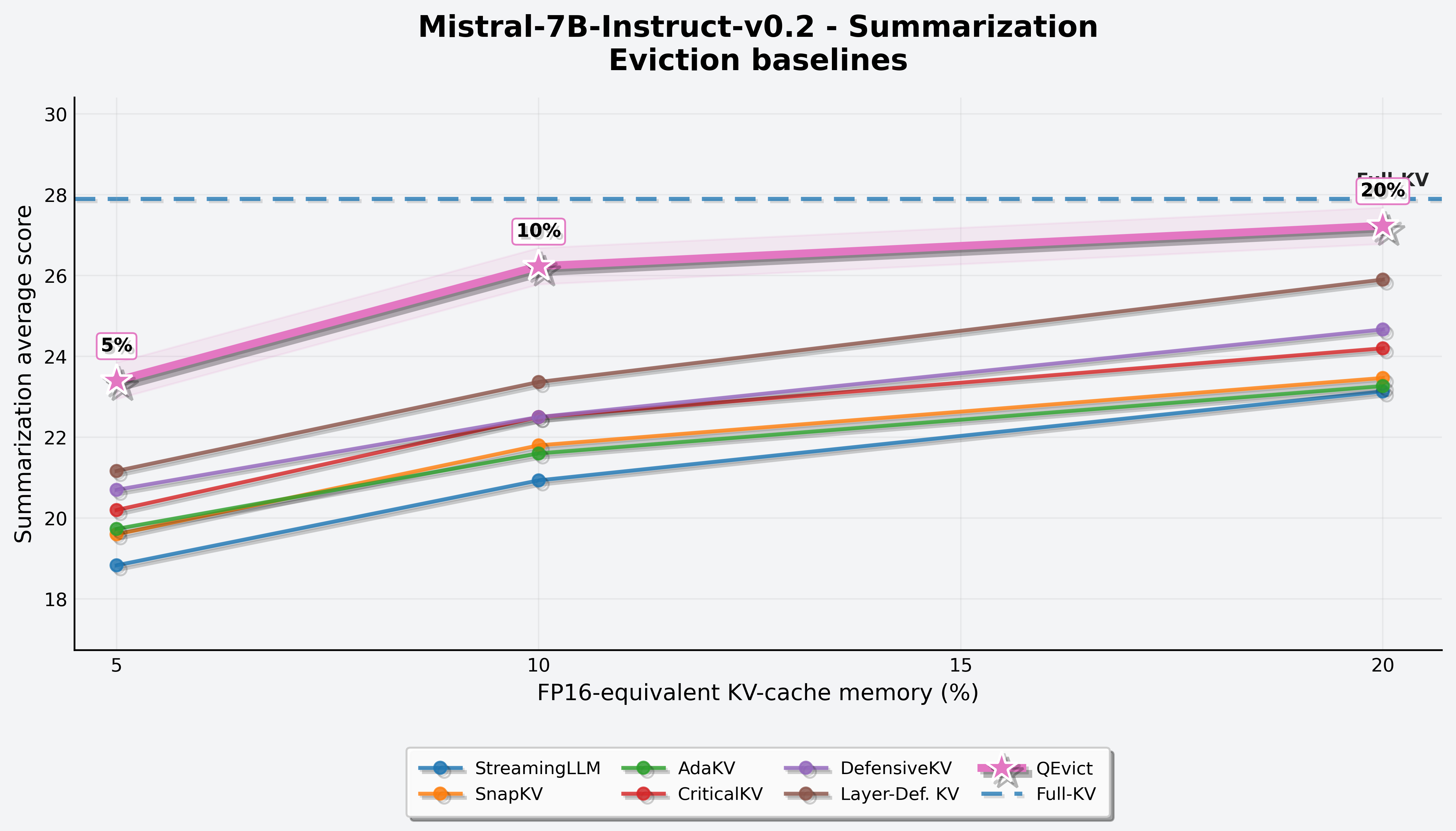}
        \caption{Summarization: eviction.}
        \label{fig:mistral-pareto-summarization-eviction}
    \end{subfigure}
    \hfill
    \begin{subfigure}[t]{0.485\textwidth}
        \centering
        \includegraphics[width=\linewidth]
        {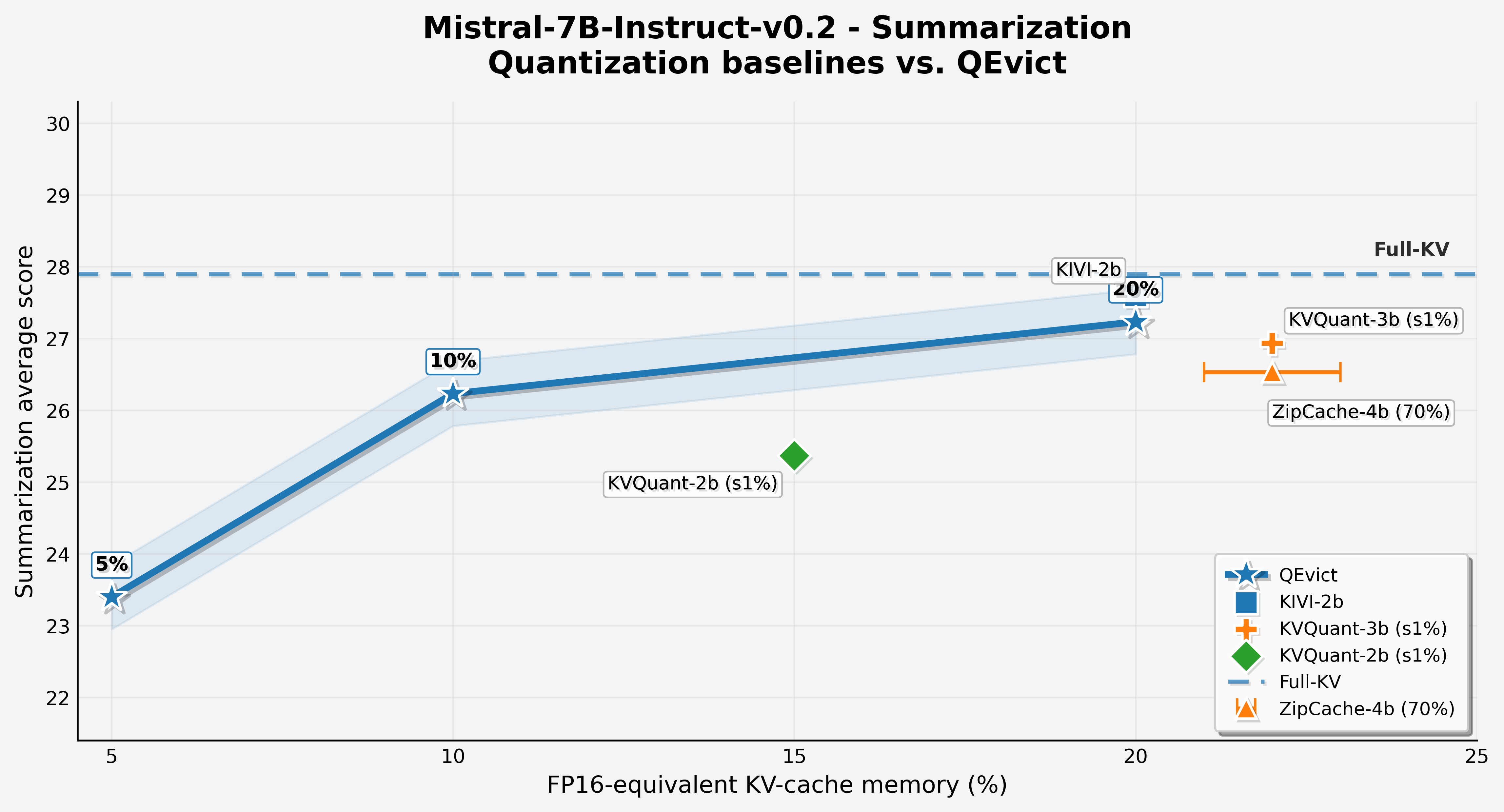}
        \caption{Summarization: quantization.}
        \label{fig:mistral-pareto-summarization-quantization}
    \end{subfigure}

    \vspace{2mm}

    \begin{subfigure}[t]{0.485\textwidth}
        \centering
        \includegraphics[width=\linewidth]
        {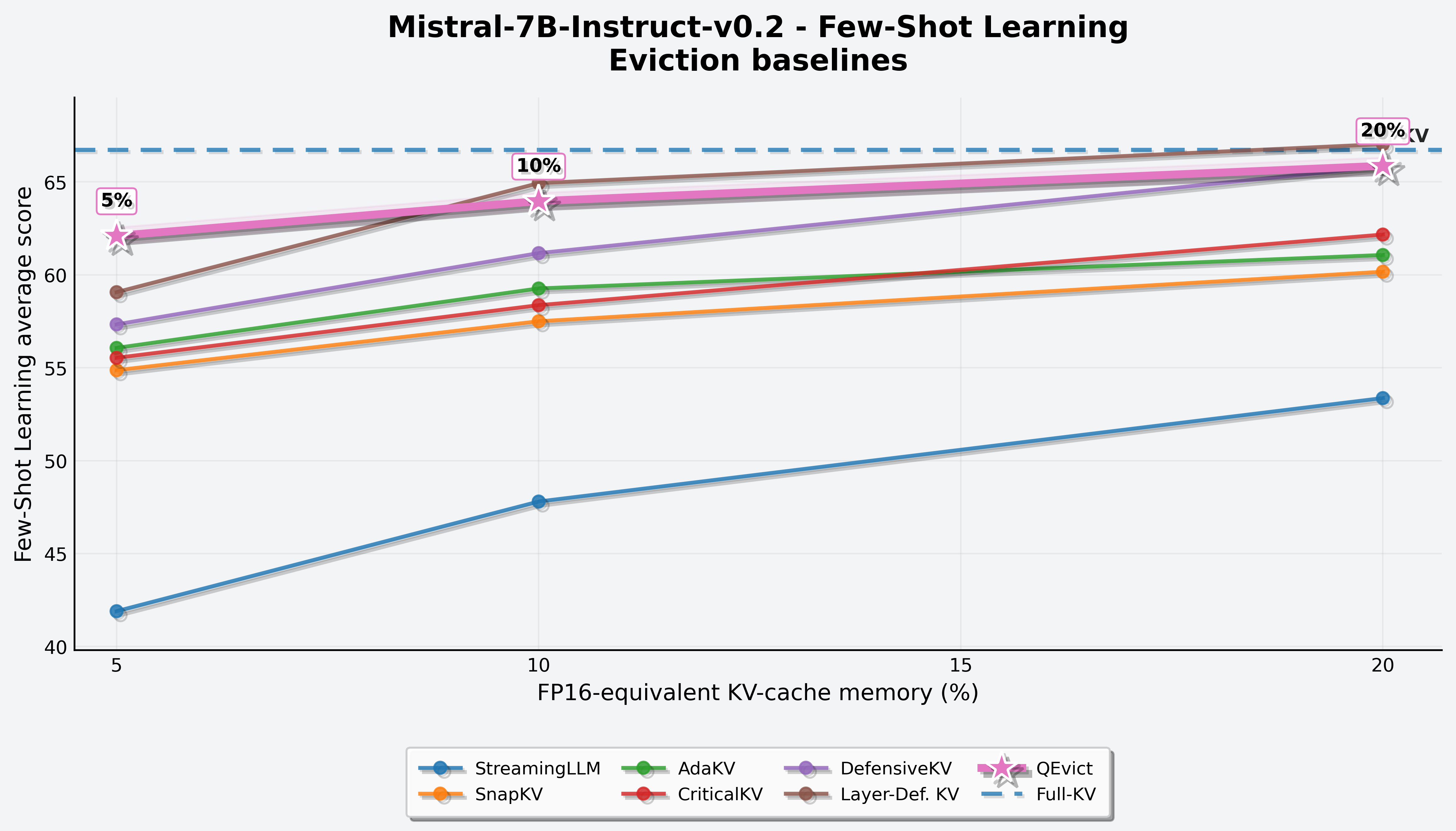}
        \caption{Few-Shot Learning: eviction.}
        \label{fig:mistral-pareto-few-shot-eviction}
    \end{subfigure}
    \hfill
    \begin{subfigure}[t]{0.485\textwidth}
        \centering
        \includegraphics[width=\linewidth]
        {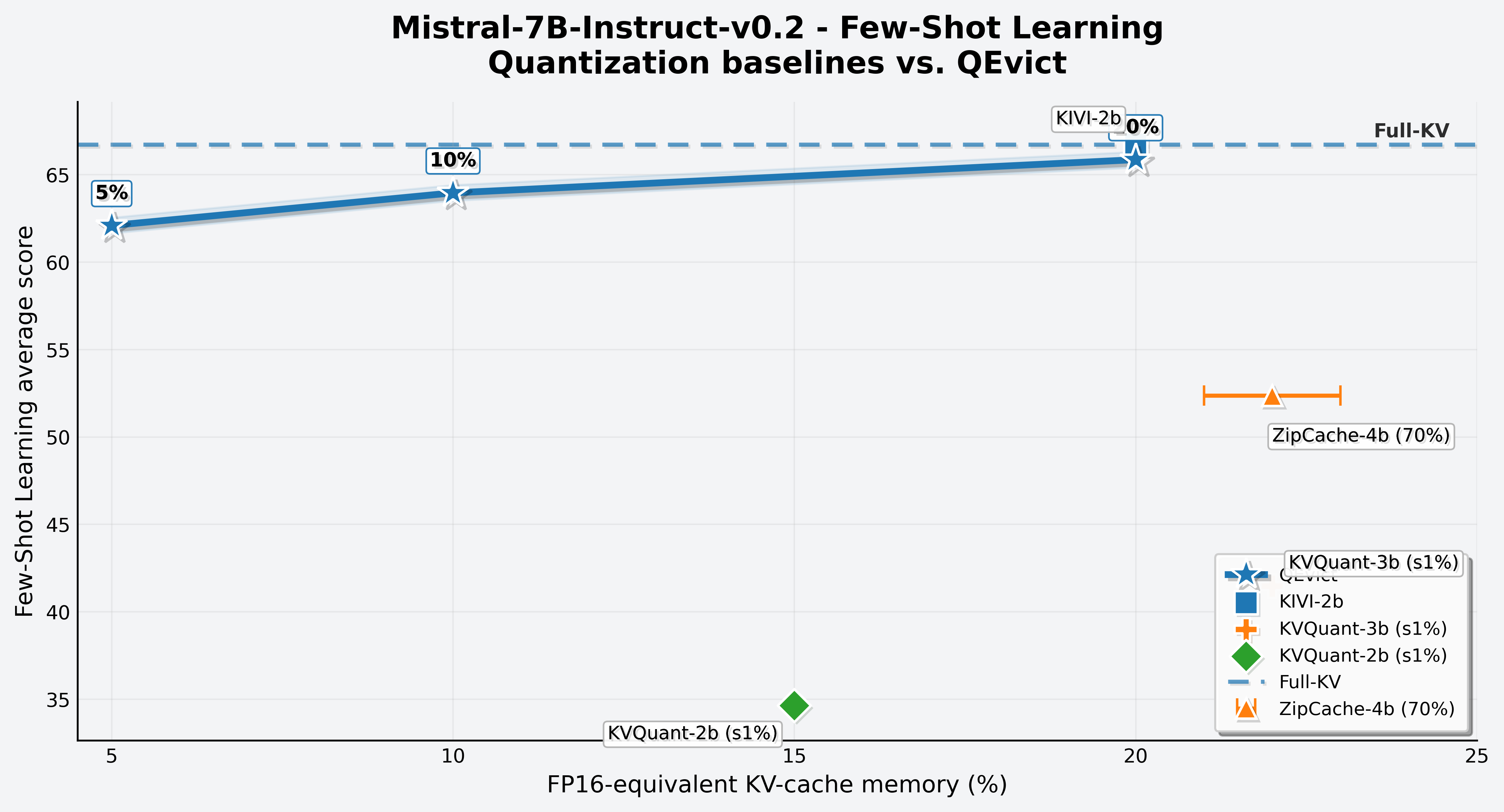}
        \caption{Few-Shot Learning: quantization.}
        \label{fig:mistral-pareto-few-shot-quantization}
    \end{subfigure}

    \caption{\textbf{LongBench performance--memory Pareto analysis for Mistral-7B-Instruct-v0.2.} Rows correspond to Single-Document QA, Multi-Document QA, Summarization, and Few-Shot Learning. The left and right columns compare \myarch{} with eviction and quantization baselines, respectively. Dashed horizontal lines denote \fullkv{} performance.}
    \label{fig:mistral-longbench-pareto}
\end{figure*}

\begin{figure*}[p]
    \centering
    \captionsetup[subfigure]{
        font=small,
        labelfont=bf,
        justification=centering,
        skip=2pt
    }

    \begin{subfigure}[t]{0.485\textwidth}
        \centering
        \includegraphics[width=\linewidth]
        {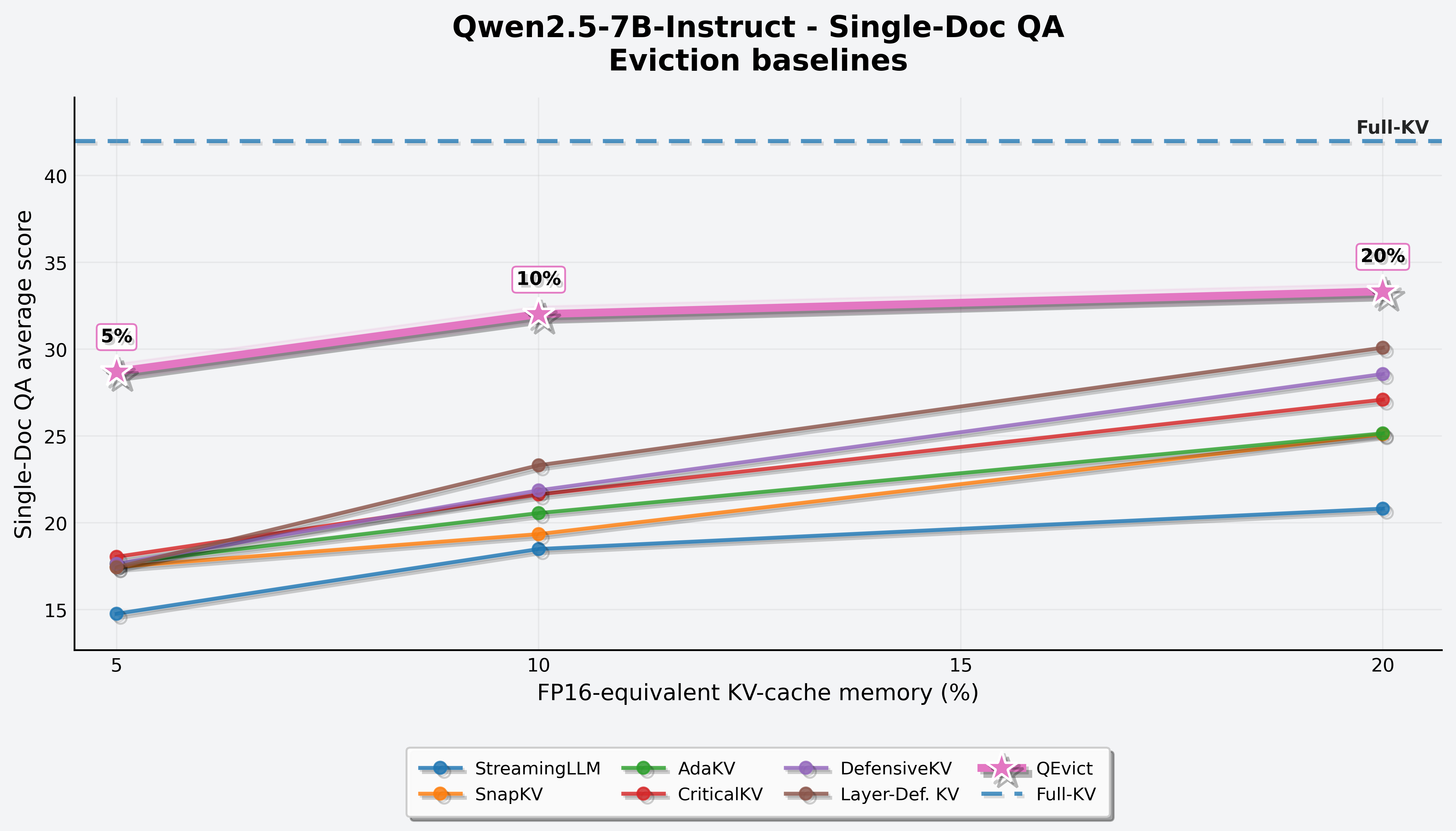}
        \caption{Single-Document QA: eviction.}
        \label{fig:qwen-pareto-single-eviction}
    \end{subfigure}
    \hfill
    \begin{subfigure}[t]{0.485\textwidth}
        \centering
        \includegraphics[width=\linewidth]
        {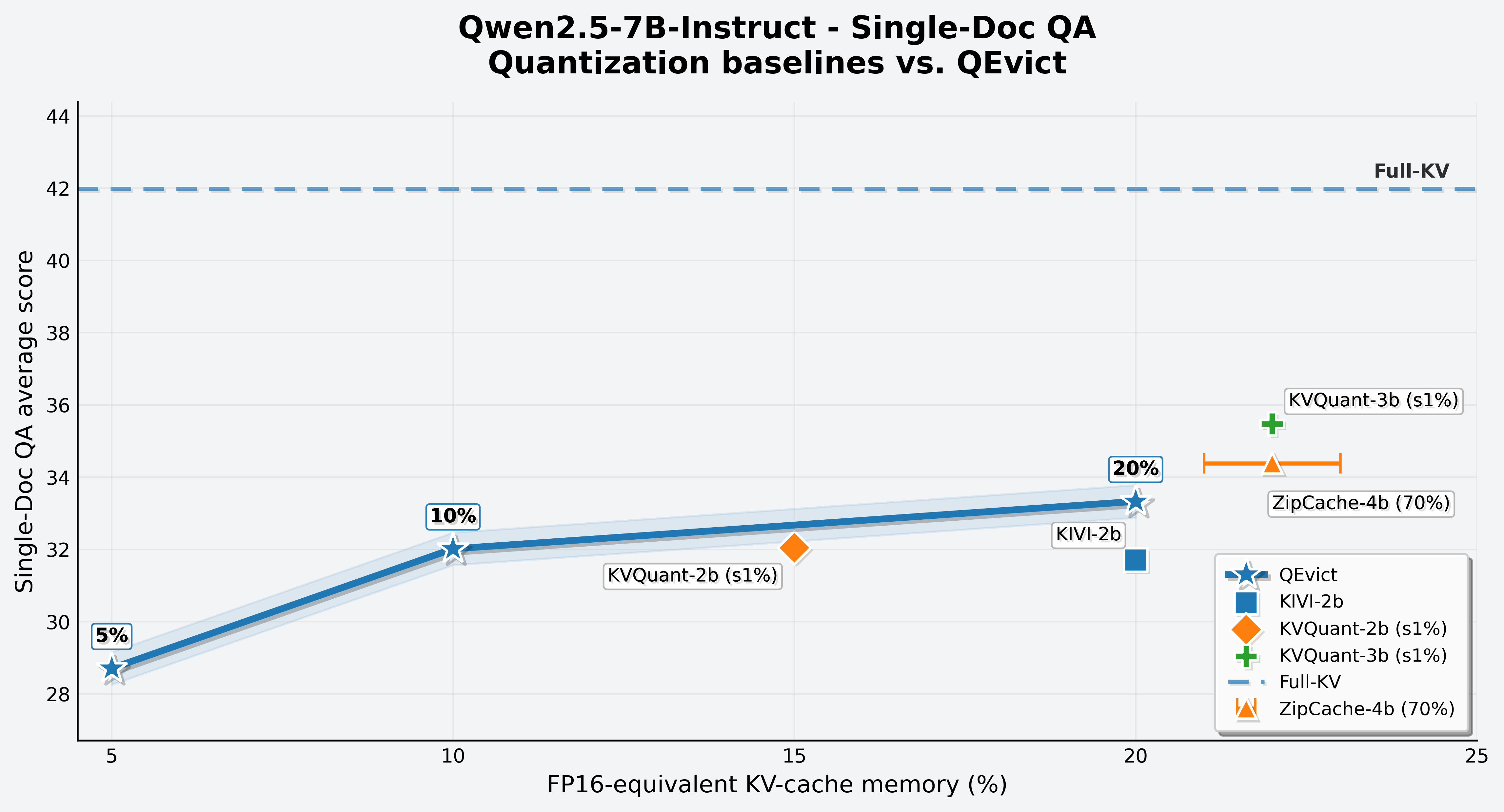}
        \caption{Single-Document QA: quantization.}
        \label{fig:qwen-pareto-single-quantization}
    \end{subfigure}

    \vspace{2mm}

    \begin{subfigure}[t]{0.485\textwidth}
        \centering
        \includegraphics[width=\linewidth]
        {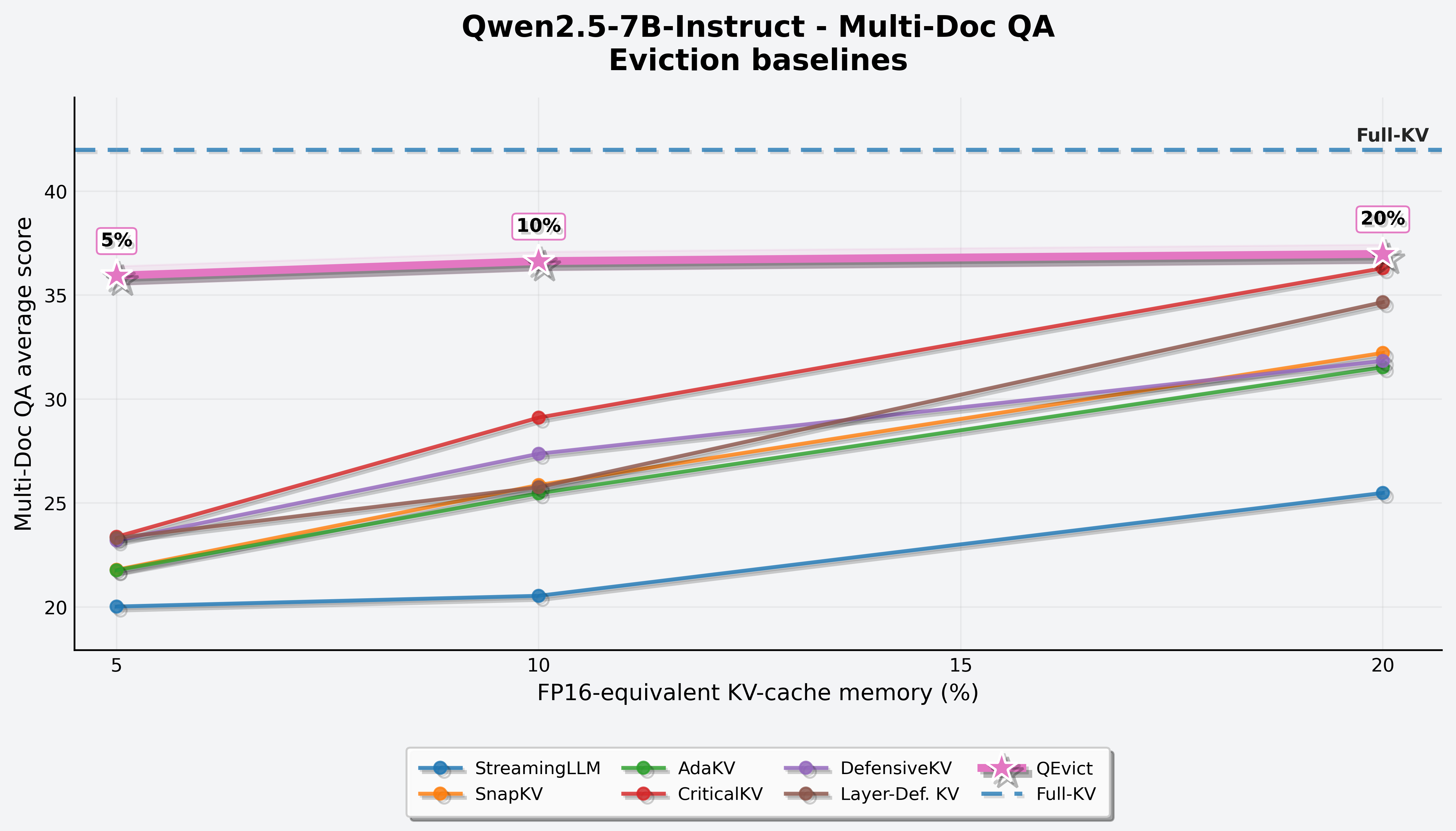}
        \caption{Multi-Document QA: eviction.}
        \label{fig:qwen-pareto-multi-eviction}
    \end{subfigure}
    \hfill
    \begin{subfigure}[t]{0.485\textwidth}
        \centering
        \includegraphics[width=\linewidth]
        {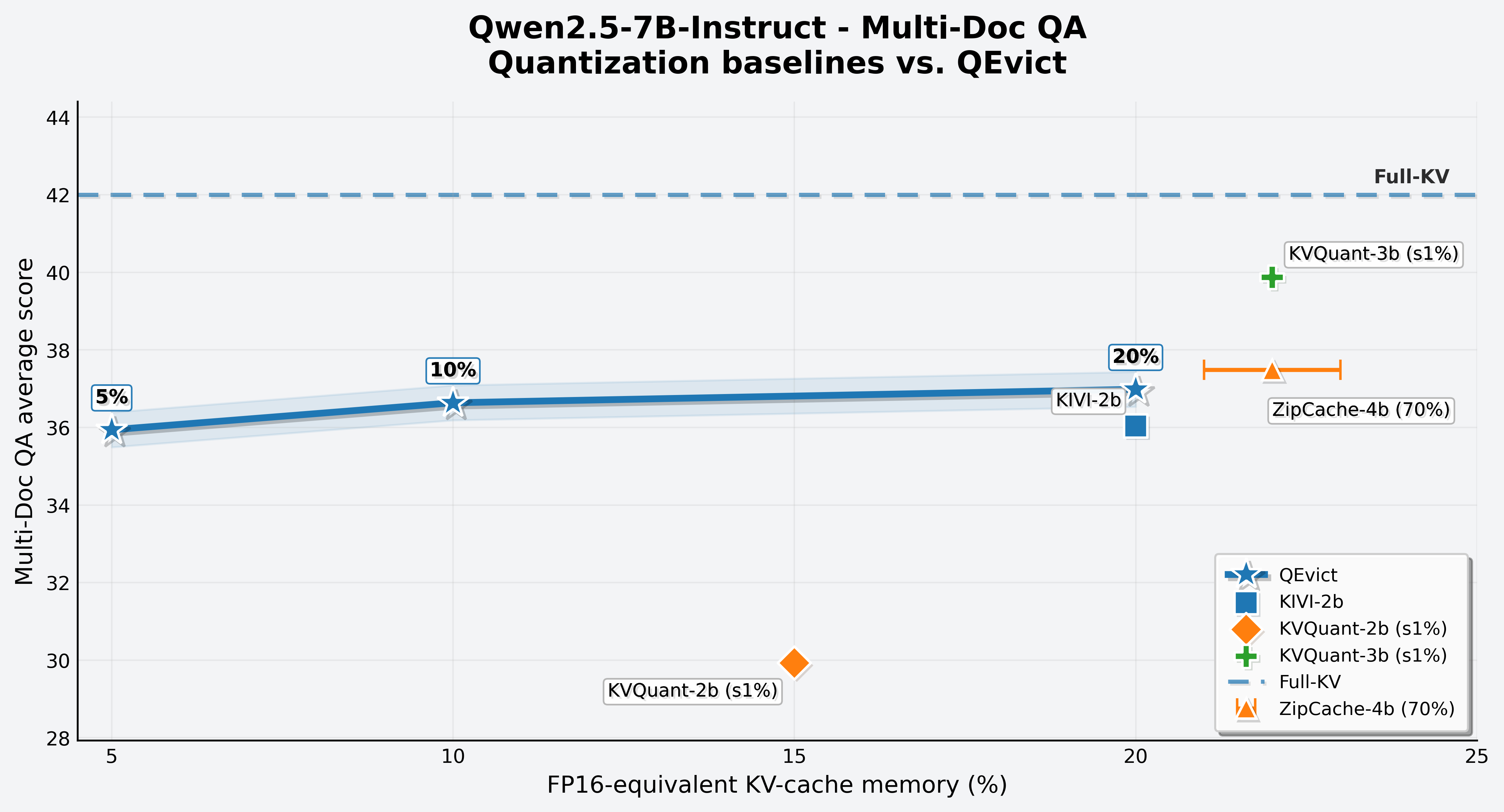}
        \caption{Multi-Document QA: quantization.}
        \label{fig:qwen-pareto-multi-quantization}
    \end{subfigure}

    \vspace{2mm}

    \begin{subfigure}[t]{0.485\textwidth}
        \centering
        \includegraphics[width=\linewidth]
        {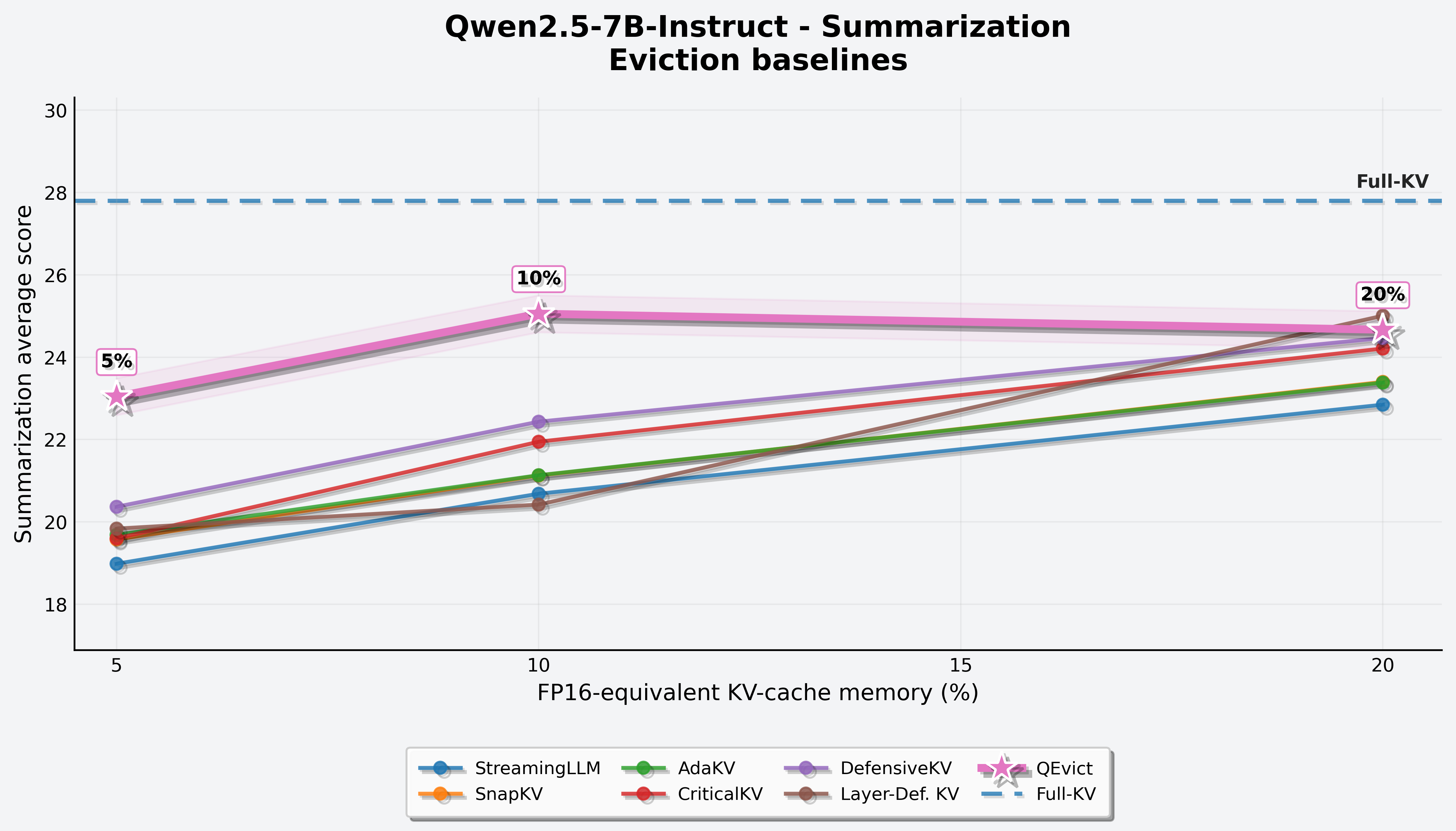}
        \caption{Summarization: eviction.}
        \label{fig:qwen-pareto-summarization-eviction}
    \end{subfigure}
    \hfill
    \begin{subfigure}[t]{0.485\textwidth}
        \centering
        \includegraphics[width=\linewidth]
        {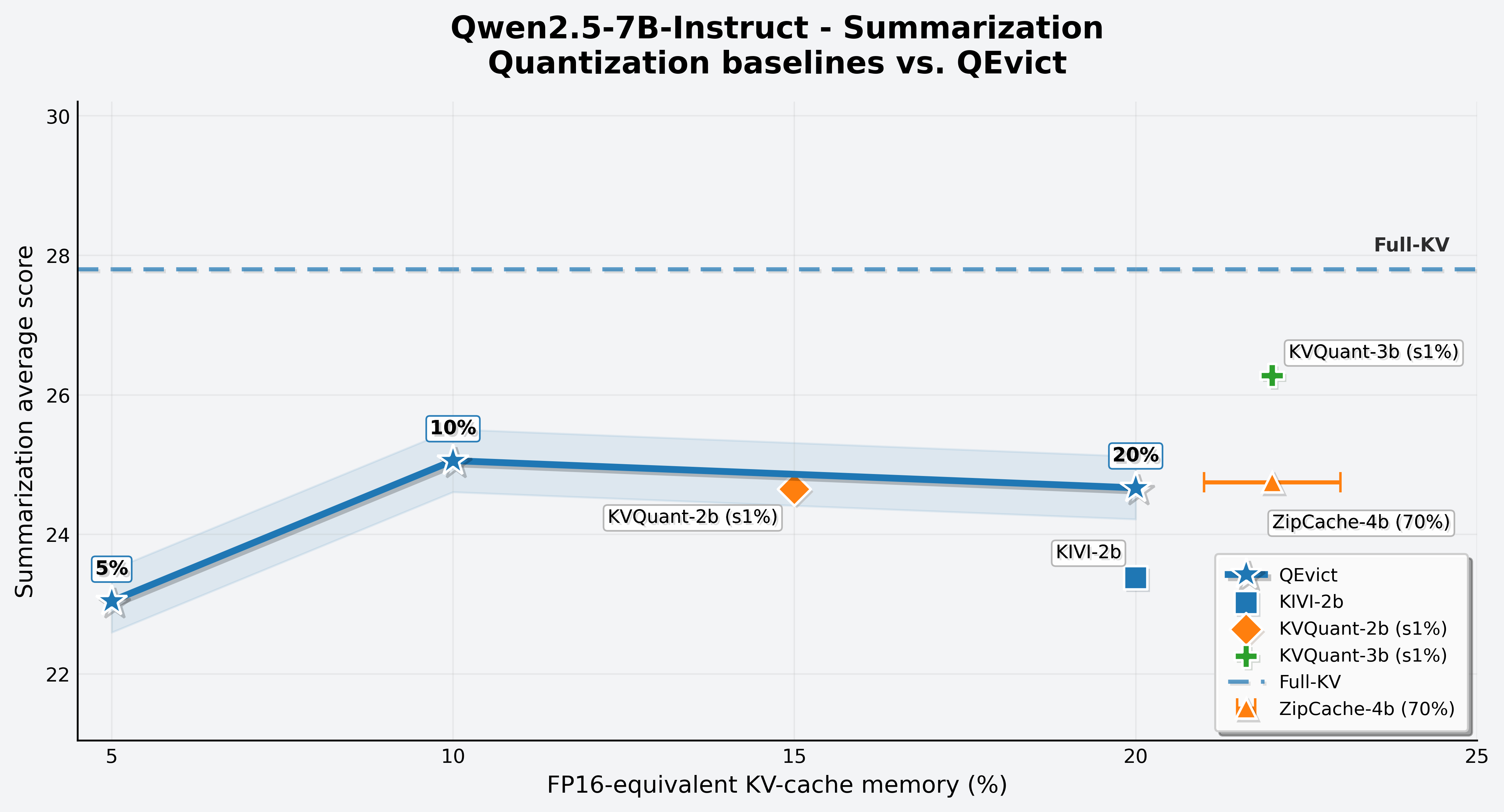}
        \caption{Summarization: quantization.}
        \label{fig:qwen-pareto-summarization-quantization}
    \end{subfigure}

    \vspace{2mm}

    \begin{subfigure}[t]{0.485\textwidth}
        \centering
        \includegraphics[width=\linewidth]
        {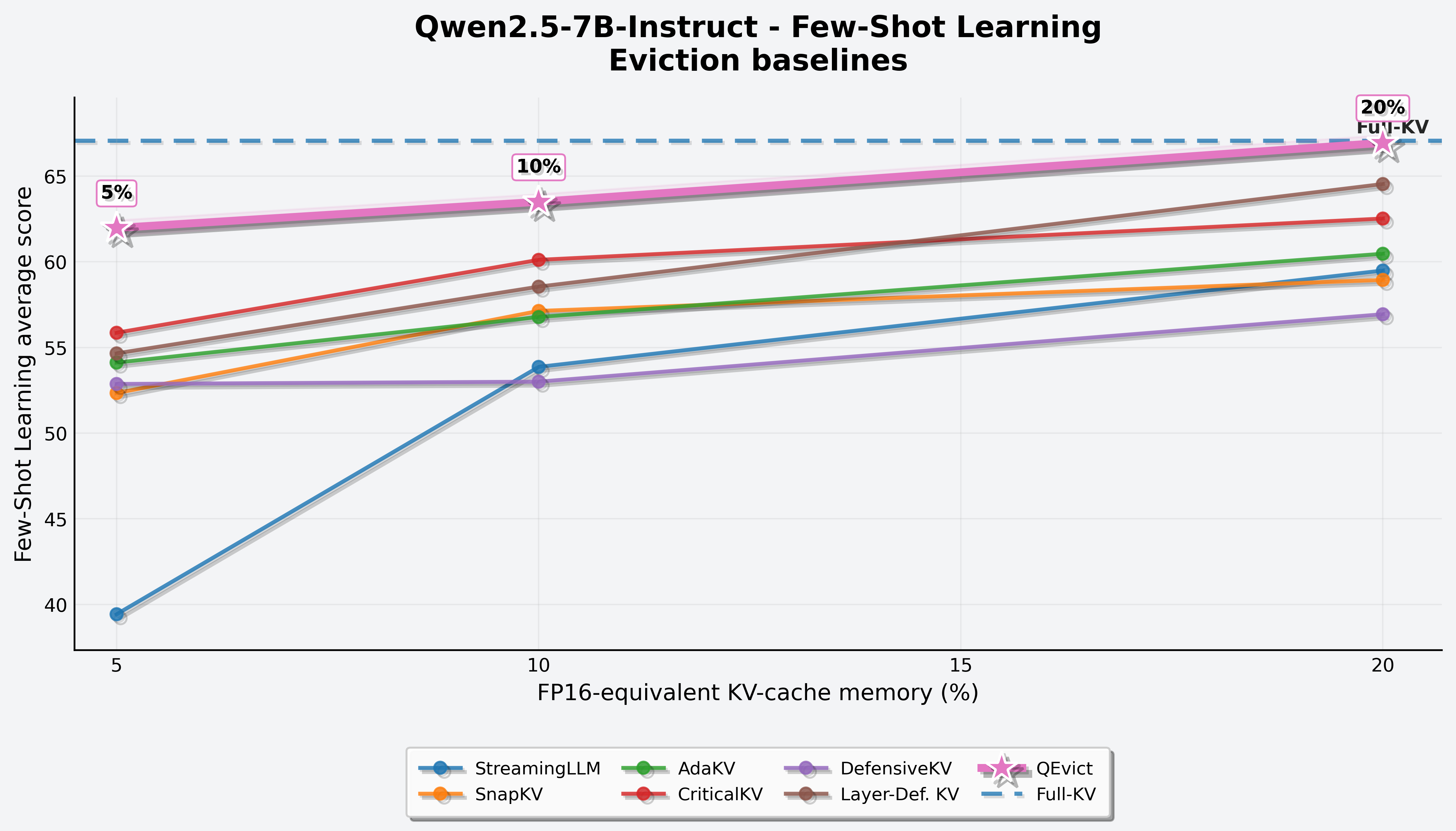}
        \caption{Few-Shot Learning: eviction.}
        \label{fig:qwen-pareto-few-shot-eviction}
    \end{subfigure}
    \hfill
    \begin{subfigure}[t]{0.485\textwidth}
        \centering
        \includegraphics[width=\linewidth]
        {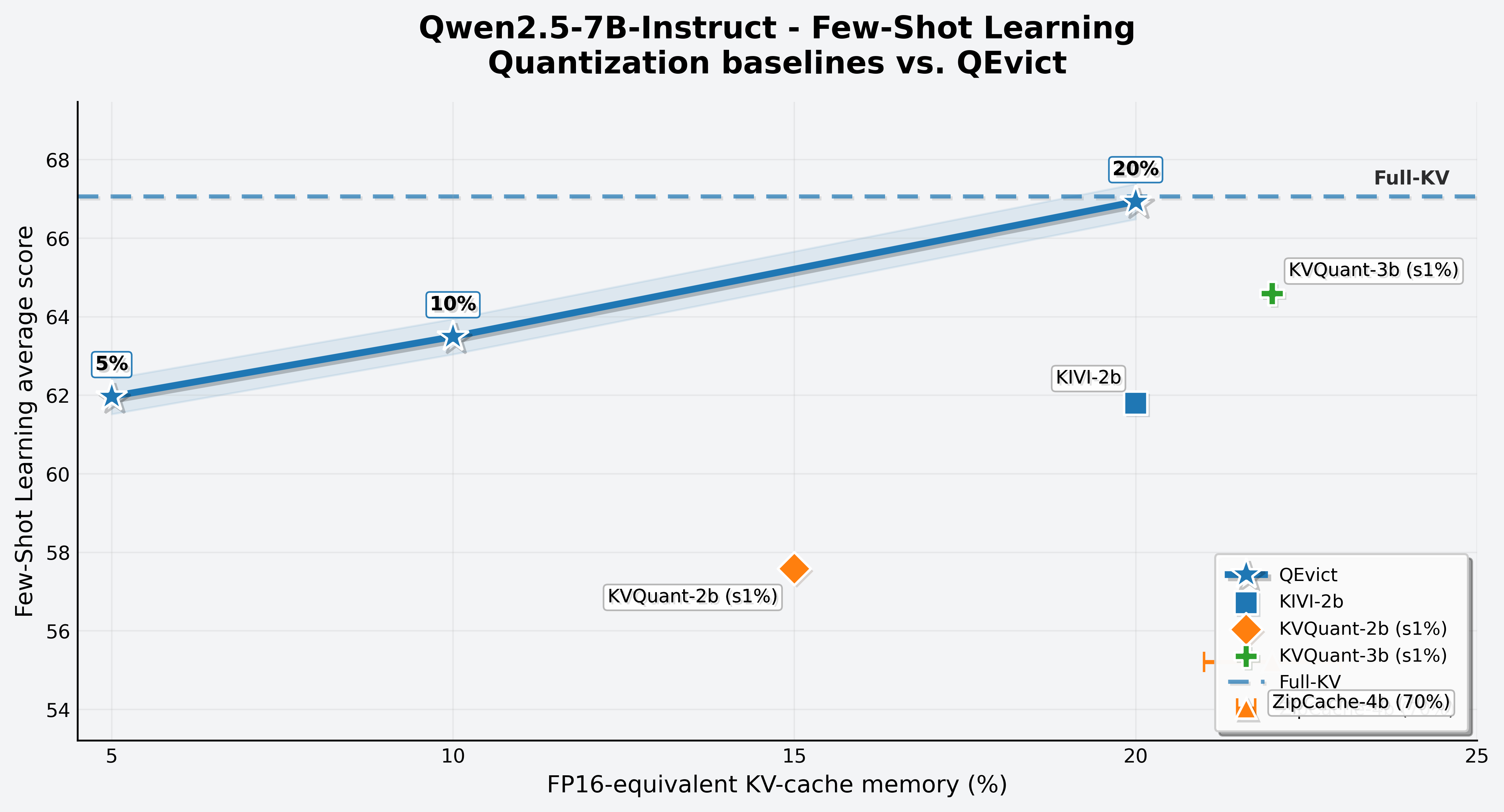}
        \caption{Few-Shot Learning: quantization.}
        \label{fig:qwen-pareto-few-shot-quantization}
    \end{subfigure}

    \caption{\textbf{LongBench performance--memory Pareto analysis for Qwen2.5-7B-Instruct.} Rows correspond to Single-Document QA, Multi-Document QA, Summarization, and Few-Shot Learning. The left and right columns compare \myarch{} with eviction and quantization baselines, respectively. Dashed horizontal lines denote \fullkv{} performance.}
    \label{fig:qwen-longbench-pareto}
\end{figure*}

We construct separate comparisons against eviction and quantization baselines for Single-Document QA, Multi-Document QA, Summarization, and Few-Shot Learning. This produces eight panels per model and 24 panels across Llama-3.1-8B-Instruct, Mistral-7B-Instruct-v0.2, and Qwen2.5-7B-Instruct. Figures~\ref{fig:llama-longbench-pareto}, \ref{fig:mistral-longbench-pareto}, and \ref{fig:qwen-longbench-pareto} report the corresponding Pareto plots. The left column of each figure compares \myarch{} with eviction methods, while the right column compares it with quantization methods. Horizontal dashed lines indicate the performance of the uncompressed \fullkv{} reference.

These plots complement the fixed-budget tables by showing whether an advantage persists across multiple memory regimes rather than at a single selected budget. They also distinguish methods that improve performance at matched memory from those that reduce memory at comparable performance.

% ============================================================
% LLAMA
% ============================================================

% ============================================================
% MISTRAL
% ============================================================

% ============================================================
% QWEN
% ============================================================

% ============================================================
% RULER
% ============================================================

\section{RULER Evaluation Details}
\label{app:ruler}

\subsection{Dataset and Evaluation Protocol}
\label{app:ruler_protocol}

RULER~\citep{hsieh2024ruler} provides a controlled evaluation of long-context retrieval, aggregation, and reasoning by varying the location, multiplicity, and dependency structure of relevant evidence. We evaluate Llama-3.1-8B-Instruct at a \(32\mathrm{K}\) context length under a \(20\%\) KV-cache budget. The complete results for this setting are reported in the main paper in Table~\ref{tab:ruler_32k_combined_llama}. This section provides additional details about the benchmark and task composition. For RULER, we use a protected recent region of 64 tokens.

RULER contains 13 tasks spanning four complementary capabilities, summarized in Table~\ref{tab:ruler_task_families}. These task families expose complementary failure modes of KV-cache compression. Needle-in-a-haystack tasks test whether sparse evidence remains accessible when its relevance is revealed only at query time. Aggregation requires retaining information distributed across the sequence, while variable tracking evaluates the preservation of dependent intermediate states. The question-answering tasks provide a more naturalistic assessment of long-range evidence use.

\begin{table*}[t]
\centering
\caption{\textbf{RULER task families used in our evaluation.}}
\label{tab:ruler_task_families}
\small
\setlength{\tabcolsep}{5pt}
\renewcommand{\arraystretch}{1.12}

\begin{tabularx}{\textwidth}{
@{}
>{\raggedright\arraybackslash\bfseries}p{0.18\textwidth}
>{\raggedright\arraybackslash}p{0.32\textwidth}
>{\raggedright\arraybackslash}X
@{}
}
\toprule
\textbf{Task Family}
& \textbf{Tasks}
& \textbf{Evaluation Focus} \\
\midrule

Aggregation
& CWE, FWE
& Aggregating frequency information distributed throughout the context. \\
\addlinespace[2pt]

Needle-in-a-Haystack
& S-1--S-3, MK-1--MK-3, MQ, MV
& Retrieving sparse evidence under varying numbers of keys, values, and
queries. \\
\addlinespace[2pt]

Question Answering
& QA-1, QA-2
& Retrieving and combining evidence from relevant passages. \\
\addlinespace[2pt]

Variable Tracking
& VT
& Preserving and following dependent variable assignments across the
sequence. \\

\bottomrule
\end{tabularx}
\end{table*}

% ============================================================
% GSM8K
% ============================================================

\section{GSM8K Evaluation}
\label{app:gsm8k}

\subsection{Evaluation Protocol}
\label{app:gsm8k_protocol}

GSM8K~\citep{mirzadeh2024gsm} evaluates KV-cache compression during autoregressive multi-step reasoning, complementing long-context benchmarks that primarily test retrieval from the input sequence. Successful generation requires maintaining intermediate decoding states that may become relevant again during later calculations or final-answer construction.

We evaluate Llama-3.1-8B-Instruct~\citep{dubey2024llama3}, Qwen2.5-7B-Instruct~\citep{yang2025qwen3}, and Mistral-7B-Instruct-v0.2~\citep{jiang2023mistral} across multiple measured KV-memory budgets. All methods use the same prompts, greedy decoding procedure, generation limits, and answer-extraction protocol. Performance is reported as exact-match accuracy, and memory is measured relative to the \fullkv{} cache. Eviction methods are compared at matched memory budgets, while quantization methods are reported at their measured operating points following Appendix~\ref{app:comparison_regimes}. Because GSM8K sequences are relatively short, we reserve 25\% of the total persistent KV-cache budget for recent tokens.

Fig~\ref{fig:gsm8k_budget_sweep} in the main paper presents the accuracy–memory curves for all three models. Tables~\ref{tab:gsm8k_llama31_8b},\ref{tab:gsm8k_mistral7b},\ref{tab:gsm8k_qwen7b} report the corresponding numerical results for every evaluated operating point.

% Required packages:
% \usepackage{booktabs}
% \usepackage{multirow}
% \usepackage{subcaption}
% \usepackage[table]{xcolor}

\providecommand{\lbbest}[1]{\textbf{#1}}
\providecommand{\lbsecond}[1]{\underline{#1}}

% Define these colors only if they are not already defined.
\providecolor{fullkvgray}{gray}{0.92}
\providecolor{qevictyellow}{RGB}{255,247,204}

% ============================================================
% Llama-3.1-8B-Instruct
% ============================================================
\begin{table*}[!h]
\centering

\caption{\textbf{GSM8K performance on Llama-3.1-8B-Instruct.}
The left subtable compares KV-cache eviction methods at different
KV-cache budgets. The right subtable compares KV-cache quantization
methods at approximately matched memory.}

\label{tab:gsm8k_llama31_8b}

\begingroup
\normalsize
\renewcommand{\arraystretch}{1.15}

% ------------------------------------------------------------
% Left: KV-cache eviction
% ------------------------------------------------------------
\begin{subtable}[t]{0.66\textwidth}
\centering
\caption{KV-cache eviction}
\label{tab:gsm8k_llama31_8b_eviction}

\setlength{\tabcolsep}{3.5pt}

\begin{tabular*}{\linewidth}{
    @{\extracolsep{\fill}}
    lccccc
    @{}
}
\toprule

\multirow{2}{*}{\textbf{Method}} &
\multicolumn{5}{c}{\textbf{KV-Cache Budget}} \\

\cmidrule(lr){2-6}

&
\textbf{10\%} &
\textbf{20\%} &
\textbf{40\%} &
\textbf{60\%} &
\textbf{80\%} \\

\midrule

\rowcolor{fullkvgray}
Full-KV
& 86.93
& 86.93
& 86.93
& 86.93
& 86.93 \\

\midrule

DefensiveKV
& 3.11
& 11.52
& 76.04
& 81.65
& \lbsecond{84.76} \\

Layer-Def.\ KV
& 3.18
& \lbsecond{31.16}
& \lbsecond{82.56}
& \lbsecond{84.08}
& 84.23 \\

CriticalKV
& \lbsecond{3.26}
& 6.97
& 31.24
& 59.14
& 81.88 \\

SnapKV
& 1.90
& 4.09
& 21.61
& 50.95
& 76.95 \\

AdaKV
& 2.20
& 5.16
& 23.97
& 52.77
& 72.82 \\

StreamingLLM
& 1.97
& 3.56
& 66.79
& 82.87
& 84.46 \\

\rowcolor{qevictyellow}
\textbf{\myarch{}}
& \lbbest{11.45}
& \lbbest{79.09}
& \lbbest{85.44}
& \lbbest{85.97}
& \lbbest{85.32} \\

\bottomrule
\end{tabular*}
\end{subtable}
\hfill
% ------------------------------------------------------------
% Right: KV-cache quantization
% ------------------------------------------------------------
\begin{subtable}[t]{0.31\textwidth}
\centering
\caption{KV-cache quantization}
\label{tab:gsm8k_llama31_8b_quantization}

\setlength{\tabcolsep}{5pt}

\begin{tabular*}{\linewidth}{
    @{\extracolsep{\fill}}
    lcc
    @{}
}
\toprule

\textbf{Method} &
\shortstack{\textbf{Equivalent}\\\textbf{KV Memory}} &
\textbf{Score} \\

\midrule

\rowcolor{fullkvgray}
Full-KV
& 100\%
& 86.93 \\

\midrule

% \rowcolor{qevictyellow}
% \textbf{\myarch{}}
% & 10\%
% & 11.45 \\

\rowcolor{qevictyellow}
\textbf{\myarch{}}
& 20\%
& 79.09 \\

KIVI-2b
& $\sim$20\%
& 79.98 \\

ZipCache-4b
& $\sim$22\%
& \lbbest{84.46} \\

KVQuant-2b
& $\sim$15\%
& 76.61 \\

KVQuant-3b
& $\sim$22\%
& \lbsecond{83.18} \\

\bottomrule
\end{tabular*}
\end{subtable}

\endgroup
\end{table*}

% ============================================================
% Mistral-7B-Instruct-v0.2
% ============================================================
\begin{table*}[!h]
\centering

\caption{\textbf{GSM8K performance on Mistral-7B-Instruct-v0.2.} The left subtable compares KV-cache eviction methods at different KV-cache budgets. The right subtable compares KV-cache quantization methods at approximately matched memory footprints. Full-KV is excluded from the best and second-best rankings.}

\label{tab:gsm8k_mistral7b}

\begingroup
\normalsize
\renewcommand{\arraystretch}{1.15}

% ------------------------------------------------------------
% Left: KV-cache eviction
% ------------------------------------------------------------
\begin{subtable}[t]{0.66\textwidth}
\centering
\caption{KV-cache eviction}
\label{tab:gsm8k_mistral7b_eviction}

\setlength{\tabcolsep}{3.5pt}

\begin{tabular*}{\linewidth}{
    @{\extracolsep{\fill}}
    lccccc
    @{}
}
\toprule

\multirow{2}{*}{\textbf{Method}} &
\multicolumn{5}{c}{\textbf{KV-Cache Budget}} \\

\cmidrule(lr){2-6}

&
\textbf{10\%} &
\textbf{20\%} &
\textbf{40\%} &
\textbf{60\%} &
\textbf{80\%} \\

\midrule

\rowcolor{fullkvgray}
Full-KV
& 46.17
& 46.17
& 46.17
& 46.17
& 46.17 \\

\midrule

DefensiveKV
& \lbsecond{2.35}
& 10.92
& \lbsecond{39.88}
& 42.68
& \lbbest{43.87} \\

Layer-Def.\ KV
& 1.90
& \lbsecond{20.09}
& 39.80
& \lbbest{43.79}
& \lbsecond{43.72} \\

CriticalKV
& 2.05
& 5.23
& 11.22
& 21.99
& 36.16 \\

SnapKV
& 1.52
& 2.73
& 7.88
& 18.50
& 32.83 \\

AdaKV
& 1.52
& 3.41
& 10.77
& 19.94
& 30.40 \\

StreamingLLM
& 1.59
& 1.97
& 29.57
& 40.26
& 42.23 \\

\rowcolor{qevictyellow}
\textbf{\myarch{}}
& \lbbest{18.92}
& \lbbest{40.12}
& \lbbest{43.67}
& \lbsecond{43.58}
& 43.44 \\

\bottomrule
\end{tabular*}
\end{subtable}
\hfill
% ------------------------------------------------------------
% Right: KV-cache quantization
% ------------------------------------------------------------
\begin{subtable}[t]{0.31\textwidth}
\centering
\caption{KV-cache quantization}
\label{tab:gsm8k_mistral7b_quantization}

\setlength{\tabcolsep}{5pt}

\begin{tabular*}{\linewidth}{
    @{\extracolsep{\fill}}
    lcc
    @{}
}
\toprule

\textbf{Method} &
\shortstack{\textbf{Equivalent}\\\textbf{KV Memory}} &
\textbf{Score} \\

\midrule

\rowcolor{fullkvgray}
Full-KV
& 100\%
& 46.17 \\

\midrule

% \rowcolor{qevictyellow}
% \textbf{\myarch{}}
% & 10\%
% & 18.92 \\

\rowcolor{qevictyellow}
\textbf{\myarch{}}
& 20\%
& 40.12 \\

KIVI-2b
& $\sim$20\%
& \lbsecond{40.94} \\

ZipCache-4b
& $\sim$22\%
& 40.64 \\

KVQuant-2b
& $\sim$15\%
& 38.12 \\

KVQuant-3b
& $\sim$22\%
& \lbbest{41.79} \\

\bottomrule
\end{tabular*}
\end{subtable}

\endgroup
\end{table*}

% ============================================================
% Qwen-7B
% ============================================================
\begin{table*}[!h]
\centering

\caption{\textbf{GSM8K performance on Qwen2.5-7B-Instruc.}
The left subtable compares KV-cache eviction methods at different KV-cache budgets. The right subtable compares KV-cache quantization methods at approximately matched memory footprints. Full-KV is excluded from the best and second-best rankings.}

\label{tab:gsm8k_qwen7b}

\begingroup
\normalsize
\renewcommand{\arraystretch}{1.15}

% ------------------------------------------------------------
% Left: KV-cache eviction
% ------------------------------------------------------------
\begin{subtable}[t]{0.66\textwidth}
\centering
\caption{KV-cache eviction}
\label{tab:gsm8k_qwen7b_eviction}

\setlength{\tabcolsep}{3.5pt}

\begin{tabular*}{\linewidth}{
    @{\extracolsep{\fill}}
    lccccc
    @{}
}
\toprule

\multirow{2}{*}{\textbf{Method}} &
\multicolumn{5}{c}{\textbf{KV-Cache Budget}} \\

\cmidrule(lr){2-6}

&
\textbf{10\%} &
\textbf{20\%} &
\textbf{40\%} &
\textbf{60\%} &
\textbf{80\%} \\

\midrule

\rowcolor{fullkvgray}
Full-KV
& 90.78
& 90.78
& 90.78
& 90.78
& 90.78 \\

\midrule

DefensiveKV
& 2.20
& \lbsecond{6.37}
& \lbsecond{75.21}
& 81.73
& 82.87 \\

Layer-Def.\ KV
& 2.20
& 3.71
& 59.14
& 80.67
& 82.18 \\

CriticalKV
& 1.59
& 3.94
& 10.39
& 30.78
& 75.66 \\

SnapKV
& 1.74
& 2.58
& 6.44
& 20.24
& 51.63 \\

AdaKV
& 2.27
& 3.56
& 13.42
& 38.74
& 74.37 \\

StreamingLLM
& \lbsecond{2.35}
& 2.50
& 68.01
& \lbbest{84.84}
& \lbbest{89.84} \\

\rowcolor{qevictyellow}
\textbf{\myarch{}}
& \lbbest{13.39}
& \lbbest{79.42}
& \lbbest{80.51}
& \lbsecond{84.08}
& \lbsecond{87.34} \\

\bottomrule
\end{tabular*}
\end{subtable}
\hfill
% ------------------------------------------------------------
% Right: KV-cache quantization
% ------------------------------------------------------------
\begin{subtable}[t]{0.31\textwidth}
\centering
\caption{KV-cache quantization}
\label{tab:gsm8k_qwen7b_quantization}

\setlength{\tabcolsep}{5pt}

\begin{tabular*}{\linewidth}{
    @{\extracolsep{\fill}}
    lcc
    @{}
}
\toprule

\textbf{Method} &
\shortstack{\textbf{Equivalent}\\\textbf{KV Memory}} &
\textbf{Score} \\

\midrule

\rowcolor{fullkvgray}
Full-KV
& 100\%
& 90.78 \\

\midrule

% \rowcolor{qevictyellow}
% \textbf{\myarch{}}
% & 10\%
% & 13.39 \\

\rowcolor{qevictyellow}
\textbf{\myarch{}}
& 20\%
& 79.42 \\

KIVI-2b
& $\sim$20\%
& 84.61 \\

ZipCache-4b
& $\sim$22\%
& \lbbest{89.99} \\

KVQuant-2b
& $\sim$15\%
& 77.30 \\

KVQuant-3b
& $\sim$22\%
& \lbsecond{89.76} \\

\bottomrule
\end{tabular*}
\end{subtable}

\endgroup
\end{table*}

% ============================================================
% ABLATIONS
% ============================================================

\section{Ablation Studies}
\label{app:ablations}

We analyse the sensitivity of \myarch{} to its three principal design parameters: routing window size, quantized-tier allocation, and recoverable precision. All experiments use Llama-3.1-8B-Instruct under a \(5\%\) KV-cache budget. To keep the ablation cost manageable while covering the major LongBench capabilities, we select one task from each of its four task families: NarrativeQA for single-document question answering, MuSiQue for multi-document question answering, QMSum for summarization, and TriviaQA for few-shot learning. We vary one parameter at a time while keeping the remaining settings fixed to the default configuration \((\Omega=8,\ q=0.70,\ \mathrm{INT2})\).

\subsection{Window Granularity}
\label{app:ablation_window}

We vary the routing window size over \(\Omega\in\{4,8,16,32\}\), while fixing \(q=0.70\) and using INT2 for the recoverable tier. Smaller windows enable finer-grained allocation, whereas larger windows provide smoother and more stable importance estimates at the cost of coarser selection. This ablation measures how routing granularity affects performance across the four representative tasks.

% Required packages:
% \usepackage{amsmath}
% \usepackage{booktabs}
% \usepackage[table]{xcolor}

\providecommand{\lbbest}[1]{\textbf{#1}}
\providecommand{\lbsecond}[1]{\underline{#1}}
\providecolor{qevictyellow}{RGB}{255,247,204}

\begin{table}[!h]
\centering

\caption{\textbf{Effect of routing-window size on LongBench.}
Results use Llama-3.1-8B-Instruct under a \(5\%\) KV-cache budget with
\(q=0.70\) and INT2 quantization. One representative task is selected
from each LongBench family. The best and second-best results in each
column are shown in bold and underlined, respectively. The highlighted
row denotes the default configuration used in the main experiments.}

\label{tab:ablation_window}

\begingroup
\normalsize
\setlength{\tabcolsep}{3pt}
\renewcommand{\arraystretch}{1.15}

\begin{tabular*}{\columnwidth}{
    @{\extracolsep{\fill}}
    crrrrr
    @{}
}
\toprule

\(\boldsymbol{\Omega}\) &
\textbf{Nar.QA} &
\textbf{MuSiQue} &
\textbf{QMSum} &
\textbf{Tri.QA} &
\textbf{Avg.} \\

\midrule

4
& \lbbest{28.17}
& 30.05
& 23.75
& \lbsecond{89.26}
& 42.81 \\

\rowcolor{qevictyellow}
8
& \lbsecond{27.30}
& \lbbest{32.50}
& \lbbest{23.90}
& \lbbest{89.60}
& \lbbest{43.33} \\

16
& 26.78
& \lbsecond{31.93}
& 23.66
& 89.17
& \lbsecond{42.89} \\

32
& 26.92
& 30.91
& \lbsecond{23.78}
& 87.67
& 42.32 \\

\bottomrule
\end{tabular*}

\endgroup
\end{table}

\subsection{Quantized-Tier Allocation}
\label{app:ablation_q}

We vary the fraction of the historical budget assigned to the recoverable tier over \[ q\in\{0.10,0.30,0.50,0.70,0.90\}, \] while fixing \(\Omega=8\) and using INT2 quantization. Increasing \(q\) expands low-bit historical coverage but reduces the capacity of the full-precision historical tier. This sweep isolates the trade-off between precise retention and broader recoverable context.

% Required packages:
% \usepackage{amsmath}
% \usepackage{booktabs}
% \usepackage[table]{xcolor}

\providecommand{\lbbest}[1]{\textbf{#1}}
\providecommand{\lbsecond}[1]{\underline{#1}}
\providecolor{qevictyellow}{RGB}{255,247,204}

\begin{table}[!h]
\centering

\caption{\textbf{Effect of quantized-tier allocation on LongBench.}
Results use Llama-3.1-8B-Instruct under a \(5\%\) KV-cache budget with
\(\Omega=8\) and INT2 quantization. One representative task is selected
from each LongBench family. The best and second-best results in each
column are shown in bold and underlined, respectively. The highlighted
row denotes the default configuration used in the main experiments.}

\label{tab:ablation_q}

\begingroup
\normalsize
\setlength{\tabcolsep}{3pt}
\renewcommand{\arraystretch}{1.15}

\begin{tabular*}{\columnwidth}{
    @{\extracolsep{\fill}}
    crrrrr
    @{}
}
\toprule

\(\boldsymbol{q}\) &
\textbf{Nar.QA} &
\textbf{MuSiQue} &
\textbf{QMSum} &
\textbf{Tri.QA} &
\textbf{Avg.} \\

\midrule

0.10
& 26.08
& 29.76
& 23.78
& 87.87
& 41.87 \\

0.30
& \lbsecond{27.20}
& 31.08
& 23.47
& 88.49
& 42.56 \\

0.50
& 27.03
& 30.67
& 23.73
& 89.39
& 42.71 \\

\rowcolor{qevictyellow}
0.70
& \lbbest{27.30}
& \lbsecond{32.50}
& \lbbest{23.90}
& \lbsecond{89.60}
& \lbbest{43.33} \\

0.90
& 27.02
& \lbbest{32.58}
& \lbsecond{23.85}
& \lbbest{89.66}
& \lbsecond{43.28} \\

\bottomrule
\end{tabular*}

\endgroup
\end{table}

\subsection{Quantization Precision}
\label{app:ablation_precision}

We compare INT2 and INT4 recoverable representations while fixing \(\Omega=8\) and \(q=0.70\). Under the same total KV-cache budget, INT2 retains more historical windows, whereas INT4 represents fewer windows with lower quantization error. This comparison evaluates whether broader recoverable coverage or higher per-window precision is more beneficial.

% Required packages:
% \usepackage{booktabs}
% \usepackage[table]{xcolor}

\providecommand{\lbbest}[1]{\textbf{#1}}
\providecommand{\lbsecond}[1]{\underline{#1}}
\providecolor{qevictyellow}{RGB}{255,247,204}

\begin{table}[!h]
\centering

\caption{\textbf{Effect of recoverable-tier precision on LongBench.}
Results use Llama-3.1-8B-Instruct under a \(5\%\) KV-cache budget with
\(\Omega=8\) and \(q=0.70\). One representative task is selected from
each LongBench family. The best and second-best results in each column
are shown in bold and underlined, respectively. The highlighted row
denotes the default configuration used in the main experiments.}

\label{tab:ablation_precision}

\begingroup
\normalsize
\setlength{\tabcolsep}{3pt}
\renewcommand{\arraystretch}{1.15}

\begin{tabular*}{\columnwidth}{
    @{\extracolsep{\fill}}
    lrrrrr
    @{}
}
\toprule

\textbf{Precision} &
\textbf{Nar.QA} &
\textbf{MuSiQue} &
\textbf{QMSum} &
\textbf{Tri.QA} &
\textbf{Avg.} \\

\midrule

\rowcolor{qevictyellow}
INT2
& \lbsecond{27.30}
& \lbbest{32.50}
& \lbbest{23.90}
& \lbbest{89.60}
& \lbbest{43.33} \\

INT4
& \lbbest{27.75}
& \lbsecond{30.09}
& \lbsecond{23.88}
& \lbsecond{88.94}
& \lbsecond{42.67} \\

\bottomrule
\end{tabular*}

\endgroup
\end{table}

% ============================================================
% EFFICIENCY
% ============================================================

\section{Efficiency and Systems Analysis}
\label{app:efficiency}

\subsection{Measurement Protocol}
\label{app:efficiency_protocol}

All efficiency measurements use identical hardware, model precision, batch size, sequence lengths, and execution backend within each comparison. We run warm-up iterations before measurement to initialize kernels and memory pools, synchronize the GPU around timed regions, and reset peak-memory statistics immediately before each run.

We report time to first token (TTFT), time per output token (TPOT), aggregate decoding throughput, and peak GPU memory. For a run producing \(N_{\mathrm{out}}\) tokens, with timestamps \(t_{\mathrm{start}}\), \(t_{\mathrm{first}}\), and \(t_{\mathrm{end}}\), we compute
\begin{equation}
    \operatorname{TPOT}
    =
    \frac{t_{\mathrm{end}}-t_{\mathrm{first}}}
         {N_{\mathrm{out}}-1},
    \qquad
    \operatorname{Throughput}
    =
    \frac{N_{\mathrm{out}}}
         {t_{\mathrm{end}}-t_{\mathrm{start}}}.
    \label{eq:appendix_latency_metrics}
\end{equation}

For batched inference, throughput is aggregated across all generated tokens in the batch. Peak GPU memory includes model parameters, persistent KV-cache storage, attention workspaces, and temporary routing buffers. The complete end-to-end results are reported in Table~\ref{tab:efficiency} of the main paper.

\subsection{Runtime Components}
\label{app:runtime_breakdown}

Compared with \fullkv{}, \myarch{} introduces periodic score aggregation, window ranking, tier assignment, and cache migration. First-time demotion also creates the persistent low-bit representation, which is reused during subsequent tier transitions. Since routing occurs every \(\Omega\) generated tokens, its amortized per-token cost is
\begin{equation}
    T_{\mathrm{route/token}}
    =
    \frac{
        T_{\mathrm{score}}
        + T_{\mathrm{rank}}
        + T_{\mathrm{migrate}}
    }{\Omega}.
    \label{eq:appendix_amortized_routing}
\end{equation}

At each decoding step, active quantized windows are dequantized and assembled with the full-precision cache before attention. Routing events additionally materialize attention probabilities for cumulative scoring. These operations explain the backend-dependent behaviour observed in the main paper: reducing the attended full-precision cache benefits eager execution, while the current FlashAttention-2 path incurs overhead from dequantization, cache reconstruction, and selective score materialization. Fusing these operations with mixed-precision attention remains the primary systems optimization opportunity.

\subsection{Hardware Settings}
\label{app:hardware_settings}

All experiments were run on a node with 8 NVIDIA A100 80GB GPUs, with a subset additionally run on 1 NVIDIA GB10 128GB GPU. Larger evaluation sweeps were parallelized by sharding the dataset round-robin across GPUs, with one full model copy per GPU. The software stack was PyTorch 2.6.0 (CUDA 12.4), Transformers 4.47.1 and flash-attn 2.8.3.

% ============================================================
% ADDITIONAL OBSERVATION DETAILS
% ============================================================

\section{Additional Observation Details}
\label{app:observation_details}

This section provides the diagnostic protocol, formal metric definitions, and supporting results for the observations in
Section~\ref{sec:observations}.

\subsection{Diagnostic Protocol}
\label{app:observation_protocol}

We collect reference attention traces from an uncompressed \fullkv{} execution because attention assigned to a discarded state cannot be recovered from the compressed execution itself. Unless stated otherwise, diagnostics are computed independently for each layer and attention head and then aggregated across eligible routing events and traces.

The analysis uses Llama-3.1-8B-Instruct~\citep{dubey2024llama3} with a 512-token prefill, 256 generated tokens, a measured \(20\%\) KV-cache budget, and five protected sink tokens. We evaluate two input articles across all 32 transformer layers, yielding 64 layer-level traces. Routing begins at decoding step zero and is repeated every \(\Omega\) generated tokens. We consider the following conditions:

\begin{itemize}
    \item \textbf{R0}: uncompressed \fullkv{} reference;
    \item \textbf{R1}: token-level eviction with \(\Omega=1\);
    \item \textbf{R2}: two-tier window eviction with \(\Omega=8\);
    \item \textbf{R3}: three-tier routing with \(\Omega=8\),
          \(K_f=7\), and \(K_q=28\);
    \item \textbf{R4}: three-tier routing with \(\Omega=32\),
          \(K_f=1\), and \(K_q=11\).
\end{itemize}

R2 and R3 use the same window size and measured byte budget, isolating the effect of replacing part of the full-precision allocation with a recoverable INT2 tier. R1, R3, and R4 are complete operating points with different routing granularities.

Let \(A^{\ell,h}_{t}(i)\) denote the \fullkv{} attention probability assigned at decoding step \(t\), layer \(\ell\), and head \(h\) to historical token \(i\). The cumulative reference score of window \(w\) is

\begin{equation}
    S^{\ell,h}_{t}(w)
    =
    \sum_{\tau\leq t}
    \sum_{i\in w}
    A^{\ell,h}_{\tau}(i).
    \label{eq:appendix_observation_score}
\end{equation}

\subsection{Observation I: \fmm{} and Selection Churn}
\label{app:observation_window_metrics}

\paragraph{\fmm{} (FMM).}
At routing event \(r\), let \(\mathcal{E}^{\ell,h}_{r}\) denote the historical positions made inaccessible by the evaluated policy. We define FMM over a future horizon \(H\) as

\begin{equation}
    \mathrm{FMM}^{\ell,h}_{H}(r)
    =
    \frac{
        \displaystyle
        \sum_{\tau=r+1}^{\min(r+H,T)}
        \sum_{\substack{
            i\in\mathcal{E}^{\ell,h}_{r}\\
            i\leq r
        }}
        A^{\ell,h}_{\tau}(i)
    }{
        \displaystyle
        \sum_{\tau=r+1}^{\min(r+H,T)}
        \sum_{i\leq r}
        A^{\ell,h}_{\tau}(i)
    }.
    \label{eq:appendix_fmm}
\end{equation}

\noindent Only states that already existed at routing event \(r\) are included. We use \(H=32\). Lower FMM indicates that the routing decision permanently discards less information required by subsequent queries.

\paragraph{Selection Churn.}
Let \(\mathcal{R}^{\ell,h}_{r}\) denote the historical positions retained after routing event \(r\). Window selections are expanded to token positions before comparison. Selection Churn is the Jaccard distance between retained sets at consecutive routing events:

\begin{equation}
    \mathrm{Churn}^{\ell,h}(r)
    =
    1-
    \frac{
        \left|
        \mathcal{R}^{\ell,h}_{r}
        \cap
        \mathcal{R}^{\ell,h}_{r+1}
        \right|
    }{
        \left|
        \mathcal{R}^{\ell,h}_{r}
        \cup
        \mathcal{R}^{\ell,h}_{r+1}
        \right|
    }.
    \label{eq:appendix_selection_churn}
\end{equation}

\noindent The protected recent region is excluded because its deterministic movement reflects the sliding local policy rather than instability in historical selection.

\begin{table}[!h]
\centering
\caption{\textbf{Future Missed Mass and historical-set churn.} FMM is averaged over eligible routing events with \(H=32\). Churn values match those reported in Section~\ref{sec:observations}.}
\label{tab:appendix_window_diagnostics}
\small
\setlength{\tabcolsep}{5pt}
\renewcommand{\arraystretch}{1.06}

\begin{tabular}{lcc}
\toprule
\textbf{Condition}
& \textbf{FMM} \(\downarrow\)
& \textbf{Churn/Route} \(\downarrow\) \\
\midrule

R1: \(\Omega=1\)
& \(51.73\%\)
& \(0.0170\) \\

R3: \(\Omega=8\)
& \(\mathbf{46.38\%}\)
& \(\mathbf{0.0012}\) \\

R4: \(\Omega=32\)
& \(47.38\%\)
& \(0.0180\) \\

R3: FP tier only
& \(60.19\%\)
& -- \\

\bottomrule
\end{tabular}
\end{table}

\noindent Table~\ref{tab:appendix_window_diagnostics} shows that R3 reduces average FMM relative to token-level R1 while lowering historical-set churn by more than an order of magnitude. R4 also reduces FMM relative to R1, but its coarser routing granularity does not improve churn in this measurement. Window aggregation is therefore beneficial, although excessively large windows can sacrifice allocation granularity and need not yield more stable assignments.

The R3 full-precision-only ablation treats quantized windows as inaccessible without reallocating their bytes. Its substantially higher FMM isolates the future attention preserved by the recoverable tier. This is a mechanistic ablation rather than an iso-memory comparison.

\subsection{Observation II: Tier-Mass Distribution and Quantized-Score Agreement}
\label{app:observation_tier_metrics}

For tier \(z\in \{\textsc{Full},\textsc{Quantized},\textsc{Evicted},\textsc{Local}\}\), we define its cumulative attention-mass share as

\begin{equation}
    \mu_t(z)
    =
    \frac{
        \displaystyle
        \sum_{w:z_t(w)=z} S_t(w)
    }{
        \displaystyle
        \sum_{w\in\mathcal{W}_t} S_t(w)
    }.
    \label{eq:appendix_tier_mass}
\end{equation}

\noindent Under the byte-matched R2--R3 comparison, R2 assigns \(33.3\%\) of the reference attention mass to full-precision windows and \(64.8\%\) to evicted windows. R3 assigns \(19.7\%\) to full precision and preserves another \(42.6\%\) in INT2, reducing the evicted share to \(35.7\%\). The protected local region contributes \(1.9\%\) in both conditions. The recoverable tier therefore reduces the reference attention mass assigned to permanent eviction by \(29.1\) percentage points.

To determine whether quantized windows remain suitable for subsequent ranking, let \(\mathbf{s}^{Q}_{r}\) denote their scores under INT2 execution and \(\mathbf{s}^{\mathrm{Full}}_{r}\) the corresponding \fullkv{} scores. We define Quantized-Score Agreement as

\begin{equation}
    \mathrm{QSA}(r)
    =
    \frac{
        \left\langle
            \mathbf{s}^{Q}_{r},
            \mathbf{s}^{\mathrm{Full}}_{r}
        \right\rangle
    }{
        \left\|
            \mathbf{s}^{Q}_{r}
        \right\|_2
        \left\|
            \mathbf{s}^{\mathrm{Full}}_{r}
        \right\|_2
    }.
    \label{eq:appendix_quantized_score_agreement}
\end{equation}

\noindent We additionally measure the ratio between the total attention mass assigned to these windows under quantized and full-precision execution:

\begin{equation}
    R_Q(r)
    =
    \frac{
        \displaystyle
        \sum_{w\in\mathcal{Q}_r} S^{Q}_r(w)
    }{
        \displaystyle
        \sum_{w\in\mathcal{Q}_r} S^{\mathrm{Full}}_r(w)
    }.
    \label{eq:appendix_quantized_mass_ratio}
\end{equation}

\begin{table}[!h]
\centering
\caption{\textbf{Agreement between INT2 and \fullkv{} scores over R3
quantized windows.}}
\label{tab:appendix_quantized_agreement}
\small
\setlength{\tabcolsep}{5pt}
\renewcommand{\arraystretch}{1.06}

\begin{tabular}{lc}
\toprule
\textbf{Statistic} & \textbf{Value} \\
\midrule
Mean cosine agreement
& \(0.9824\) \\
First / last routing event
& \(1.0000 / 0.9676\) \\
Per-layer minimum / maximum
& \(0.9503 / 0.9958\) \\
Per-head minimum / maximum
& \(0.7874 / 0.9995\) \\
Mean attention-mass ratio
& \(0.8660\) \\
\bottomrule
\end{tabular}
\end{table}

\noindent The high cosine agreement in Table~\ref{tab:appendix_quantized_agreement} shows that INT2 largely preserves the relative ordering of quantized windows. The lower mass ratio indicates attenuation in attention magnitude, so low-bit execution is not numerically identical to \fullkv{}. Nevertheless, the ranking signal remains sufficiently stable for continued scoring and subsequent promotion.

\subsection{Observation III: \lir{}}
\label{app:observation_lir}

Let \(X_{r,w}\in\{0,1\}\) indicate whether window \(w\) belongs to a selected full-precision set at routing event \(r\). We analyse two selections:

\begin{itemize}
    \item \textbf{Oracle}: the top-\(K_f\) windows according to the R0
          \fullkv{} ranking;
    \item \textbf{Policy FP}: the full-precision tier selected by R3 or R4.
\end{itemize}

\noindent An inactive episode is a maximal sequence of routing events for which \(X_{r,w}=0\). After \(m\) consecutive inactive events, the episode becomes eligible. It is rescued if the window re-enters the selected set at any later routing event before generation terminates. We define

\begin{equation}
    \mathrm{GlobalLIR}(m)
    =
    \frac{
        \#\{
        \text{eligible inactive episodes that later re-enter}
        \}
    }{
        \#\{
        \text{eligible inactive episodes}
        \}
    }.
    \label{eq:appendix_global_lir}
\end{equation}

\noindent Each maximal inactive episode contributes once. Episodes that do not return before generation terminates remain in the denominator and are counted as not rescued. We use \(m=3\) by default and verify the same trend for \(m\in\{1,2,4,8\}\).

\begin{table}[!h]
\centering
\caption{\textbf{\lir{} for oracle and policy full-precision selections.} Confidence intervals are \(95\%\). Median time-to-return is measured in routing events.}
\label{tab:appendix_global_lir}
\small
\setlength{\tabcolsep}{4pt}
\renewcommand{\arraystretch}{1.06}

\begin{tabular}{llccc}
\toprule
\textbf{Run}
& \textbf{Selection}
& \textbf{\lir}
& \textbf{Rescued}
& \textbf{Median TTR} \\
\midrule

R3
& Oracle
& \(0.98\%\,[0.92\%,1.03\%]\)
& \(51\)
& \(12\) \\

R3
& Policy FP
& \(\mathbf{6.18\%}\,[6.09\%,6.27\%]\)
& \(341\)
& \(8\) \\

R4
& Oracle
& \(0.09\%\,[0.00\%,0.17\%]\)
& \(1\)
& \(4\) \\

R4
& Policy FP
& \(\mathbf{2.59\%}\,[1.90\%,3.27\%]\)
& \(30\)
& \(2\) \\

\bottomrule
\end{tabular}
\end{table}

\noindent The low oracle LIR in Table~\ref{tab:appendix_global_lir} confirms that the true top-ranked region is largely persistent. Under R3, however, the deployed full-precision tier exhibits a \(6.18\%\) return rate, approximately \(6.3\times\) the oracle rate. Approximate scoring and constrained allocation therefore produce substantially greater re-entry pressure than the oracle ranking alone suggests.

The median policy return times are eight routing events for R3 and two for R4. Given routing intervals of eight and 32 tokens, respectively, both correspond to approximately 64 decoding steps. Relevance revival can therefore occur well beyond the immediately following routing event.

We complement \lir{} with the lagged transition probability

\begin{equation}
    P^{(\Delta)}_{ab}
    =
    \Pr
    \left[
        X_{r+\Delta,w}=b
        \mid
        X_{r,w}=a
    \right],
    \qquad
    a,b\in\{0,1\}.
    \label{eq:appendix_binary_transition}
\end{equation}

\noindent For the R3 policy, \(P^{(\Delta)}_{01}\) increases from \(0.31\%\) at
\(\Delta=1\) to \(2.39\%\) at \(\Delta=8\), while
\(P^{(\Delta)}_{10}\) increases from \(3.05\%\) to \(22.12\%\). At
\(\Delta=8\), departures from the full-precision tier are approximately
\(9.3\times\) more frequent than returns.

These dynamics are asymmetric. Most oracle-important windows remain active, many policy-selected windows subsequently decline, and a smaller but systematic subset eventually returns. This behaviour motivates frequent demotion together with recoverable low-bit storage that preserves the possibility of later promotion.

\section{Compatibility with Alternative Ranking Functions}
\label{app:ranking_generality}

The three-tier hierarchy in \myarch{} requires only an ordering over candidate windows and is therefore not tied to cumulative attention as its importance estimator. To evaluate this modularity, we replace the default ranking function with three representative alternatives while leaving the cache hierarchy, byte allocation, and migration policy unchanged.

All experiments use Llama-3.1-8B-Instruct on LongBench under a \(5\%\) KV-cache budget with \(\Omega=8\), \(q=0.70\), and an INT2 recoverable tier. The evaluation covers the 12 tasks used in the main study together with PassageCount and PassageRetrieval-en. We compare the following scoring functions:

\begin{itemize}
    \item \textbf{Cumulative attention}, the default estimator used by \myarch{}.

    \item \textbf{Exponentially decayed attention}, which discounts older observations and places greater weight on recent queries. This variant is motivated by historical-attention and persistence-based cache selection \citep{zhang2023h2o,liu2023scissorhands}.

    \item \textbf{Attention \(p\)-norm}, which replaces additive aggregation with a peak-sensitive \(p\)-norm over observed window scores. We use \(p=3\). This is a power-norm generalization of the attention-based ranking signals used by method such SnapKV~\citep{li2024snapkv}.

    \item \textbf{Key-vector norm}, an attention-free estimator based on the \(L_2\)-norm of cached keys. Following \citet{devoto2024simple}, lower key norms receive higher retention priority.
\end{itemize}

Let \(u_t^\ell(w)\) denote the attention received by window \(w\) since the previous routing event:

\begin{equation}
    u_t^\ell(w)
    =
    \frac{1}{H_q}
    \sum_{h=1}^{H_q}
    \sum_{\tau=t-\Omega+1}^{t}
    \sum_{i\in w}
    a_{\tau}^{\ell,h}(i).
    \label{eq:ranking_event_score}
\end{equation}

The cumulative and exponentially decayed scores are

\begin{align}
    S_{t,\mathrm{cum}}^\ell(w)
    &=
    S_{t-\Omega,\mathrm{cum}}^\ell(w)
    + u_t^\ell(w),
    \label{eq:ranking_cumulative}
    \\
    S_{t,\mathrm{decay}}^\ell(w)
    &=
    \lambda S_{t-\Omega,\mathrm{decay}}^\ell(w)
    + u_t^\ell(w),
    \qquad \lambda=0.99.
    \label{eq:ranking_decay}
\end{align}

For the \(p\)-norm variant, we aggregate the event-level window scores as

\begin{equation}
    S_{t,p}^\ell(w)
    =
    \left(
        \sum_{r\leq t}
        \left[u_r^\ell(w)\right]^p
    \right)^{1/p},
    \qquad p=3.
    \label{eq:ranking_pnorm}
\end{equation}

The key-norm estimator is defined as

\begin{equation}
    S_{\mathrm{key}}^\ell(w)
    =
    -\frac{1}{|w|}
    \sum_{i\in w}
    \left\|
        \mathbf{k}_i^\ell
    \right\|_2,
    \label{eq:ranking_key_norm}
\end{equation}

where the negative sign converts the low-norm preference into the
higher-is-better convention used by the routing algorithm.

\begin{table*}[t]
\centering
\caption{\textbf{Compatibility of \myarch{} with alternative window-ranking functions on LongBench.} All results use Llama-3.1-8B-Instruct under a \(5\%\) KV-cache budget with \(\Omega=8\), \(q=0.70\), and INT2 quantization. Only the ranking function is changed. The average is computed over all 14 tasks. Best and second-best distinct results are shown in bold and underlined, respectively.}
\label{tab:ranking_generality}

\scriptsize
\setlength{\tabcolsep}{2.5pt}
\renewcommand{\arraystretch}{1.08}

\resizebox{\textwidth}{!}{%
\begin{tabular}{l*{15}{r}}
\toprule
\multirow{2}{*}{\textbf{Ranking function}}
& \multicolumn{3}{c}{\textbf{Single-Document QA}}
& \multicolumn{3}{c}{\textbf{Multi-Document QA}}
& \multicolumn{3}{c}{\textbf{Summarization}}
& \multicolumn{3}{c}{\textbf{Few-Shot Learning}}
& \multicolumn{2}{c}{\textbf{Synthetic}}
& \multirow{2}{*}{\textbf{Avg.}} \\
\cmidrule(lr){2-4}
\cmidrule(lr){5-7}
\cmidrule(lr){8-10}
\cmidrule(lr){11-13}
\cmidrule(lr){14-15}
& Nar.QA
& Qasper
& MF-en
& Hot.QA
& 2Wi.QA
& MuSiQue
& Gov.Re.
& QMSum
& M.News
& TREC
& Tri.QA
& SAMSum
& PCount
& PR-en
& \\
\midrule

Cumulative attention
& 27.30
& 28.00
& \underline{43.50}
& \underline{56.40}
& 45.10
& \underline{32.50}
& \underline{28.80}
& \underline{23.90}
& \textbf{19.30}
& \textbf{63.00}
& \underline{89.60}
& 38.20
& \underline{8.67}
& \underline{92.00}
& \underline{42.59} \\

Exp.\ decay, \(\lambda=0.99\)
& 27.34
& \underline{28.03}
& \textbf{43.54}
& \textbf{56.47}
& \underline{45.19}
& \textbf{32.59}
& \textbf{29.11}
& \textbf{24.13}
& 19.20
& \textbf{63.00}
& \textbf{89.66}
& 38.21
& \underline{8.67}
& \underline{92.00}
& \textbf{42.65} \\

Attention \(p\)-norm, \(p=3\)
& \textbf{29.64}
& \textbf{28.80}
& 38.55
& 54.54
& \textbf{46.51}
& 30.98
& 26.49
& 23.84
& \underline{19.27}
& \underline{61.00}
& 87.47
& \textbf{40.03}
& \textbf{9.00}
& \textbf{96.00}
& 42.29 \\

Key \(L_2\)-norm
& \underline{28.73}
& 24.91
& 40.02
& 54.12
& 43.56
& 30.02
& 27.34
& 22.52
& 18.57
& \underline{61.00}
& 87.31
& \underline{39.79}
& \textbf{9.00}
& \underline{92.00}
& 41.35 \\

\bottomrule
\end{tabular}%
}
\end{table*}

% Requires:
% \usepackage{booktabs}

\section{Effect of Q$\rightarrow$F Residency Promotion}
\label{app:promotion_ablation}

\paragraph{Objective.}
We examine whether allowing quantized windows to return to the full execution tier improves the behavior of the R3 three-tier policy. We compare the one-way policy
\[
\textsc{Full}
\rightarrow
\textsc{Quantized}
\rightarrow
\textsc{Evicted}
\]
against the bidirectional policy
\[
\textsc{Full}
\leftrightarrow
\textsc{Quantized}
\rightarrow
\textsc{Evicted}.
\]
A promoted window is reconstructed from its persistent INT2 representation. Promotion therefore restores full-tier residency and execution handling rather than the original pre-quantization KV values.

\paragraph{Experimental setting.}
We use Llama-3.1-8B-Instruct~\citep{dubey2024llama3} with a 256-token prefill, 1024 decoding tokens, a measured 20\% persistent KV-cache budget, five protected sink tokens, and the R3 three-tier configuration with window size $\Omega=8$. The two variants use the same tier capacities, cumulative-attention scoring rule, routing schedule, quantization configuration, and memory budget. They differ only in whether a window already assigned to the quantized tier may later be promoted to the full execution tier. Under the no-promotion variant, quantized windows may subsequently be evicted but cannot return to the full tier.

The current pilot conditions use different source articles. The comparison should therefore be interpreted as directional evidence of the behavior enabled by promotion rather than as a paired estimate of its exact causal effect.

\paragraph{Metrics.}
We reuse the diagnostics introduced in Appendix~\ref{app:observation_details}: Future Missed Mass (FMM), Quantized-Score Agreement (QSA), the attention-mass ratio $R_Q$, and Global LIR. We additionally report the number of observed Q$\rightarrow$F transitions as an auxiliary measure of promotion activity.

For this analysis, FMM measures the fraction of future FullKV attention assigned to historical states that have become inaccessible under R3. Lower FMM indicates that fewer future-relevant states are permanently discarded.

QSA measures the cosine agreement between the score vector of the windows currently assigned to the quantized tier and their corresponding FullKV reference scores:
\[
\operatorname{QSA}(r)
=
\frac{
    \left\langle
        \mathbf{s}^{Q}_{r},
        \mathbf{s}^{\mathrm{Full}}_{r}
    \right\rangle
}{
    \left\lVert \mathbf{s}^{Q}_{r} \right\rVert_2
    \left\lVert \mathbf{s}^{\mathrm{Full}}_{r} \right\rVert_2
}.
\]

\noindent
The attention-mass ratio measures the fraction of the FullKV reference attention mass preserved over the windows currently assigned to the quantized tier:
\[
R_Q(r)
=
\frac{
    \sum_{w\in Q_r} S^{Q}_{r}(w)
}{
    \sum_{w\in Q_r} S^{\mathrm{Full}}_{r}(w)
}.
\]
We report $R_Q$ as a percentage. A value of $100\%$ indicates that quantized execution preserves the total FullKV attention mass over the current Q-tier membership.

Global LIR measures the percentage of eligible inactive episodes that later re-enter the selected full-tier set. The transition count records the total number of Q$\rightarrow$F movements observed during decoding.

\begin{table}[t]
\centering
\small
\setlength{\tabcolsep}{4.5pt}
\caption{
Effect of Q$\rightarrow$F residency promotion for the R3 three-tier policy with window size $\Omega=8$. QSA and $R_Q$ are computed over the windows currently assigned to the quantized tier. Lower is better for FMM; higher is better for the remaining metrics.
}
\label{tab:promotion_ablation}
\begin{tabular}{lccc}
\toprule
\textbf{Metric}
& \textbf{No promotion}
& \textbf{With promotion}
& \textbf{$\Delta$} \\
\midrule

R3 FMM $\downarrow$
& 16.8\%
& \textbf{16.1\%}
& $-0.7$ pp \\

Quantized-Score Agreement $\uparrow$
& 0.947
& \textbf{0.985}
& $+0.038$ \\

FullKV attention mass preserved, $R_Q$ $\uparrow$
& 88.7\%
& \textbf{95.5\%}
& $+6.8$ pp \\

Global LIR $\uparrow$
& 0.0\%
& \textbf{6.6\%}
& $+6.6$ pp \\

Recorded Q$\rightarrow$F transitions
& 0
& \textbf{639}
& $+639$ \\

\bottomrule
\end{tabular}
\end{table}

\paragraph{Observation 1: Promotion is associated with lower future missed mass.} R3 FMM decreases from 16.8\% under one-way demotion to 16.1\% when Q$\rightarrow$F promotion is enabled. This directional reduction suggests that bidirectional routing can reduce the premature permanent removal of windows whose importance later re-emerges.

\paragraph{Observation 2: Promotion improves policy-level Q-tier score agreement.} QSA increases from 0.947 to 0.985, showing that the score vectors of the windows currently remaining in the quantized tier align more closely with their FullKV reference scores under the promotion-enabled policy.

\paragraph{Observation 3: Promotion preserves more FullKV attention mass in the Q tier.} The attention-mass ratio $R_Q$ increases from 88.7\% to 95.5\%, a gain of 6.8 percentage points. Thus, the current Q-tier population under bidirectional routing preserves a larger fraction of its FullKV reference attention mass and moves closer to the ideal value of 100\%.

Because promotion changes which windows remain in the quantized tier, the improvements in QSA and $R_Q$ should be interpreted as evidence of better tier composition and policy-level score agreement. They do not imply that promotion reverses the quantization error of a fixed window.

\paragraph{Observation 4: Promotion is actively exercised.} Global LIR increases from 0\% under the one-way policy to 6.6\% when promotion is enabled. The zero value without promotion holds by construction, whereas the nonzero value confirms that previously inactive windows re-enter the full-tier selection under bidirectional routing.

\paragraph{Observation 5: Re-entry is frequent rather than purely hypothetical.} The promotion-enabled run records 639 Q$\rightarrow$F transitions, compared with none under one-way demotion. This shows that promotion is an active component of the routing policy rather than only a permitted but unused state transition.

\paragraph{Overall observation.} Taken together, the lower FMM, higher QSA and $R_Q$, and nonzero re-entry activity provide consistent directional evidence that Q$\rightarrow$F promotion improves the dynamic allocation of historical windows across the full and quantized tiers.

\end{document}